\documentclass{article} 
\usepackage{iclr2027_conference,times}

\usepackage{amsmath,amsfonts,bm}

\def\eqref#1{equation~\ref{#1}}

\def\1{\bm{1}}

\DeclareMathAlphabet{\mathsfit}{\encodingdefault}{\sfdefault}{m}{sl}
\SetMathAlphabet{\mathsfit}{bold}{\encodingdefault}{\sfdefault}{bx}{n}

\usepackage{hyperref}
\hypersetup{hidelinks}
\usepackage{url}
\usepackage{amsmath,amssymb}
\usepackage{booktabs}
\usepackage{multirow}
\usepackage{graphicx}
\usepackage{xcolor}
\usepackage{wrapfig}   

\newcommand{\fitwidth}[1]{\resizebox{\ifdim\width>\linewidth\linewidth\else\width\fi}{!}{#1}}
\newcommand{\sig}[1]{\textsuperscript{#1}}
\newcommand{\cmark}{\ensuremath{\checkmark}}
\newcommand{\xmark}{\ensuremath{\times}}

\newif\ifshowcomments
\showcommentstrue

\title{On the Token Collapse Phenomenon in\\ Textual Graph Soft Prompting}

\author{%
  \parbox{\textwidth}{%
    \centering
    \normalfont\normalsize
    \setlength{\tabcolsep}{0pt}
    \begin{tabular}{@{}c@{\hspace{1.5em}}c@{\hspace{1.5em}}c@{}}
      \textbf{Andrea Giuseppe Di Francesco}\textsuperscript{1,2,*} &
      \textbf{Rishabh Jain}\textsuperscript{3} &
      \textbf{Franco Maria Nardini}\textsuperscript{2}
      \\[0.55em]
      \textbf{Nicola Tonellotto}\textsuperscript{4} &
      \textbf{Pietro Li\`{o}}\textsuperscript{3} &
      \textbf{Fabrizio Silvestri}\textsuperscript{1}
    \end{tabular}
    \\[1em]
    {\small
      \textsuperscript{1}Sapienza University of Rome, Rome, Italy\\[0.2em]
      \textsuperscript{2}ISTI-CNR, Pisa, Italy\\[0.2em]
      \textsuperscript{3}University of Cambridge, Cambridge, UK\\[0.2em]
      \textsuperscript{4}University of Pisa, Pisa, Italy
      \textsuperscript{*}Corresponding author
    }
    \\[0.7em]
    {\footnotesize
      \href{mailto:difrancesco@diag.uniroma1.it}{%
        \texttt{difrancesco@diag.uniroma1.it}%
      }
    }
  }%
}
\iclrfinalcopy 
\begin{document}

\maketitle
\fancyhead{}         
\lhead{Preprint}

\begin{abstract}
Recent methods for injecting graph structure into LLM prompts include \textit{graph soft prompting}, in which a graph neural network (GNN) learns graph-specific tokens that capture relational information among entities in textual prompts.  
A major development in this line of work is G-Retriever, which combines graph soft prompts with a verbalized description of the same graph to strengthen the textual prompt signal. In this paper, we conduct an in-depth analysis of G-Retriever and uncover a pathological behavior we term \emph{token collapse}. Across three benchmarks, three graph encoders, two LLM families, and both frozen and LoRA-adapted models, G-Retriever's graph tokens become almost indistinguishable across inputs. We provide evidence that the same failure is inherited by subsequent methods to G-Retriever, even with multi-token pooling. Under the standard setup, swapping a token with the dataset mean, a token from another example, or one derived from a corrupted graph produces no resolved answer-level effect in accuracy. Moreover, a learned constant prompt that never accesses the graph matches or surpasses G-Retriever on all three benchmarks. These findings indicate that, rather than conveying graph content as intended, the collapsed token functions primarily as a generic soft prompt.  
We examine conditions linked to this collapse and propose a two-stage procedure that yields non-collapsed tokens while maintaining competitive performance in most cases. However, token variability alone does not prove that the model is using graph information: via interventions and decoding probes, we show that depending on the configuration, a token can reflect the question, an answer prior, or graph-derived content that the LLM may or may not actually leverage. Overall, our results offer a practical approach to evaluating graph-token alignment rather than inferring it from task accuracy alone. Code to reproduce our experiments is available at \url{https://github.com/difra100/Token_Collapse_KGQA}.
\end{abstract}

\section{Introduction}
\label{sec:intro}

Large language models (LLMs) store knowledge in their parameters, making information outside or newer than their pre-training data hard to access or update~\citep{lewis2020retrieval}. Such external knowledge may be organised as text-attributed graphs~\citep{he2024gretriever}. We can incorporate this information into an LLM either by \emph{textualizing the graphs}, i.e., constructing verbalized descriptions of nodes and edges, or through \emph{graph soft prompting}, i.e., providing a learned representation generated by a graph neural network (GNN). Verbalisation has two limitations: its token cost grows with the amount of graph content included, and serialisation imposes a sequence order on facts that have no intrinsic order. Expressive Graph soft prompting aims to offer an alternative interface for conveying relational information without verbalising every node and edge. G\nobreakdash-Retriever~\citep{he2024gretriever} supplies the same underlying graph via both channels. However, this design introduces an identifiability issue: because the graph content is already present in textual form, improvements in answer accuracy alone cannot distinguish whether the model relies on the graph token, the verbalized graph, or both. Furthermore, the training objective does not explicitly enforce that the graph-derived token encodes graph-specific information when the textual channel suffices. Consequently, a token computed from the graph may degenerate into a graph-invariant soft prompt.

\begin{wrapfigure}{r}{0.43\textwidth}
\centering
\includegraphics[width=\linewidth]{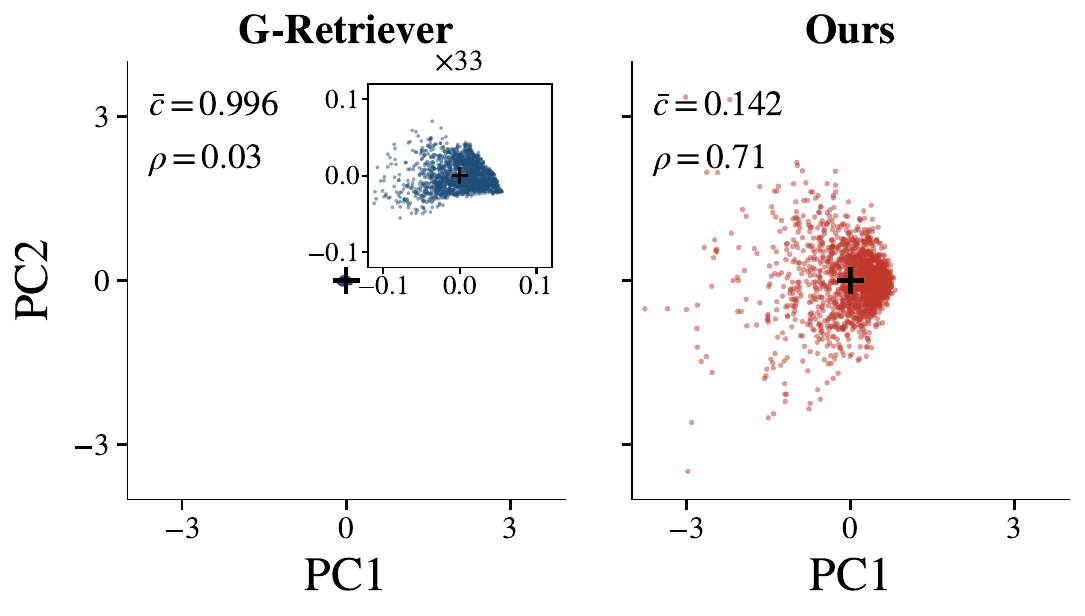}
\caption{\textbf{Graph tokens of all 1,628 \texttt{WebQSP} test questions}, projected onto two principal components after removing the mean token $\boldsymbol{\mu}$, with pairwise cosines shown below.}
\label{fig:tokenspace}
\end{wrapfigure}

G-Retriever interprets its ablations as evidence that the graph encoder and verbalised graph provide complementary graph representations~\citep{he2024gretriever}. We revisit this interpretation by testing what the graph token contains and whether the LLM uses it.

This study evaluates the proposed mechanism by quantifying both the information encoded in graph-derived tokens and the information that the large language model (LLM) causally exploits during inference. The experimental protocol integrates token-space statistical analyses with fixed-weight causal interventions: individual tokens are substituted with (i) the training-set mean token, (ii) a token sampled from a different example, or (iii) a token recomputed from corrupted or different graph inputs. In addition to answer accuracy, we assess gold-answer log-likelihood to capture subtler shifts in the model’s output distribution, and we employ decoding-based probes to measure the recoverability of question, answer, and graph attributes from internal representations. Graph-free control conditions preserve the projector architecture and the full training pipeline while ablating the graph input itself.

Applying this protocol to G-Retriever uncovers an unanticipated phenomenon that we call \emph{token collapse}, which also recurs in subsequent graph-prompting methods. Specifically, across several graph encoders, benchmark datasets, LLM families, and fine-tuning regimes, the token representations generated by G-Retriever are nearly invariant with respect to the input graph (Figure~\ref{fig:tokenspace}). Moreover, substituting these input-dependent projected tokens with a fixed alternative yields no resolved answer-level effect under the default configuration; indeed, a learned constant token that never observes any graph matches or even surpasses the performance of the full model. These results indicate that the collapsed token operates primarily as a soft prompt rather than as a carrier of graph-specific information. Although likelihood-based analyses show that perturbing the token can modulate the output distribution, such sensitivity alone is insufficient to demonstrate input- or graph-conditioned utilization.

We associate collapse with two factors: a projector that suppresses variation in the encoder representations and an answer objective that does not require this variation to be preserved. Our ablations implicate both factors, although they do not isolate either as the sole cause. We address them with a two-stage pipeline that first aligns and freezes the graph encoder, then trains separate graph-token and prompt-token pathways. The resulting tokens no longer collapse, but non-collapse alone guarantees neither graph use nor improved accuracy.
Depending on the benchmark and training configuration, they may encode the question, an answer prior, graph information made redundant by the text, or graph information that the LLM uses to answer. This distinction matters because task improvements may otherwise be attributed to graph--language alignment even when they arise from the verbalised graph or a graph-independent soft prompt. By separating token variation, graph content, and downstream use, our protocol makes alignment empirically testable and provides concrete criteria for developing graph--language models that preserve and use task-relevant graph information beyond what is already supplied by the textual channel.
\paragraph{Contributions.}
We identify \emph{token collapse} across G-Retriever benchmarks, graph encoders,
LLM families, and tuning regimes, and find the same failure in recent
graph-prompting methods. Using matched graph-free controls and fixed-weight
interventions, we show that the collapsed token acts primarily as a soft prompt
rather than as a graph representation used to answer. We associate collapse
with the projector and end-to-end answer training, and propose a two-stage
pipeline that mitigates it across the tested settings. Finally, we show that a
non-collapsed token may encode \textit{question} or \textit{answer priors}, \textit{unused graph
information}, or \textit{graph information} that changes the answer. Together, these
results provide a practical protocol for testing graph-token alignment rather
than inferring it from task accuracy.

\section{Related work}
\label{sec:related}

\textbf{Graph soft prompting.}
Soft prompting conditions an LLM with learned continuous vectors \citep{pt,li2021prefixtuningoptimizingcontinuousprompts}; graph soft prompting instead projects graph-encoder outputs into one or more such tokens. GraphToken~\citep{fatemi2024talk}, GraphLLM~\citep{chai2024graphllm},
GNP~\citep{tian2024gnp}, GraphGPT~\citep{tang2024graphgpt}, and G-Retriever~\citep{he2024gretriever} differ in their encoders, token counts, retrieval, and LLM adaptation. In this paper, we test whether G-Retriever and subsequent methods preserve and use graph-specific information in their projected tokens.

\textbf{Evaluating graph--language models.}
Recent work questions whether existing objectives and benchmarks require graph--language integration. \citet{GTLLM} show that autoregressive training can favour a text-dominant solution, while \citet{petkar2026graph} find that single-modality and raw soft-prompt baselines remain competitive and introduce
CLEGR to test compositional graph--text reasoning. We complement these studies by auditing the graph token-- LLM interface of graph soft prompting. Extended related work discussion is in Appendix~\ref{app:related}.

\section{Method}
\label{sec:method}

\subsection{Preliminaries: graph soft prompting in G-Retriever}
\label{sec:prelim}

A \emph{textual graph}
$G=(\mathcal{V},\mathcal{E},\{x_v\}_{v\in\mathcal{V}},
\{x_e\}_{e\in\mathcal{E}})$
associates every node $v$ with a text attribute $x_v$ and every edge $e$
with a text attribute $x_e$. Each example $i$ consists of a graph, a question
$q_i$, and an answer $y_i$.

\textbf{Retrieval.}
Node, edge, and question texts are embedded by a frozen sentence encoder $\mathrm{LM}(\cdot)$ into $\mathbb{R}^{d}$~\citep{sbert}. The respective representations are $\mathbf{z}_v=\mathrm{LM}(x_v)$, $\mathbf{z}_e=\mathrm{LM}(x_e)$, and
$\mathbf{q}_i=\mathrm{LM}(q_i)$.

When the complete graph exceeds the LLM's context length or a predefined budget, only a relevant subgraph can be supplied to the model. Following G-Retriever, similarities between $\mathbf{q}_i$ and the node and edge embeddings define retrieval relevance, and a prize-collecting Steiner tree is used to select a compact, connected subgraph~\citep{bienstock1993pcst}. We denote the resulting question-specific subgraph by $G_i$. When the complete graph fits within the budget, no retrieval is needed and $G_i=G$.

\textbf{Graph token.}
A graph encoder $f_\phi$ processes $G_i$ using node features $\mathbf{z}_v$, edge features $\mathbf{z}_e$, and optionally the question representation $\mathbf{q}_i$. G-Retriever uses a graph transformer by
default~\citep{shi2020graphtransformer}; we also evaluate GAT~\citep{velickovic2018gat}
and GCN~\citep{kipf2017gcn}. The node representations are mean-pooled into
$\mathbf{h}_i\in\mathbb{R}^{d_g}$, and a projector $P_\psi$ maps the result
into the LLM embedding space:
\begin{equation}
\mathbf{t}_i
=
P_\psi(\mathbf{h}_i)
=
\mathbf{W}_2
\sigma(\mathbf{W}_1\mathbf{h}_i+\mathbf{b}_1)
+\mathbf{b}_2
\in\mathbb{R}^{d_\ell},
\label{eq:projector}
\end{equation}
where $\sigma$ is the logistic sigmoid and the projector has hidden width
$d_p$. We refer to $\mathbf{t}_i$ as the \emph{graph token}.

\textbf{Generation.}
The retrieved subgraph is also verbalised as a list of nodes and edges, denoted by $\tau(G_i)$, and truncated to a budget of $L_{\max}$ tokens. In the default frozen-LLM setting, the model receives
\[
[\,\mathbf{e}_{\mathrm{bos}};\,
\mathbf{t}_i;\,
\mathrm{Emb}(\tau(G_i));\,
\mathrm{Emb}(q_i)\,].
\]
The graph token therefore precedes the verbalised graph and the question.
The LLM parameters $\theta$ remain fixed, while the graph encoder and
projector are trained with the answer loss
\begin{equation}
\mathcal{L}_{\mathrm{QA}}(\phi,\psi)
=
-\sum_i\sum_k
\log p_\theta
\bigl(
y_{i,k}\mid
y_{i,<k},\mathbf{t}_i,\tau(G_i),q_i
\bigr).
\label{eq:qa}
\end{equation}
The graph encoder does not receive the question directly, so in G-Retriever
the token depends on the question only through retrieval.

\subsection{The Graph Token Collapse Phenomenon}
\label{sec:protocol}

Task accuracy conflates three distinct questions: \emph{Does the token vary across inputs? Does the LLM use the graph token? What information does it contain?} We address them with complementary measurements of token geometry, fixed-weight interventions, gold-answer likelihood, and decodability.

\textbf{Does the token vary across inputs?}
We measure inter-sample cosine, the input-specific residual relative to the mean token, and token norm relative to ordinary word embeddings. Collapse appears as cosine near one and a small residual ratio. We also compute inter-sample cosine before projection, on the pooled encoder representations $\mathbf{h}_i$, to determine where collapse arises. Formal definitions are
given in Appendix~\ref{app:protocol}.

\textbf{Does the LLM use graph-specific token variation?} At fixed weights, \textsc{real} uses the original token, \textsc{mean} replaces it with the training-set mean, and \textsc{swap} uses another example's token. Our main graph-specific intervention, \textsc{gswap}, recomputes the token from another graph while preserving the original question conditioning. \textsc{node}, \textsc{edge}, and \textsc{rel} corrupt individual graph components, while \textsc{notxt} removes the verbalised graph. Retrained no-entity and no-question variants isolate the information available during training. Because accuracy observes only changes in the decoded answer, we also measure the paired change in gold-answer negative log-likelihood under teacher forcing. This detects smaller changes in the output distribution. Full details on our interventions are in Appendix~\ref{app:protocol}.

\textbf{What does the token contain?}
\label{sec:decodability}
Recoverable content and downstream use are not equivalent: a token may contain information that the LLM ignores. This matters for our pipeline because Stage~1 is conditioned on the question and trained with an answer-related target. A non-collapsed token may therefore encode the \textbf{question}, an \textbf{answer} prediction or prior, the \textbf{graph}, or a
\textbf{mixture}. The answer can be a supervision target rather than an input, so it cannot be removed through a fixed-weight intervention; appendix Table~\ref{tab:taxonomy} maps the information retained by the remaining conditions.

We complement the interventions with a post-hoc probe that maps stored tokens
to frozen embeddings of the question, answer, or verbalised graph. The map is
fitted on half of the test examples; on the held-out half, we report the median
rank of each example's correct target among all candidates. A constant token
provides the control rank. Estimation details and limitations are given in
Appendix~\ref{app:protocol}.

\textbf{Graph-free controls.} \textbf{CONST} passes a learned graph-independent vector through G-Retriever's projector, while \textbf{MLPQ} replaces the graph encoder with an MLP over the question. \textbf{PT} is the raw prompt-tuning baseline, the same as that used by \citet{he2024gretriever}, and does not use the projector. These controls distinguish the contribution of graph input from that of token parameterisation.

\subsection{Preventing collapse: a two-stage pipeline}
\label{sec:pipeline}

\begin{figure}[t]
\centering
\includegraphics[width=\linewidth]{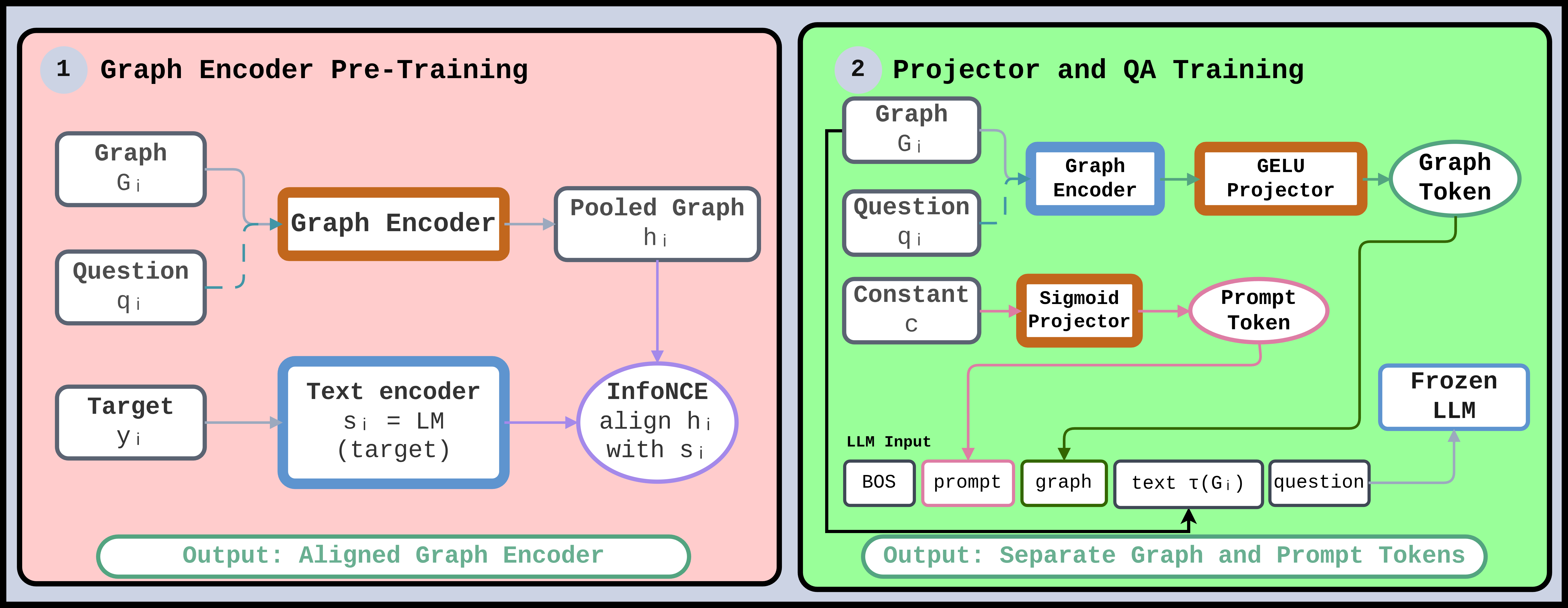}
\caption{\textbf{Two-stage pipeline.} Stage~1 aligns the question-conditioned graph encoder. Stage~2 freezes it and trains separate graph-token and prompt-token pathways. Orange modules are trained; blue modules are frozen.}
\label{fig:pipeline}
\end{figure}

Our ablations associate collapse with two factors: a Sigmoid projector that suppresses encoder variation and end-to-end answer training that does not require this variation to be preserved. We address both while retaining G-Retriever's data, prompt, answer loss, and optimisation setup (Figure~\ref{fig:pipeline}).

\textbf{Stage 1: preserve input-specific variation.}
Without the LLM, we train the graph encoder with an InfoNCE objective~\citep{oord2019representationlearningcontrastivepredictive}. Its pooled representation $\mathbf{h}_i$ is aligned with a frozen embedding of a sample-specific target and contrasted against negative targets. The question conditions each node through
$\mathbf{x}_v=\mathbf{z}_v+\mathbf{W}_q\mathbf{q}_i$, with $\mathbf{W}_q$ initialised at zero. We use the answer as the target, or the verbalized graph, when the answer space is too small to distinguish samples  (e.g., answers are Yes/No). The encoder is then frozen.

\textbf{Stage 2: separate graph representation from prompting.}
Using the original answer loss, we jointly train a GELU projector over the frozen encoder and a separate graph-independent prompt token. Freezing protects the variation learned in Stage~1, while the second pathway gives the shared prompting function its own token rather than forcing it to compete with input-specific information.

The full objective, targets, and design ablations are reported in
Appendix~\ref{app:design}.
\section{Experiments}
\label{sec:experiments}

We investigate four questions. \textbf{RQ1:} Do graph tokens collapse, and how general is this behaviour? \textbf{RQ2:} What role does the collapsed token play when the same graph is also verbalised in the prompt? \textbf{RQ3:} What is collapse associated with, and can it be prevented without sacrificing task performance? \textbf{RQ4:} What do non-collapsed tokens encode, and which of that information does the LLM use?
\subsection{Setup}
\label{sec:setup}
\textbf{Benchmarks.}
We use the three GraphQA benchmarks introduced by~\citet{he2024gretriever}: \texttt{ExplaGraphs}, \texttt{SceneGraphs}, and \texttt{WebQSP}. Following their official splits, retrieval procedures, and evaluators, we report accuracy on \texttt{ExplaGraphs} and \texttt{SceneGraphs} and Hit@1 on \texttt{WebQSP}. Dataset statistics and retrieval settings are in Appendix~\ref{app:data}.

\textbf{Models.}
We reproduce G-Retriever with its default graph transformer (GT) and frozen Llama-2-7B \citep{touvron2023llama2openfoundation}. Robustness experiments use GAT, GCN, Qwen2.5-3B \citep{qwen2025}, and LoRA adaptation. Architecture, optimisation, and pipeline settings are in Appendix~\ref{app:impl}.

\textbf{Evaluation stability.}
Re-evaluating identical \texttt{WebQSP} weights in a separate process changes Hit@1 by $1.04$ points. We therefore require an answer-level difference larger than $2$ points with $p<.05$. Gold-answer likelihood is more stable; we require $|\Delta\mathrm{NLL}|>0.03$ and a bootstrap interval excluding the replicate floor. Calibration results are in Appendix~\ref{app:protocol}.

\textbf{Statistical testing.}
Accuracy and Hit@1 comparisons are paired with the exact McNemar test~\citep{McNemar_1947}. We denote $p<.10$, $.05$, $.01$, and $.001$ by \sig{\dag}, \sig{*}, \sig{**}, and \sig{***}, respectively; unmarked comparisons have $p\geq.10$. Reported $p$-values are unadjusted, so claims rely on the effect-size criteria above and consistent evidence rather than isolated significant cells.

\subsection{RQ1: Do graph tokens collapse, and how general is collapse?}
\label{sec:rq1}

\begin{table}[t]
\caption{\textbf{Collapse across datasets, graph encoders, LLMs, and tuning regimes.} Scores are accuracy for \texttt{ExplaGraphs} and \texttt{SceneGraphs} and Hit@1 for \texttt{WebQSP}.}
\label{tab:collapse}
\begin{center}
\small
\fitwidth{
\begin{tabular}{llccc}
\toprule
Dataset & Model & Score & Encoder cosine & Token cosine \\
\midrule
\multirow{3}{*}{\texttt{ExplaGraphs}}
 & G-Retriever (GT)  & 85.56 & 0.678 & 0.997 \\
 & G-Retriever (GAT) & 85.20 & 0.457 & 0.998 \\
 & G-Retriever (GCN) & 78.88 & 0.423 & 0.985 \\
\midrule
\multirow{3}{*}{\texttt{SceneGraphs}}
 & G-Retriever (GT)  & 81.35 & 0.630 & 0.993 \\
 & G-Retriever (GAT) & 80.20 & 0.693 & 0.998 \\
 & G-Retriever (GCN) & 80.55 & 0.428 & 0.998 \\
\midrule
\multirow{3}{*}{\texttt{WebQSP}}
 & G-Retriever (GT)   & 70.33 & 0.593 & 0.996 \\
 & \quad + LoRA       & 74.32 & 0.621 & 0.997 \\
 & \quad + Qwen2.5-3B & 55.47 & 0.589 & 0.997 \\
\bottomrule
\end{tabular}}
\end{center}
\end{table}

\textbf{Collapse occurs in every tested configuration.}
Token cosine ranges from $0.985$ to $0.998$, while the input-specific residual is only 4--12\% of the shared mean, whose norm is 22--97 times that of an ordinary word embedding (Appendix~\ref{app:collapse-full}). Figure~\ref{fig:tokenspace} illustrates this geometry on \texttt{WebQSP}. Models with similarly collapsed tokens nevertheless obtain substantially different scores, so task performance does not reveal collapse.

The encoder representations are considerably more variable, with cosines between $0.423$ and $0.693$. This makes GNN oversmoothing \citep{rusch2023surveyoversmoothinggraphneural} an insufficient explanation for token collapse:
even if node representations are smoothed within a graph, the pooled outputs
might remain distinct across inputs, and near-identity appears only after projection.
Collapse also persists across three graph encoders, two LLM families, and frozen and LoRA-adapted LLMs. Moreover, a controlled \texttt{WebQSP} audit finds the same geometry in later single- and multi-token graph-prompting methods (Appendix~\ref{app:successors}). This answers \textbf{RQ1}: collapse is a recurring failure mode rather than a peculiarity of one encoder, LLM, or
single-token implementation. We next ask what the collapsed token contributes to the answer.
\subsection{RQ2: What role does the collapsed token play?}
\label{sec:rq2}

\begin{table}[t]
\caption{\textbf{Role of G-Retriever's collapsed token.}
\textsc{real}, \textsc{notxt}, and \textsc{const} are absolute scores; other
columns report differences from their corresponding reference. Full results in Appendices~\ref{app:interventions-full},
\ref{app:magnitude}, and~\ref{app:controls-full}.}
\label{tab:collapsed-role}
\begin{center}
\small
\fitwidth{
\begin{tabular}{lccccccc}
\toprule
Dataset & \textsc{real} & $\Delta$\textsc{mean} & $\Delta$\textsc{swap}
& $\Delta$\textsc{gswap} & \textsc{notxt}
& $\Delta$\textsc{notxt-mean} & \textsc{const} \\
\midrule
\texttt{ExplaGraphs}
& 85.56 & $-0.18$ & $-1.26$ & $-1.08$
& 66.43 & $-0.90$ & 85.56 \\
\texttt{SceneGraphs}
& 81.35 & $-0.35$ & $-1.20$\sig{\dag} & $-1.10$\sig{\dag}
& 30.20 & $+0.45$ & 81.40 \\
\texttt{WebQSP}
& 70.33 & $-1.84$\sig{*} & $-0.92$ & $-0.55$
& 41.34 & $+1.04$ & 71.25 \\
\bottomrule
\end{tabular}}
\end{center}
\end{table}
\textbf{The input-specific component has no resolved answer-level effect.}
Across all three benchmarks, \textsc{mean}, \textsc{swap}, and \textsc{gswap} remain within the two-point threshold (Table~\ref{tab:collapsed-role}); corrupting entities, edges, or relations gives the same result (more evidence in Appendix~\ref{app:interventions-full}). Yet replacing the token with zero costs 59.64--81.95 points (see Appendix~\ref{app:magnitude}). The model therefore relies on the token's shared value, but not measurably on its graph-specific variation.

\textbf{The verbalised graph is the dominant graph channel.}
Removing it costs 19--51 points, whereas replacing the remaining token with its mean changes performance by at most 1.04 points. Thus, in the default Llama-2 configuration, task-relevant graph information is supplied primarily through text.

\textbf{The shared token acts as a soft prompt.}
\textsc{const}, a graph-independent vector passed through the same projector, matches or exceeds G-Retriever on every benchmark. Graph input is therefore unnecessary for the benefit attributed to the collapsed token. Raw prompt tuning, namely the existing ablation performed by \citet{he2024gretriever}, performs worse (Appendix~\ref{app:controls-full}), but changes the parameterisation and is not a controlled graph-versus-no-graph comparison.

Small distributional effects remain: on \texttt{WebQSP}, \textsc{swap} and \textsc{gswap} increase gold-answer NLL by $0.0183$ and $0.0182$, both below the $0.03$ criterion (Appendix~\ref{app:nll}). These results answer \textbf{RQ2}: the collapsed token functions primarily as a
projector-parameterised soft prompt, while the verbalised graph is the dominant graph-information channel. \textbf{RQ3} asks what collapse is associated with and whether it can be prevented.

\subsection{RQ3: What is collapse associated with, and can it be prevented?}
\label{sec:rq3}
\label{sec:mechanism}

\begin{table}[t]
\caption{\textbf{Factors associated with collapse on \texttt{ExplaGraphs}.} Step~0 precedes projector training, and significance markers compare task score with the first row. Full projector diagnostics and intervention results are in Appendices~\ref{app:mechanism} and~\ref{app:interventions-full}.}
\label{tab:mechanism}
\begin{center}
\small
\fitwidth{
\begin{tabular}{llcccc}
\toprule
Encoder & Projector & Encoder cos. & Step-0 token cos. & Token cos. & Score \\
\midrule
trained end to end   & Sigmoid & 0.695 & 1.000 & 0.997 & 85.56 \\
trained end to end   & GELU    & 0.787 & 0.992 & 0.914 & 73.29\sig{***} \\
frozen after Stage~1 & Sigmoid & 0.024 & 0.987 & 0.992 & 84.48 \\
frozen after Stage~1 & GELU    & 0.024 & 0.091 & 0.323 & 57.76\sig{***} \\
\bottomrule
\end{tabular}}
\end{center}
\end{table}

\textbf{Factors associated with collapse.}
Table~\ref{tab:mechanism} identifies two associations. With the aligned Stage~1 encoder frozen, its representations are highly distinct (cosine $0.024$), yet the Sigmoid projector produces nearly identical tokens before and after training ($0.9870$ and $0.9918$). GELU instead yields cosine $0.0910$ before training and $0.3229$ afterwards. Untrained-projector experiments reproduce this contrast on random inputs (Appendix~\ref{app:mechanism}), linking the Sigmoid parameterisation to suppression of encoder variation.

Changing the projector alone is insufficient: under end-to-end answer training, GELU still produces highly similar tokens (cosine $0.9139$). Conversely, combining the aligned frozen encoder with GELU prevents collapse, but the single-token model falls to approximately majority-class performance ($57.76$ versus $58.8$). Thus, both factors are associated with collapse, but this experiment does not identify either as the sole cause. The cross-method audit supports this qualification (Appendix~\ref{app:successors}). GRAG \citep{hu2024grag} and GRETPool~\citep{grover2026one} inherit the same Sigmoid projector and also collapse, while GraSP~\citep{wang2026topology} collapses despite using the hyperbolic tangent. 

\begin{table}[t]
\caption{\textbf{Token geometry, performance, and intervention effects.}
\emph{Ours} denotes the complete two-token pipeline; Intervention columns report condition minus \textsc{real}. Bold scores mark the highest score within each dataset--LLM block; bold \textsc{gswap} effects satisfy $|\Delta|>2$ and $p<.05$. Complete interventions and $p$-values are in Appendix~\ref{app:interventions-full}, and pipeline ablations are in Appendix~\ref{app:pipeline-ablations}.}
\label{tab:geometry-use}
\begin{center}
\small
\fitwidth{
\begin{tabular}{llccccc}
\toprule
Dataset & Model & Score & Token cos. &
$\Delta$\textsc{mean} & $\Delta$\textsc{swap} & $\Delta$\textsc{gswap} \\
\midrule
\multicolumn{7}{l}{\emph{Llama-2-7B}} \\
\cmidrule(lr){1-7}
\multirow{4}{*}{\texttt{WebQSP}}
 & G-Retriever              & 70.33 & 0.996 & $-1.84$\sig{*}   & $-0.92$            & $-0.55$ \\
 & \textbf{Ours}            & 69.90 & 0.142 & $-3.13$\sig{***} & $-2.15$\sig{*}     & $-1.97$\sig{\dag} \\
 & \quad no entity features & $\mathbf{72.73}$ & 0.125 & $-6.14$\sig{***} & $-5.65$\sig{***}   & $-1.97$\sig{*} \\
 & \quad no question        & 69.84 & 0.169 & $-0.06$           & $-3.69$\sig{***}   & $\mathbf{-2.33}$\sig{*} \\

\midrule
\multirow{4}{*}{\texttt{SceneGraphs}}
 & G-Retriever              & $\mathbf{81.35}$ & 0.993 & $-0.35$           & $-1.20$\sig{\dag} & $-1.10$\sig{\dag} \\
 & \textbf{Ours}            & 72.50 & 0.133 & $-1.40$           & $-3.40$\sig{***}  & $\mathbf{-5.30}$\sig{***} \\
 & \quad no entity features & 79.65 & 0.125 & $-5.75$\sig{***} & $-9.60$\sig{***}  & $+0.35$ \\
  & \quad no question        & 76.35 & 0.225 & $-1.15$ & $-0.25$ & $-0.25$ \\

\midrule
\multirow{4}{*}{\texttt{ExplaGraphs}}
 & G-Retriever              & 85.56 & 0.997 & $-0.18$           & $-1.26$ & $-1.08$ \\
 & \textbf{Ours}            & 85.74 & 0.099 & $+0.36$           & $-0.54$ & $-1.44$ \\
 & \quad no entity features & 84.48 & 0.113 & $+3.07$\sig{**}  & $0.00$  & $-0.36$ \\
 & \quad no question
 & $\mathbf{87.55}$ & 0.108 & $+1.08$ & $+0.90$ & $+1.26$ \\
\midrule
\multicolumn{7}{l}{\emph{Qwen2.5-3B, matched training budget}} \\
\cmidrule(lr){1-7}
\multirow{2}{*}{\texttt{WebQSP}}
 & G-Retriever & 57.13 & 0.997 & $-0.55$ & $-1.41$ & $-1.60$\sig{\dag} \\
 & \textbf{Ours} & $\mathbf{65.79}$ & 0.200
 & $-2.09$\sig{**} & $-1.90$\sig{**}
 & $\mathbf{-2.09}$\sig{**} \\
\midrule
\multirow{2}{*}{\texttt{SceneGraphs}}
 & G-Retriever & $\mathbf{76.05}$ & 0.997
 & $-0.30$ & $-1.05$\sig{\dag} & $-1.15$\sig{*} \\
 & \textbf{Ours} & 72.65 & 0.139
 & $-7.00$\sig{***} & $-5.75$\sig{***}
 & $\mathbf{-6.50}$\sig{***} \\
\midrule
\multirow{2}{*}{\texttt{ExplaGraphs}}
 & G-Retriever & 74.37 & 0.998
 & $-1.62$ & $-2.17$ & $\mathbf{-4.51}$\sig{*} \\
 & \textbf{Ours} & $\mathbf{83.94}$ & 0.103
 & $+0.90$ & $0.00$ & $+0.36$ \\

\bottomrule
\end{tabular}}
\end{center}
\end{table}

\textbf{Preventing collapse while retaining performance.}
Table~\ref{tab:geometry-use} shows that the complete two-token pipeline mitigates collapse in every tested setting. For the standard pipeline, token cosine falls from above $0.99$ to $0.099$--$0.142$ with Llama-2 and to
$0.103$--$0.200$ with Qwen2.5-3B.

With Llama-2, the pipeline retains performance on \texttt{WebQSP} ($69.90$ versus $70.33$) and \texttt{ExplaGraphs} ($85.74$ versus $85.56$), with both differences below our resolution threshold. On \texttt{ExplaGraphs}, the separate prompt pathway recovers the performance lost by the non-collapsed single-token model, raising its score from $57.76$ to $85.74$. On \texttt{SceneGraphs}, however, \textsc{gswap} produces a resolved $-5.30$-point effect while the score falls from $81.35$ to $72.50$.

The Qwen2.5-3B replication further separates performance from graph use. The Qwen rows in Table~\ref{tab:geometry-use} use a matched training budget and are distinct from the robustness checkpoint in Table~\ref{tab:collapse}. On \texttt{WebQSP}, our pipeline improves the matched baseline from $57.13$ to $65.79$ ($p=4.5\times10^{-20}$), while \textsc{gswap} produces a resolved $-2.09$-point effect ($p=.0027$). On \texttt{ExplaGraphs}, it improves from $74.37$ to $83.94$ ($p=1.5\times10^{-5}$), but \textsc{gswap} has no effect ($+0.36$). Conversely, on \texttt{SceneGraphs}, it produces the strongest graph-replacement effect in the study ($-6.50$, $p<.001$) while scoring $3.40$ points below G-Retriever. Qwen G-Retriever on \texttt{ExplaGraphs} provides the opposite exception: its tokens remain collapsed, yet \textsc{gswap} causes a resolved $-4.51$-point effect. Near-collapse therefore does not imply complete insensitivity to the remaining variation. 

These results answer \textbf{RQ3}. We found that collapse is associated with both the projector and end-to-end answer training, without either being isolated as its sole cause. The two-stage pipeline mitigates collapse across both LLM families and can retain or improve performance. However, geometric collapse, graph-specific token use, and task performance remain distinct properties. \textbf{RQ4} takes a step forward and examines what the tokens encode and which information the LLM uses.
\subsection{RQ4: What does a non-collapsed token carry?}
\label{sec:rq4}

Non-collapse shows that tokens differ across inputs, but not what they represent or whether the LLM uses those differences. Because Stage~1 receives the question and an answer-related target, the token may encode the question,
an answer prior, the graph, or a mixture.

We separate recoverable content from downstream use.
Table~\ref{tab:decomposition} reports question, answer, and graph retrieval ranks, answer-level effects under \textsc{gswap}, and likelihood changes under \textsc{swap} and \textsc{gswap}. Table~\ref{tab:geometry-use} measures reliance on sample-specific token information through \textsc{mean} and \textsc{swap}, while the appendix Table~\ref{tab:notext-pipeline} repeats the interventions without the retrieved verbalised graph. Complete cross-LLM results are reported in Appendix~\ref{app:decodability}.

The retrieval probe maps each token to the corresponding question, answer, or verbalised graph among held-out candidates. We report the median rank of the correct target: lower is better, and a constant token provides the control rank. Recoverability does not imply use. A low answer rank, for example, means that the answer is predictable from the token, not that it was computed from
the graph or used by the LLM.

We use five diagnoses. \textsc{Used} means that \textsc{gswap} changes the task score by more than two points with $p<.05$. \textsc{Masked} means that
this effect resolves only after removing the verbalised graph.
\textsc{Prior} identifies a no-entity model whose token carries question and answer information and is used under \textsc{mean} or \textsc{swap}, without resolved graph-specific use. \textsc{Decodable} means that graph information is recoverable without a resolved answer-level effect. \textsc{Soft prompt} denotes a collapsed baseline without resolved graph-specific use; any information recoverable from its residual variation is reported separately by the retrieval ranks.

\begin{table}[t]
\caption{\textbf{Token content and use with Llama-2.}
Q, A, and G are median retrieval ranks (lower is better); constant controls are
407 for \texttt{WebQSP}, 500 for \texttt{SceneGraphs} (A: 476--500), and 139
for \texttt{ExplaGraphs} (A omitted). $\Delta$Score is the with-text
\textsc{gswap} effect; NLL reports \textsc{swap}/\textsc{gswap}. Bold denotes
effects meeting the criteria in Section~\ref{sec:setup}.}
\label{tab:decomposition}
\begin{center}
\small
\fitwidth{
\begin{tabular}{llcccccl}
\toprule
Dataset & Model & Q rank & A rank & G rank
& $\Delta$Score \textsc{gswap}
& $\Delta$NLL S/GS & Diagnosis \\
\midrule

\multirow{4}{*}{\texttt{WebQSP}}
 & G-Retriever
 & 293 & 317 & 369
 & $-0.55$
 & $+0.0183/+0.0182$
 & \textsc{soft prompt} \\
 & \textbf{Ours}
 & 65 & 60 & 315
 & $-1.97$\sig{\dag}
 & $\mathbf{+0.0895}/\mathbf{+0.0332}$
 & \textsc{masked} \\
 & \quad no entity features
 & 34 & 51 & 303
 & $-1.97$\sig{*}
 & $\mathbf{+0.1111}/+0.0262$
 & \textsc{prior} \\
 & \quad no question
 & 120 & 134 & 329
 & $\mathbf{-2.33}$\sig{*}
 & $\mathbf{+0.0848}/\mathbf{+0.0845}$
 & \textsc{used} \\

\midrule
\multirow{4}{*}{\texttt{SceneGraphs}}
 & G-Retriever
 & 271 & 436 & 248
 & $-1.10$\sig{\dag}
 & $+0.0169/+0.0167$
 & \textsc{soft prompt} \\
 & \textbf{Ours}
 & 47 & 187 & 427
 & $\mathbf{-5.30}$\sig{***}
 & $+0.0229/\mathbf{+0.1665}$
 & \textsc{used} \\
 & \quad no entity features
 & 8 & 59 & 369
 & $+0.35$
 & $\mathbf{+0.1553}/+0.0045$
 & \textsc{prior} \\
 & \quad no question
 & 116 & 395 & 185
 & $-0.25$
 & $+0.0214/+0.0213$
 & \textsc{masked} \\

\midrule
\multirow{4}{*}{\texttt{ExplaGraphs}}
 & G-Retriever
 & 138 & -- & 139
 & $-1.08$
 & $+0.0014/+0.0022$
 & \textsc{soft prompt} \\
 & \textbf{Ours}
 & 54 & -- & 41
 & $-1.44$
 & $+0.0010/+0.0032$
 & \textsc{decodable} \\
 & \quad no entity features
 & 67 & -- & 67
 & $-0.36$
 & $+0.0026/+0.0056$
 & \textsc{decodable} \\
 & \quad no question
 & 84 & -- & 70
 & $+1.26$
 & $-0.0063/-0.0073$
 & \textsc{decodable} \\

\bottomrule
\end{tabular}}
\end{center}
\end{table}

\textbf{\texttt{WebQSP}: the token's role depends on training.}
The standard token makes the question and answer strongly recoverable, at ranks $65$ and $60$ against the control of $407$. With the verbalised graph present, \textsc{gswap} affects likelihood but not the decoded score; after removing the text, it causes a resolved $-2.27$-point effect (Tables~\ref{tab:decomposition} and~\ref{tab:notext-pipeline}). Removing entity features further strengthens question and answer recovery and improves Hit@1 without resolved graph use, supporting the \textsc{prior} diagnosis. Conversely, removing question conditioning produces resolved graph use. Thus, the same architecture can favour priors or graph information depending on its training inputs.

\textbf{\texttt{SceneGraphs}: graph use and accuracy diverge.}
Our pipeline uses graph-specific token content: \textsc{gswap} lowers accuracy by $5.30$ points and increases NLL by $0.1665$, but it remains less accurate
than G-Retriever. G-Retriever's collapsed token contains recoverable graph information, yet shows no resolved graph use, so its functional diagnosis remains \textsc{soft prompt}. The no-entity model nearly recovers its accuracy
while eliminating graph use, confirming that performance and token use are distinct.

\textbf{\texttt{ExplaGraphs}: graph information is decodable but unused.}
Our token makes the graph substantially more recoverable, at rank $41$ against the control of $139$, but \textsc{gswap} has no resolved effect with or without
the verbalised graph. Because these graphs are small and supplied in full, their textual representation may already be sufficient. However, removing that text does not make the model rely on the token. The evidence therefore establishes recoverable graph information, not downstream use.

Together, these results answer \textbf{RQ4}: preventing collapse still does not uniquely determine what a token represents or whether the LLM uses it. Depending on the benchmark and training inputs, a non-collapsed token may encode a question or answer prior, unused graph information, or graph information that changes the answer.

\section{Discussion}
\label{sec:discussion}

\textbf{What standard evaluation misses.}
Task accuracy cannot isolate graph-token use, while raw prompt tuning changes the token parameterisation and no-text retraining changes the learning problem. A minimal audit should combine token geometry, a graph-free control using the same projector, and fixed-weight \textsc{mean}, \textsc{swap}, and \textsc{gswap} interventions. The controlled \texttt{WebQSP} audit in Appendix~\ref{app:successors} finds collapse with query-conditioned and multi-token approaches, extending the phenomenon beyond G-Retriever up to most recent graph soft prompting methods.

\textbf{Graph use and accuracy are different outcomes.}
Our pipeline can make answers depend on graph-specific token variation without improving accuracy. The textual channel may be easier to exploit, while continuous graph tokens may be lossy or difficult for the LLM to interpret.
The Qwen2.5-3B results show that this trade-off is not inevitable; alignment claims should therefore report graph use and performance separately.

\textbf{Limitations and future directions.}
Our detailed analysis covers G-Retriever on three question-answering benchmarks; the cross-method audit is limited to \texttt{WebQSP}, runnable implementations, and controlled variants using G-Retriever's retrieved subgraphs. The mechanism evidence is associative and does not isolate the projector or end-to-end training as a sole cause.

The audited recipe trains from scratch with answer cross-entropy; collapse may differ with pretrained encoders or explicit contrastive, reconstructive, or counterfactual objectives~\citep{GTLLM,xu2025aligngrag}. Stage~1 explores one alternative, but its conditioning and targets can introduce priors. \textsc{Gswap} tests graph--question pairing rather than fact-level reasoning, and decodability measures recoverability rather than use. Our pipeline is
therefore a diagnostic proof of concept. Future work should preserve task-relevant graph information and verify that the LLM uses it.

\section{Conclusion}

We uncover token collapse in G-Retriever across benchmarks, graph encoders, LLM families, and tuning regimes, and find the same geometry in controlled tests of later single- and multi-token interfaces. Despite competitive task
scores, the projected tokens are nearly identical across inputs.
We also find that graph-free controls and fixed-weight interventions match the performance of the original G-retriever model. These findings show that strong task performance can conceal a graph-token interface whose principal contribution is a graph-independent soft prompt, while the verbalised graph provides the dominant graph-information channel.
Our two-stage pipeline mitigates this collapse, but also demonstrates that distinct tokens do not guarantee graph use or improved accuracy. Token variation, recoverable graph information, downstream use, and task performance must therefore be evaluated separately. Our protocol provides a practical way to distinguish these outcomes and turns alignment from an architectural assumption into an empirical claim that will foster the development of future graph--language models.

\subsection*{AI use statement}

Generative AI tools assisted with prose revision and parts of the analysis and plotting code. They did not generate experimental results: all newly reported experimental numbers were produced by our pipeline and checked against stored predictions, tokens, checkpoints, or logs. The authors reviewed all AI-assisted text and code, verified the experimental claims, and take full responsibility for the paper.

\subsection*{Reproducibility statement}

Appendix~\ref{app:impl} specifies the repository and code modification, optimisation settings, model dimensions, LoRA configuration, Stage-1 hyperparameters, trainable parameter counts, hardware, and evaluation procedure. Appendix~\ref{app:data} reports dataset statistics, splits,
retrieval settings, metrics, and intervention subsets.
Appendix~\ref{app:interventions-full} provides the complete fixed-weight intervention results; the remaining appendices report the geometry, likelihood, decodability, controls, and pipeline measurements.
\bibliography{main}
\bibliographystyle{iclr2027_conference}

\appendix

\section*{Appendix Overview}

\begin{itemize}\setlength\itemsep{1pt}
\item[] \emph{Setup.} \ref{app:impl}~implementation and hyperparameters; \ref{app:data}~datasets and retrieval; \ref{app:encoders}~graph encoders.
\item[] \emph{Protocol.} \ref{app:protocol}~condition taxonomy, likelihood and decodability estimators, evaluation stability and the statistical protocol.
\item[] \emph{Collapse.} \ref{app:collapse-full}~complete collapse statistics and per-encoder interventions; \ref{app:successors}~complete collapse statistics for methods subsequent to G-Retriever; \ref{app:tokenspace}~token space of every model; \ref{app:mechanism}~projector diagnostics behind Section~\ref{sec:mechanism}.
\item[] \emph{Causal audit.} \ref{app:interventions-full}~full fixed-weight interventions; \ref{app:magnitude}~magnitude and geometric conditions; \ref{app:controls-full}~complete graph-free controls; \ref{app:nll}~gold-answer likelihood.
\item[] \emph{Content and pipeline.} \ref{app:decodability}~complete decodability results; \ref{app:design}~design choices and Stage-1 diagnostics; \ref{app:pipeline-ablations}~pipeline and robustness ablations; \ref{app:xfit}~cross-fitting; \ref{app:seeds}~initialisations and evaluation noise.
\item[] \emph{Context.} \ref{app:motivating}~the verbalised graph is not enough; \ref{app:related}~extended related work.
\end{itemize}

\section{Implementation details}
\label{app:impl}

\textbf{G-Retriever reproduction.}
We use the authors' public implementation at a pinned commit. For the baseline reproduction, our only modification is explicit device placement, allowing each run to fit on a single GPU. The frozen sentence encoder is all-roberta-large-v1 with output width $d=1024$. The default graph encoder is a four-layer, four-head graph transformer with pooled width $d_g=1024$. Its representation is mapped to the LLM embedding space by a two-layer projector with hidden width $d_p=2048$. The default LLM is frozen Llama-2-7B, whose embedding width is $d_\ell=4096$. The verbalised graph is truncated to $L_{\max}=512$ tokens, and generation is limited to 32 tokens.

Optimisation follows the authors' defaults: AdamW with learning rate $10^{-5}$, weight decay $0.05$, and at most 10 epochs, with early stopping after two epochs without validation improvement. Every model in this paper is trained on a single 80\,GB GPU.

\textbf{Stage 1.} InfoNCE with temperature 0.07 against a memory bank of about 2,000 targets, with no LLM loaded. Trainable: 20,994,048 parameters in the encoder and 1,049,600 in the question-conditioning layer $\mathbf{W}_q$.

\textbf{Stage 2.}
The Stage-1 encoder and $\mathbf{W}_q$ parameters are frozen, while their running normalisation statistics remain in training mode. The projector's Sigmoid is replaced by GELU, and a learned constant of width $d_g$ passes through its own Linear $\rightarrow$ Sigmoid $\rightarrow$ Linear projector to produce the prompt token. The two projector pathways are initialised and optimised jointly; checkpoint-derived trainable parameter counts are reported in Table~\ref{tab:controls-full}. Interventions modify only the graph token, never the prompt token.

\textbf{Controls and parameter counts.}
Table~\ref{tab:controls-full} reports checkpoint-derived trainable parameter counts for the evaluated models. Stage~1 separately trains $22{,}043{,}648$ parameters; because the two stages are trained sequentially, we do not report their sum as a simultaneous trainable count. Accuracy does not track capacity: CONST improves over PT by 19 points, while our larger Stage-2 model scores 1.35 points below CONST. Since PT and CONST differ in both parameterisation and capacity, their gap should not be attributed to the projector alone. The controlled graph-input comparison is instead between CONST and G-Retriever, which share the projector parameterisation.

\section{Datasets}
\label{app:data}

We use the GraphQA benchmark assembled by \citet{he2024gretriever}, whose statistics are reproduced in Table~\ref{tab:datasets}. Retrieval is needed on two of the three: verbalising an average \texttt{WebQSP} graph takes 100,627 tokens, reduced to 610 after retrieval, and an average \texttt{SceneGraphs} graph 1,396 tokens, reduced to 235. \texttt{ExplaGraphs} graphs fit in the prompt, so the whole graph is used. Retrieval keeps the top 3 nodes and edges on \texttt{SceneGraphs} with edge cost 1, and the top 3 nodes and 5 edges on \texttt{WebQSP} with edge cost 0.5.

\begin{table}[h]
\caption{GraphQA, textual graphs benchmark. Numbers are from  \citet{he2024gretriever}.}
\label{tab:datasets}
\begin{center}
\small
\fitwidth{
\begin{tabular}{lccc}
\toprule
 & \texttt{ExplaGraphs} & \texttt{SceneGraphs} & \texttt{WebQSP} \\
\midrule
Graphs & 2,766 & 100,000 & 4,737 \\
Average nodes & 5.17 & 19.13 & 1,370.89 \\
Average edges & 4.25 & 68.44 & 4,252.37 \\
Node attributes & commonsense concepts & objects and attributes & Freebase entities \\
Edge attributes & commonsense relations & actions, spatial relations & Freebase relations \\
Task & stance prediction & scene-graph QA & knowledge-graph QA \\
Metric & accuracy & accuracy & Hit@1 \\
Test questions & 554 & 20,025 & 1,628 \\
\bottomrule
\end{tabular}}
\end{center}
\end{table}

Metrics use the benchmark's own evaluators, applied to the stored per-question predictions: on \texttt{ExplaGraphs} the first ``support'' or ``counter'' in the output must equal the label; on \texttt{SceneGraphs} the gold label must appear in the output; on \texttt{WebQSP} a prediction counts as a hit when it contains one of the gold answers. Interventions use all test questions on \texttt{ExplaGraphs} and \texttt{WebQSP}, and the first 2,000 on \texttt{SceneGraphs}, of which 1,302 are not yes/no questions. The one comparison drawn on the full \texttt{SceneGraphs} test set, prompt tuning against the graph-free constant in Section~\ref{sec:rq2}, is stated as such where it appears.

\section{Graph encoders}
\label{app:encoders}

GAT~\citep{velickovic2018gat} and GCN~\citep{kipf2017gcn} replace the graph transformer~\citep{shi2020graphtransformer} inside G-Retriever's pipeline, using the repository's own implementations with the same depth and hidden width. Their collapse statistics and interventions appear in Tables~\ref{tab:collapse} and~\ref{tab:collapsed-role}, their token spaces in Figures~\ref{fig:tokenspace-expla} and~\ref{fig:tokenspace-scene}, and every condition measured for them in Appendix~\ref{app:interventions-full}. The three encoders differ in accuracy on \texttt{ExplaGraphs}, where GCN loses 6.7 points to the graph transformer, but not in the behaviour of their tokens.

\section{Protocol details}
\label{app:protocol}

\textbf{Token-geometry statistics.}
Let $\mathbf{t}_1,\ldots,\mathbf{t}_n$ be the test tokens,
$\boldsymbol{\mu}=\frac{1}{n}\sum_i\mathbf{t}_i$ their mean, and $\bar e$ the
median norm of the LLM's word embeddings. We measure
\begin{equation}
\begin{gathered}
\operatorname{cos}_{\mathrm{inter}}
=
\frac{2}{n(n-1)}
\sum_{i<j}
\frac{\mathbf{t}_i^\top\mathbf{t}_j}
     {\lVert\mathbf{t}_i\rVert\lVert\mathbf{t}_j\rVert},
\qquad
\operatorname{RR}
=
\frac{1}{n}\sum_i
\frac{\lVert\mathbf{t}_i-\boldsymbol{\mu}\rVert}
     {\lVert\boldsymbol{\mu}\rVert},
\\
\operatorname{NR}
=
\frac{\frac{1}{n}\sum_i\lVert\mathbf{t}_i\rVert}{\bar e}.
\end{gathered}
\label{eq:stats}
\end{equation}
Inter-sample cosine measures whether tokens point in the same direction,
the residual ratio (RR) measures their input-specific component relative to
the shared mean, and the norm ratio (NR) compares their magnitude with an
ordinary word embedding. We compute the same inter-sample cosine for the
pooled encoder representations $\mathbf{h}_i$ to distinguish collapse in the
encoder from collapse introduced by the projector.

\textbf{GSWAP.} The partner graph is drawn by a fixed derangement of the test set, a cyclic shift, and two further random derangements are reported as replicates in Appendix~\ref{app:interventions-full}. The partner's graph replaces the batch's graph before the encoder, while the question conditioning $\mathbf{W}_q\mathbf{q}_i$ still uses question $i$, which is what makes the condition specific to the graph.

\textbf{The conditions.} \textsc{mean} replaces the token by the mean token over the training graphs, identical for every question. \textsc{swap} replaces it by the token of another test question under a fixed permutation. \textsc{node} recomputes the token after replacing node features with real rows from other graphs, keeping the edges; \textsc{edge} after rewiring every edge uniformly at random inside its own graph, keeping features and edge count; \textsc{rel} after resampling relation attributes. The conditions on the token's geometry are the residual scaling $\boldsymbol{\mu}+a(\mathbf{t}_i-\boldsymbol{\mu})$ with $a\in\{0,2\}$, the zero token $\mathbf{0}$, and rescaling the token to a chosen norm. \textsc{notxt-x} evaluates condition x with the verbalised graph deleted from the prompt, which puts every model out of its training distribution and is read as a comparison between models on the same items.

\textbf{The measured floor.} Table~\ref{tab:floor} is the calibration behind the thresholds of Appendix~\ref{app:protocol}. The null replicate reruns the identical condition on the identical weights in a second process and compares the generated strings; models with a GNN are not bit-reproducible across processes, because scatter reductions accumulate in a nondeterministic order, and a difference of order $10^{-7}$ in the prefix token rewrites most greedy decodings. MLPQ's tokens are identical to within $3\times10^{-4}$, so its \textsc{swap} carries no information and is still significant.

\begin{table}[h]
\caption{\textbf{The measured floor}, \texttt{WebQSP}, $n=1{,}628$. Churn is the fraction of generated strings that change; no perturbation here carries graph information.}
\label{tab:floor}
\begin{center}
\small
\fitwidth{
\begin{tabular}{llccc}
\toprule
Control & Perturbation & Churn & Accuracy & Exact McNemar $p$ \\
\midrule
Null replicate, same weights, same condition & none at all & 63.2\% & $-1.04$ & .247 \\
MLPQ, \textsc{swap} & $3\times10^{-4}$, cosine $\ge0.999999$ & 38\% & $-1.72$ & \textbf{.015} \\
No entity features, \textsc{node} & zero, features already zeroed & 43\% & $-0.61$ & .415 \\
CONST, \textsc{swap} & zero, all token rows identical & 0\% & $0.00$ & 1 \\
\bottomrule
\end{tabular}}
\end{center}
\end{table}

\begin{table}[h]
\centering
\small
\setlength{\tabcolsep}{4.5pt}
\caption{\textbf{Information from the original sample retained under each condition.}
\cmark{} indicates that the information remains available when constructing the token;
\xmark{} indicates that it is removed or replaced. The final three rows are separately
trained ablations rather than fixed-weight interventions.}
\label{tab:taxonomy}
\begin{tabular}{@{}llcccc@{}}
\toprule
& & \multicolumn{4}{c}{\textbf{information from the original sample retained}} \\
\cmidrule(l){3-6}
\textbf{Condition} & \textbf{What it replaces}
& question & entities & structure & relations \\
\midrule
\multicolumn{6}{@{}l}{\emph{Interventions at fixed weights}} \\
\textsc{mean}
& token $\to$ training-set mean $\boldsymbol{\mu}$
& \xmark & \xmark & \xmark & \xmark \\
\textsc{swap}
& token $\to$ another question's token
& \xmark & \xmark & \xmark & \xmark \\
\textsc{gswap}
& graph $\to$ another question's graph
& \cmark & \xmark & \xmark & \xmark \\
\textsc{node}
& node features $\to$ resampled features
& \cmark & \xmark & \cmark & \cmark \\
\textsc{edge}
& edges $\to$ uniformly rewired edges
& \cmark & \cmark & \xmark & \cmark \\
\textsc{rel}
& relation features $\to$ resampled features
& \cmark & \cmark & \cmark & \xmark \\
\midrule
\multicolumn{6}{@{}l}{\emph{Pipelines retrained with one input removed}} \\
no entity features
& node features zeroed during training and testing
& \cmark & \xmark & \cmark & \cmark \\
no question
& question conditioning removed from Stage~1
& \xmark & \cmark & \cmark & \cmark \\
question MLP
& graph encoder replaced by an MLP on $q$
& \cmark & \xmark & \xmark & \xmark \\
\bottomrule
\end{tabular}
\end{table}

\textbf{Gold-answer likelihood.} With $y_i$ the gold answer of question $i$, truncated to the generation budget and terminated by the end-of-sequence token, and $\mathbf{t}_i^{c}$ the graph token under condition $c$:
\begin{equation}
\mathrm{NLL}_i(c) = -\frac{1}{|y_i|}\sum_{k=1}^{|y_i|}
  \log p_\theta\big(y_{i,k}\mid y_{i,<k},\, \mathbf{t}_i^{c},\, \tau(G_i),\, q_i\big),
\quad
\Delta_{\mathrm{NLL}}(c) = \frac{1}{n}\sum_{i=1}^{n}\big[\mathrm{NLL}_i(c) - \mathrm{NLL}_i(\textsc{real})\big].
\label{eq:nll}
\end{equation}
This is the token-averaged cross-entropy of the gold answer under teacher forcing, so nothing is generated and no decoding is involved; a positive $\Delta_{\mathrm{NLL}}$ means the intervention made the gold answer less likely, that is, the token was helping. The gold answer is truncated to 32 tokens plus the end-of-sequence token, the batch size is 1, and paired differences are reported with a 2,000-sample bootstrap confidence interval and a Wilcoxon signed-rank test.

\textbf{Decodability.} With $\mathbf{T}\in\mathbb{R}^{m\times d_\ell}$ the stored test tokens of the fitting half, $\mathbf{Y}$ the frozen sentence embeddings of the target texts and $\tilde{\mathbf{T}}=[\mathbf{T},\mathbf{1}]$,
\begin{equation}
\mathbf{W}^\star = \big(\tilde{\mathbf{T}}^\top\tilde{\mathbf{T}} + \alpha \mathbf{I}\big)^{-1}
                    \tilde{\mathbf{T}}^\top \mathbf{Y},
\label{eq:ridge}
\end{equation}
solved in double precision with $\alpha=10^{3}$. The held-out half is scored by retrieval: each question's predicted embedding is ranked against all held-out targets by cosine, and we report the median rank of its own target, with ties counted against us. The split is 50/50 with seed 0, the sentence encoder is the same frozen all-roberta-large-v1 used everywhere else, and the control replaces every token by the fitting half's mean, which lands at the median rank by construction.

Two caveats bound the graph column. The sentence encoder truncates at 512 wordpieces, and the verbalised subgraphs are post-retrieval: the median is 76 wordpieces on \texttt{ExplaGraphs} with nothing truncated, 260 on \texttt{WebQSP} with 14.5\% truncated and 256 on \texttt{SceneGraphs} with 23\%, so the instrument is intact on the benchmark where we lean on it and is a lower bound on the other two. Compressing a subgraph into one 1024-dimensional sentence vector is lossy in any case, so the column measures whether the token aligns with the gist of the verbalisation. Failing to decode is therefore not proof of absence, while decoding successfully is proof of presence, which is the direction every claim in the paper uses.

\textbf{Likelihood floor.}
The same null replicate gives a paired likelihood difference of $+0.00026$
with a 95\% confidence interval of $[-0.00027,+0.00080]$ and Wilcoxon
$p=.785$.

\section{Complete collapse statistics and per-encoder interventions}
\label{app:collapse-full}

Table~\ref{tab:collapse-geometry} gives the two token statistics dropped from Table~\ref{tab:collapse}, and Table~\ref{tab:collapse-encoders} the per-encoder interventions dropped from Table~\ref{tab:collapsed-role}. Every remaining condition of every model is in Appendix~\ref{app:interventions-full}. 

\section{Beyond G-Retriever: token collapse in subsequent graph-prompting methods}
\label{app:successors}

\textbf{Motivation and scope.}
We test whether collapse is confined to G-Retriever or recurs in later graph-prompting methods. We consider methods evaluated on \texttt{WebQSP} that construct continuous prompts from graphs and have runnable public implementations. These criteria retain GRAG~\citep{hu2024grag}, GRETPool~\citep{grover2026one}, and GraSP~\citep{wang2026topology}. We exclude Graph2Token~\citep{wang2025bridgingmoleculargraphslarge}, which targets molecular graphs, and GraphAdapter~\citep{li2023graphadaptertuningvisionlanguagemodels}, which adapts vision--language models. Align-GRAG~\citep{xu2025aligngrag} could not be reproduced because its anonymous repository was no longer available.

\textbf{Controlled setup.}
All methods use the 1,628-example \texttt{WebQSP} test split and the PCST subgraphs retrieved for G-Retriever. Holding retrieval fixed prevents graph size and relevance from explaining differences in token geometry, but makes GRAG-PCST and GraSP-PCST controlled adaptations rather than complete reproductions of the published systems. GRAG-PCST retains the released query-conditioned GAT, projector, and LoRA-tuned Llama configuration. GRETPool uses its reported TopK \citep{gao2019graphunets} and SAG \citep{lee2019selfattentiongraphpooling} ratios of $0.44$ and DiffPool \citep{ying2019hierarchicalgraphrepresentationlearning} budgets of 8 and 100 tokens. GraSP-PCST uses the released \textsc{SubgraphGNN} \citep{alsentzer2020subgraphneuralnetworks} and \textsc{Graph2Prefix} with a local Llama model, producing one token per
retained node. Table~\ref{tab:successor-configurations} summarises the resulting cardinality regimes.

\begin{table}[t]
\caption{\textbf{Cross-method configurations.}
$K$ is the number of graph-derived tokens and $N$ the retained PCST node count.
The projector column reports the final graph-to-LLM mapping.}
\label{tab:successor-configurations}
\begin{center}
\small
\fitwidth{
\begin{tabular}{lllcc}
\toprule
Method & Configuration & Projector & $K$ regime & Mean $K$ \\
\midrule
GRAG-PCST~\citep{hu2024grag}
 & released GAT + LoRA
 & Linear--Sigmoid--Linear
 & $1$ & 1.000 \\
GRETPool~\citep{grover2026one}
 & TopK~\citep{gao2019graphunets}, ratio $0.44$
 & Linear--Sigmoid--Linear
 & $\lceil0.44N\rceil$ & 7.848 \\
--
 & SAG~\citep{lee2019selfattentiongraphpooling}, ratio $0.44$
 & Linear--Sigmoid--Linear
 & $\lceil0.44N\rceil$ & 7.848 \\
--
 & DiffPool~\citep{ying2019hierarchicalgraphrepresentationlearning}
 & Linear--Sigmoid--Linear
 & $8$ & 8.000 \\
--
 & DiffPool~\citep{ying2019hierarchicalgraphrepresentationlearning}
 & Linear--Sigmoid--Linear
 & $100$ & 100.000 \\
GraSP-PCST~\citep{wang2026topology}
 & \textsc{SubgraphGNN}~\citep{alsentzer2020subgraphneuralnetworks} + Llama
 & Linear--Tanh--Linear
 & $N$ & 16.719 \\
\bottomrule
\end{tabular}}
\end{center}
\end{table}

\textbf{Diagnostics across cardinalities.}
For inter-sample comparison, we mean-pool each example's $K$ prompt tokens and compute the geometry statistics of Appendix~\ref{app:protocol}. This pooling is used only for measurement and does not alter inference. For $K>1$, we also report the mean pairwise cosine among tokens from the same graph. All interventions preserve cardinality: donor examples have the same $K$, and
node randomisation and edge rewiring preserve node and edge counts.

\begin{table*}[p]
\caption{\textbf{Cross-method token-collapse audit on \texttt{WebQSP}.}
Panel (a) reports geometry after mean-pooling each example's $K$ tokens;
within-example cosine is computed before pooling. Panels (b) and (c) report
Hit@1 differences with exact McNemar $p$-values in parentheses. Panel (b) uses
the text-present \textsc{real} condition as reference; Panel (c) uses each
model's \textsc{notxt-real} condition. Significance symbols follow
Section~\ref{sec:setup}. Results are single-seed and $p$-values are unadjusted;
NLL and multi-seed uncertainty are unavailable.}
\label{tab:successor-geometry}
\begin{center}
\scriptsize

\textbf{(a) Geometry and unmodified performance}

\medskip
\fitwidth{
\begin{tabular}{lccccccc}
\toprule
Method & Mean $K$ & Hit@1 & Inter cos. & RR & Norm ratio
& Within cos. & Eff.\ rank \\
\midrule
GRAG-PCST
 & 1.000 & 74.570 & 0.999171 & 0.028223 & 42.935 & -- & 10.300 \\
TopK
 & 7.848 & 71.437 & 1.000000 & 0.000155 & 28.258 & 1.000000 & -- \\
SAG
 & 7.848 & 72.727 & 1.000000 & 0.000531 & 25.802 & 0.999999 & -- \\
DiffPool ($K=8$)
 & 8.000 & 52.703 & 0.999712 & 0.009284 & 25.477 & 0.999856 & -- \\
DiffPool ($K=100$)
 & 100.000 & 61.364 & 0.999994 & 0.001291 & 27.170 & 1.000000 & -- \\
GraSP-PCST
 & 16.719 & 81.450 & 0.997783 & 0.039442 & 4.271 & 0.999465 & -- \\
\bottomrule
\end{tabular}}

\medskip
\textbf{(b) With-text interventions: $\Delta$Hit@1 ($p$) relative to \textsc{real}}

\medskip
\fitwidth{
\begin{tabular}{lcccccc}
\toprule
Method & \textsc{swap} & \textsc{w-shuf} & \textsc{gswap}
& \textsc{mean} & \textsc{node} & \textsc{edge} \\
\midrule
GRAG-PCST
 & $-0.246$ (.791)
 & --
 & $+0.061$ (1.000)
 & $+0.676$ (.379)
 & $-0.061$ (1.000)
 & $-1.413$ (.0616)\sig{\dag} \\
TopK
 & $+0.430$ (.659)
 & --
 & $-0.676$ (.514)
 & $-0.184$ (.877)
 & $-0.799$ (.390)
 & $+1.044$ (.254) \\
SAG
 & $-0.921$ (.298)
 & --
 & $-0.676$ (.462)
 & $+0.553$ (.543)
 & $-0.369$ (.685)
 & $+0.614$ (.493) \\
DiffPool ($K=8$)
 & $-0.614$ (.612)
 & --
 & $+0.614$ (.610)
 & $-0.184$ (.909)
 & $+0.676$ (.556)
 & $+2.088$ (.0525)\sig{\dag} \\
DiffPool ($K=100$)
 & $-0.430$ (.689)
 & --
 & $+0.184$ (.894)
 & $+0.246$ (.839)
 & $-0.553$ (.569)
 & $+0.430$ (.674) \\
GraSP-PCST
 & $-0.553$ (.460)
 & $+0.123$ (.930)
 & $-0.676$ (.379)
 & $+0.799$ (.263)
 & $-0.430$ (.586)
 & $-0.123$ (.924) \\
\bottomrule
\end{tabular}}

\medskip
\textbf{(c) Without-text interventions: \textsc{real} Hit@1 and
$\Delta$Hit@1 ($p$) relative to \textsc{notxt-real}}

\medskip
\fitwidth{
\begin{tabular}{lcccc}
\toprule
Method & \textsc{notxt-real} & \textsc{notxt-swap}
& \textsc{notxt-gswap} & \textsc{notxt-mean} \\
\midrule
GRAG-PCST
 & 41.585
 & $-1.044$ (.303)
 & $+0.184$ (.904)
 & $-0.369$ (.756) \\
TopK
 & 28.194
 & $-0.184$ (.904)
 & $+0.860$ (.519)
 & $-0.921$ (.395) \\
SAG
 & 31.511
 & $+0.307$ (.802)
 & $+0.369$ (.798)
 & $-0.369$ (.758) \\
DiffPool ($K=8$)
 & 27.150
 & $+2.334$ (.0530)\sig{\dag}
 & $+0.491$ (.719)
 & $+0.921$ (.463) \\
DiffPool ($K=100$)
 & 22.789
 & $-1.167$ (.224)
 & $-0.553$ (.605)
 & $-1.904$ (.0472)\sig{*} \\
GraSP-PCST
 & 38.391
 & $-0.123$ (.963)
 & $+2.518$ (.0767)\sig{\dag}
 & $+1.658$ (.234) \\
\bottomrule
\end{tabular}}
\end{center}
\end{table*}

\textbf{Results.}
Collapse occurs in all six configurations and every tested cardinality: inter-sample cosine is $0.998$--$1.000$, and the multi-token methods are also internally almost collinear. GRAG and GRETPool share G-Retriever's Linear--Sigmoid--Linear projector, so their results show that collapse persists under alternative encoders and pooling operators without changing this bottleneck. GraSP provides the independent case: despite using a Linear--Tanh--Linear projector and one token per retained node, it reaches
inter-sample cosine $0.9978$, within-example cosine $0.9995$, and residual ratio $0.039$. Its smaller norm ratio ($4.27$) shows that collapse need not involve the unusually large token norms observed with the Sigmoid projectors. Thus, collapse can persist with multiple tokens and with either Sigmoid or Tanh projectors.

With the verbalised graph present, none of the direct token-content controls resolves: \textsc{swap} changes Hit@1 by at most $0.921$ points ($p=.298$), \textsc{gswap} by at most $0.676$ ($p\geq.379$), and \textsc{mean} by at most $0.799$ ($p=.263$). Node randomisation is likewise
unresolved. Edge rewiring produces two marginal effects: $-1.413$ for GRAG-PCST ($p=.0616$) and $+2.088$ for DiffPool with $K=8$ ($p=.0525$). Neither satisfies the paper's joint requirement of $|\Delta|>2$ and $p<.05$,
and the latter improves rather than degrades performance.

The no-text tests use \textsc{notxt-real}, rather than the text-present score, as their paired reference. They provide limited, configuration-dependent evidence: DiffPool with $K=100$ decreases under the mean-token diagnostic ($-1.904$, $p=.0472$), while DiffPool with $K=8$ under \textsc{swap} ($+2.334$, $p=.0530$) and GraSP-PCST under \textsc{gswap} ($+2.518$, $p=.0767$) are marginal. None satisfies both the effect-size and $p$-value criteria. Removing the verbalised graph itself lowers performance by $25.6$--$43.2$ points, but this comparison measures removal of textual evidence rather than graph-token specificity.

These results extend the collapse observation beyond G-Retriever: it recurs under GRAG's query-conditioned single-token method and under fixed, ratio-selected, and node-preserving multi-token pooling. They do not show that
every graph-prompting method collapses or that graph information is universally unnecessary. The paired tests establish that token-content replacements are unresolved in every text-present configuration, while the no-text conditions show only limited and inconsistent sensitivity. Because this audit remains single-seed and lacks likelihood measurements, it supports a recurrence claim
rather than the stronger conclusions established for G-Retriever in the main experiments.

\begin{table}[h]
\caption{Residual and norm ratios of the trained tokens; norm ratio is relative to the median word embedding.}
\label{tab:collapse-geometry}
\begin{center}
\small
\fitwidth{
\begin{tabular}{llcc}
\toprule
Dataset & Model & Residual ratio & Norm ratio \\
\midrule
\multirow{3}{*}{\texttt{ExplaGraphs}} & G-Retriever (GT) & 0.050 & 21.9$\times$ \\
 & G-Retriever (GAT) & 0.040 & 24.4$\times$ \\
 & G-Retriever (GCN) & 0.121 & 24.3$\times$ \\
\midrule
\multirow{3}{*}{\texttt{SceneGraphs}} & G-Retriever (GT) & 0.088 & 84.8$\times$ \\
 & G-Retriever (GAT) & 0.046 & 87.6$\times$ \\
 & G-Retriever (GCN) & 0.048 & 97.0$\times$ \\
\midrule
\multirow{3}{*}{\texttt{WebQSP}} & G-Retriever (GT) & 0.057 & 29.4$\times$ \\
 & \quad + LoRA & 0.050 & 37.8$\times$ \\
 & \quad + Qwen2.5-3B & 0.047 & 22.8$\times$ \\
\bottomrule
\end{tabular}}
\end{center}
\end{table}

\begin{table}[h]
\caption{Interventions on the GAT and GCN encoders, as differences from REAL in accuracy points.}
\label{tab:collapse-encoders}
\begin{center}
\small
\fitwidth{
\begin{tabular}{llcccccc}
\toprule
Dataset & Model & REAL & MEAN & SWAP & NODE & EDGE & NOTXT-REAL \\
\midrule
\multirow{2}{*}{\texttt{ExplaGraphs}} & G-Retriever (GAT) & 85.20 & $-0.36$ & $0.00$ & $-0.18$ & $+0.18$ & 76.90 \\
 & G-Retriever (GCN) & 78.88 & $+0.90$ & $+0.36$ & $+1.08$ & $+0.72$ & 69.86 \\
\midrule
\multirow{2}{*}{\texttt{SceneGraphs}} & G-Retriever (GAT) & 80.20 & $+0.60$ & $+0.20$ & $-0.20$ & $+0.10$ & 28.65 \\
 & G-Retriever (GCN) & 80.55 & $-0.10$ & $-1.05$ & $-1.25$\sig{\dag} & $-1.25$\sig{\dag} & 29.80 \\
\bottomrule
\end{tabular}}
\end{center}
\end{table}

\section{The token space of every model}
\label{app:tokenspace}

The figures show the graph tokens of every test question. Their top rows project the mean-centred tokens onto two principal components and scale them by $\lVert\boldsymbol{\mu}\rVert$, making models with different token magnitudes comparable. The reported mean radius is measured in this two-dimensional projection and is therefore smaller than the full-dimensional residual ratio in Table~\ref{tab:collapse-geometry}. Panels with radius below $0.1$ include a magnified inset. The bottom rows show inter-sample cosine distributions on a fixed axis.

\begin{figure}[h]
\begin{center}
\includegraphics[width=\linewidth]{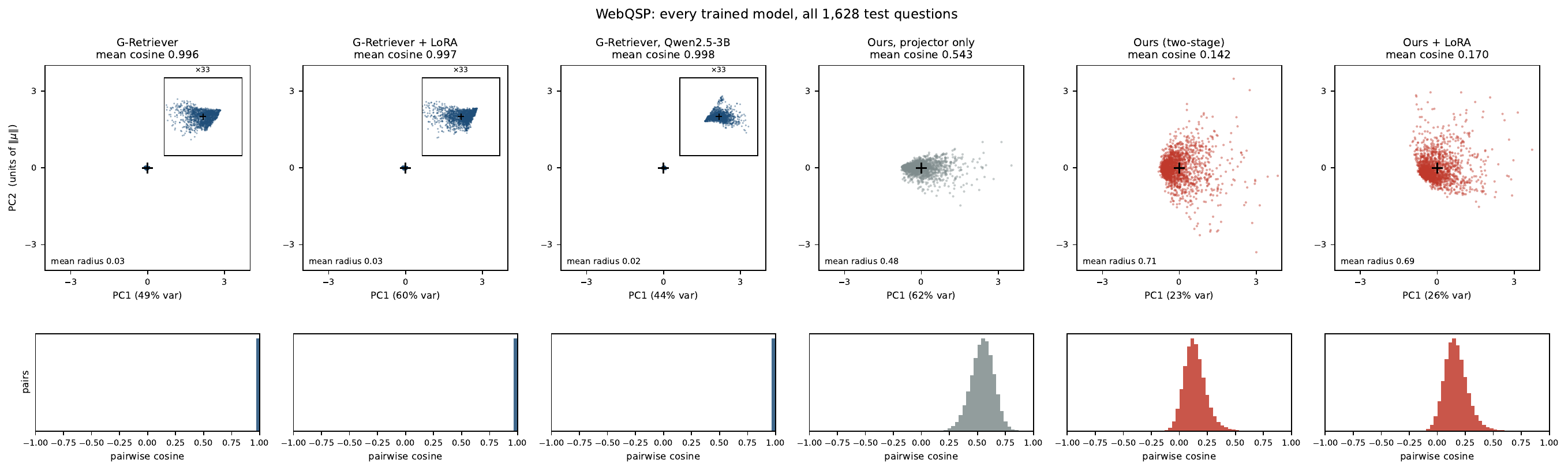}
\end{center}
\caption{\texttt{WebQSP}, all 1,628 test questions, six models. G-Retriever 0.9963 cosine and 0.031 mean radius; with LoRA 0.9970 and 0.032; with Qwen2.5-3B 0.9975 and 0.022; our projector-only ablation 0.5425 and 0.479; our two-stage pipeline 0.1424 and 0.714; ours with LoRA 0.1702 and 0.686. Fine-tuning the LLM and changing the LLM family leave the token as collapsed as it was, and the projector-only ablation sits between the two regimes, which is the visual counterpart of the trade-off in Table~\ref{tab:decomposition}.}
\label{fig:tokenspace-webqsp}
\end{figure}

\begin{figure}[h]
\begin{center}
\includegraphics[width=\linewidth]{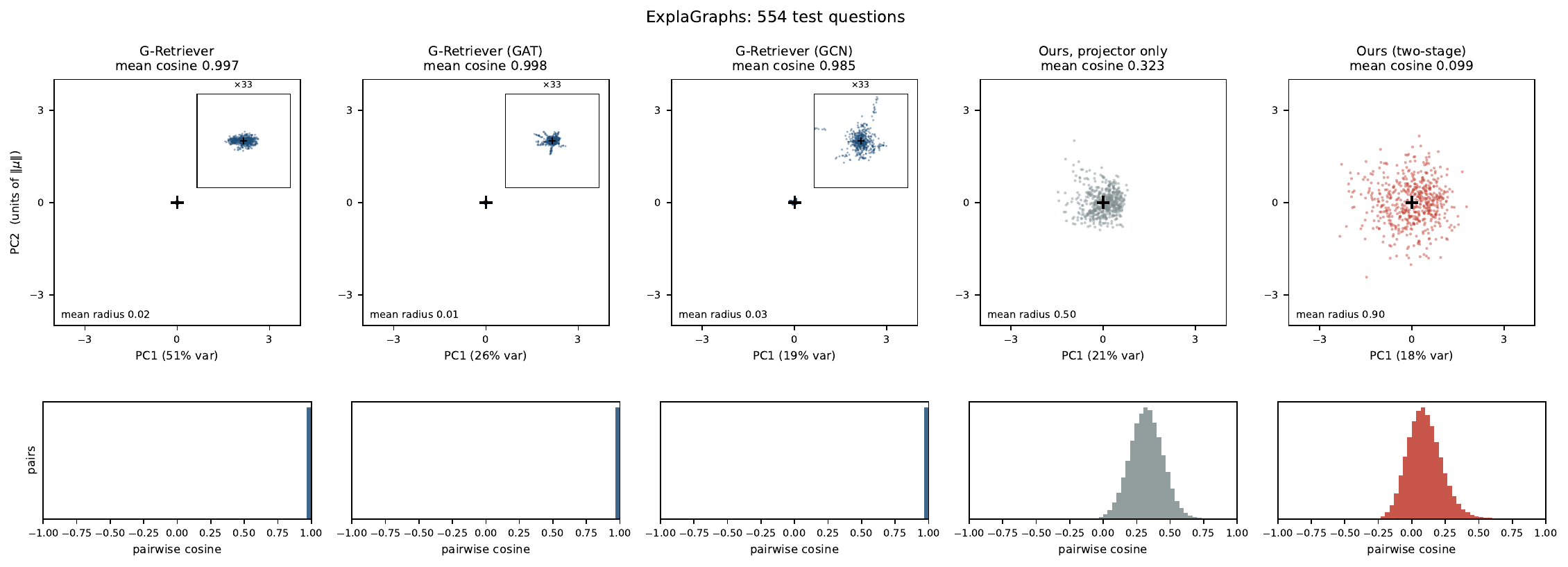}
\end{center}
\caption{\texttt{ExplaGraphs}, all 554 test questions, five models: the three graph encoders of G-Retriever, with cosine 0.9974, 0.9983 and 0.9852 and mean radius 0.017, 0.012 and 0.026; our projector-only ablation, 0.3229 and 0.503; and our two-stage pipeline under the cross-fitted recipe, 0.0986 and 0.895. Collapse is the same for all three encoders.}
\label{fig:tokenspace-expla}
\end{figure}

\begin{figure}[h]
\begin{center}
\includegraphics[width=\linewidth]{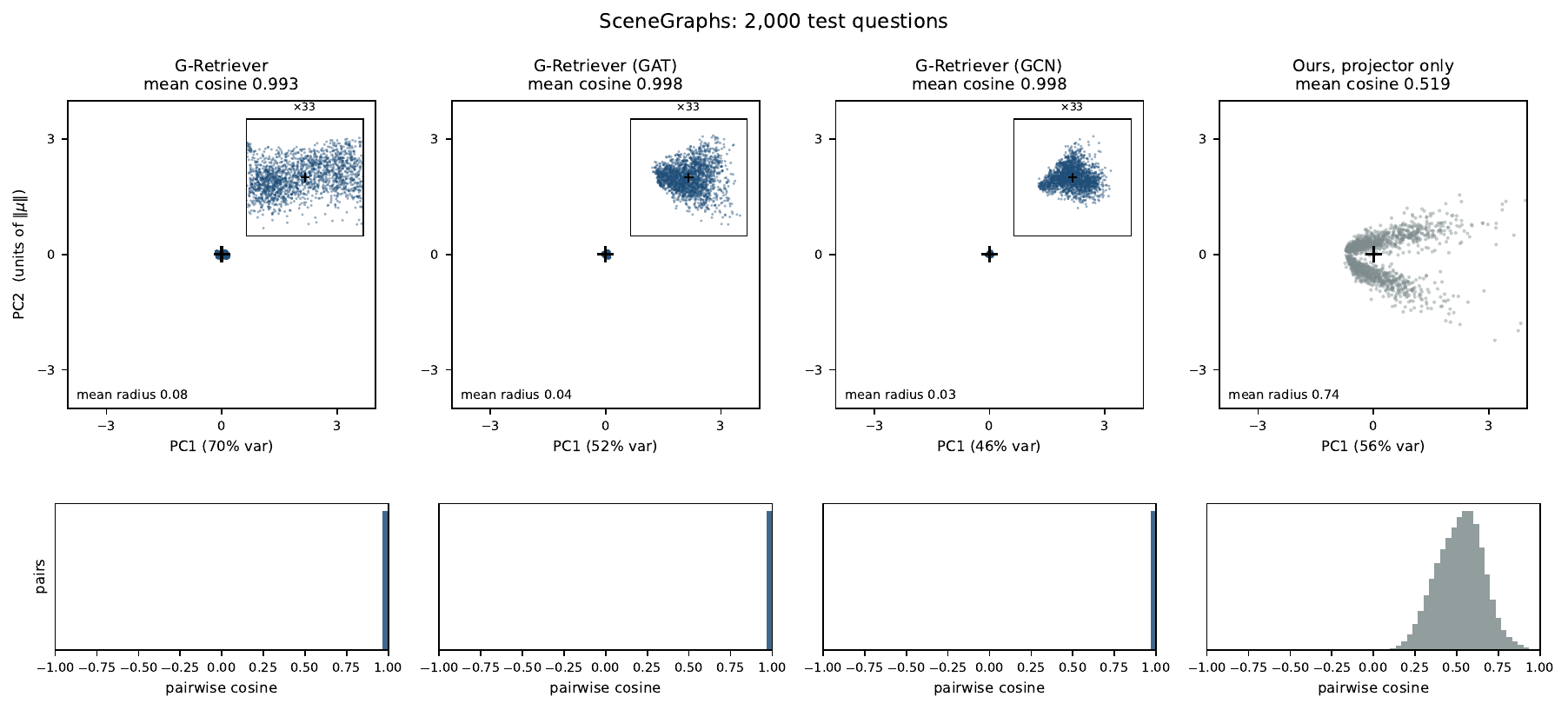}
\end{center}
\caption{\texttt{SceneGraphs}, the first 2,000 test questions, four models: G-Retriever with a graph transformer (cosine 0.9934, mean radius 0.077), with GAT (0.9980, 0.040), with GCN (0.9979, 0.032), and our projector-only ablation (0.5194, 0.736).}
\label{fig:tokenspace-scene}
\end{figure}

\begin{figure}[h]
\begin{center}
\includegraphics[width=\linewidth]{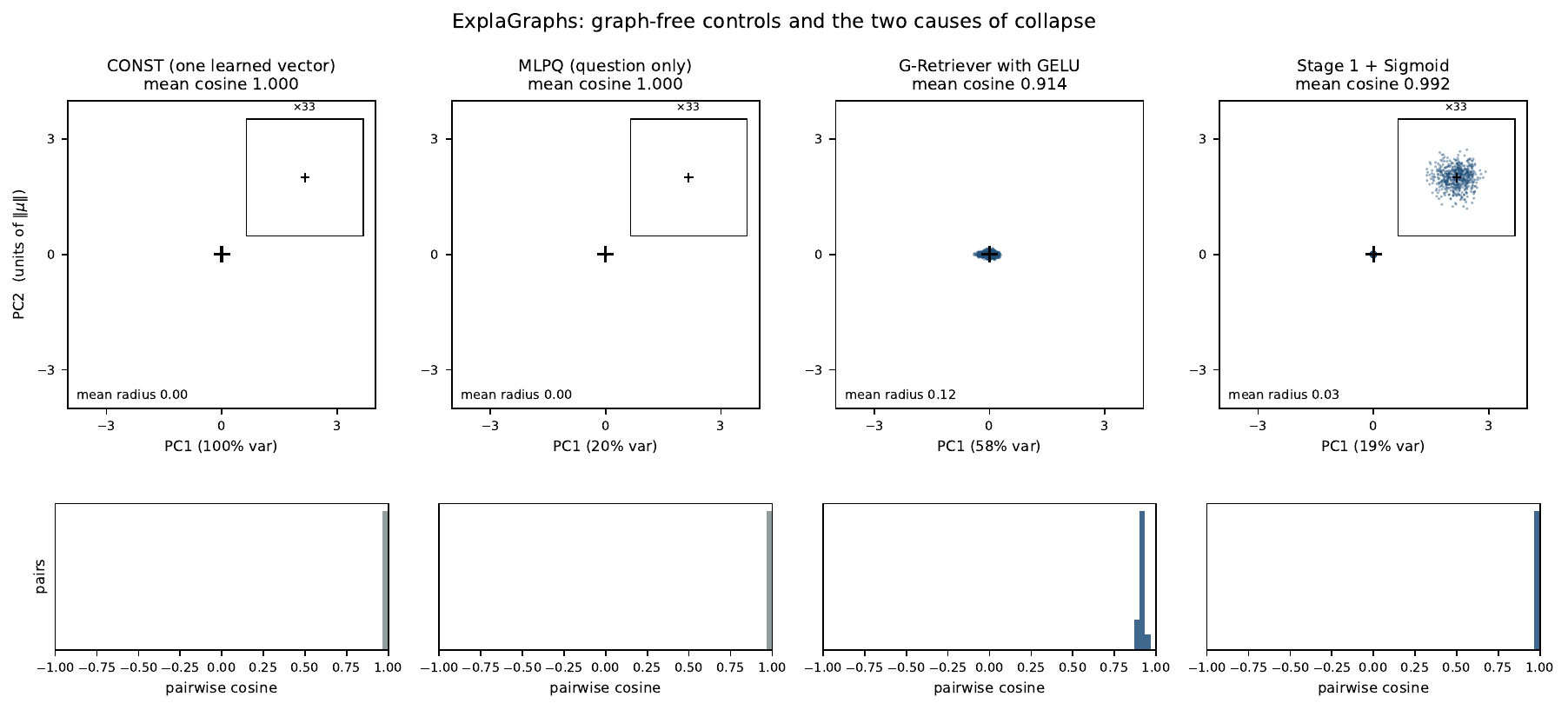}
\end{center}
\caption{\texttt{ExplaGraphs}: the graph-free controls and the two causes of collapse. CONST (cosine 1.0000, radius 0.000) is a single point by construction and is the reference for a fully collapsed token; MLPQ (1.0000, 0.000) shows that a token computed from the question collapses just as completely; G-Retriever with GELU (0.9139, 0.116) shows that removing the saturating non-linearity alone is not enough while the encoder trains; Stage~1 with a Sigmoid projector (0.9918, 0.027) shows that an encoder whose outputs are nearly orthogonal still produces collapsed tokens. The last two panels are the visual form of Table~\ref{tab:mechanism}.}
\label{fig:tokenspace-controls}
\end{figure}

\clearpage

\section{Mechanism diagnostics}
\label{app:mechanism}

Table~\ref{tab:untrained} supports the first association in Section~\ref{sec:mechanism}. An untrained projector, at its default initialisation, is applied to 500 random inputs of fixed norm, and the tokens it produces are almost identical to each other unless the input norm is two orders of magnitude larger than anything a trained encoder produces.

\begin{table}[h]
\caption{Untrained projector (Equation~\ref{eq:projector}, default initialisation) applied to 500 random inputs of fixed norm. Collapse precedes training and is a property of the parameterisation.}
\label{tab:untrained}
\begin{center}
\small
\fitwidth{
\begin{tabular}{lccc}
\toprule
Input norm & Token norm & Token cosine & Residual ratio \\
\midrule
1 & 18.4 & 0.9999 & 0.009 \\
10 & 18.4 & 0.9919 & 0.090 \\
100 & 21.4 & 0.7346 & 0.600 \\
\bottomrule
\end{tabular}}
\end{center}
\end{table}

\textbf{Why Sigmoid can create a shared direction.}
Let $\mathbf{x}=\mathbf{W}_1\mathbf{h}_i+\mathbf{b}_1$. If these preactivations are near zero, then
$\sigma(\mathbf{x})\approx\frac12\mathbf{1}+\frac14\mathbf{x}$, and the projector is approximately \[
P_\psi(\mathbf{h}_i)
\approx
\frac12\mathbf{W}_2\mathbf{1}
+\frac14\mathbf{W}_2\mathbf{b}_1
+\mathbf{b}_2
+\frac14\mathbf{W}_2\mathbf{W}_1\mathbf{h}_i.
\]
The first three terms are input-independent, while the factor $\frac14$ attenuates input-specific variation. This provides one possible explanation for the shared direction produced by the untrained projector. It is not a complete explanation: GraSP remains collapsed with a zero-centred Tanh projector (Appendix~\ref{app:successors}), showing that Sigmoid is neither necessary nor the sole cause.

The approximation therefore motivates a mechanism consistent with Table~\ref{tab:untrained}, rather than establishing that the projector operates entirely in this local regime.

\section{Full fixed-weight interventions}
\label{app:interventions-full}

The following tables report the available fixed-weight interventions for each model. Each table gives absolute accuracy, the difference from the reference condition, and the exact McNemar $p$-value. The reference is \textsc{real}, or \textsc{notxt-real} for conditions without the verbalised graph. The second and third \textsc{swap} rows, where present, use independent derangements. The reference is REAL, and NOTXT-REAL for the block without the verbalised graph. Conditions are defined in Section~\ref{sec:protocol} and Table~\ref{tab:taxonomy}; the second and third \textsc{swap} rows are replicates under different derangements. Both stages of our pipeline are trained on all training questions unless a caption says cross-fitted. No effect below 2.0 accuracy points licenses a claim, whatever its $p$-value (Appendix~\ref{app:protocol}).

\subsection{\texttt{ExplaGraphs}}

\begin{table}[h]
\caption{\textbf{G-Retriever (graph transformer)}, 554 test questions.}
\label{tab:full1}
\begin{center}\small
\begin{tabular}{lccc}
\toprule
Condition & Accuracy & Difference & $p$ \\
\midrule
REAL & 85.56 & -- & -- \\
MEAN & 85.38 & $-$0.18 & 1 \\
SWAP & 84.30 & $-$1.26 & 0.118 \\
$\boldsymbol{\mu}+0\cdot(\mathbf{t}_i-\boldsymbol{\mu})$ & 85.20 & $-$0.36 & 0.727 \\
$\boldsymbol{\mu}+2(\mathbf{t}_i-\boldsymbol{\mu})$ & 84.48 & $-$1.08 & 0.146 \\
NODE (real rows) & 84.66 & $-$0.90 & 0.227 \\
NODE (Gaussian) & 85.20 & $-$0.36 & 0.727 \\
EDGE (rewired) & 85.02 & $-$0.54 & 0.581 \\
REL (resampled) & 85.74 & $+$0.18 & 1 \\
zero token & 3.61 & $-$81.95\sig{***} & $1.09\times10^{-131}$ \\
norm $\rightarrow$ word size & 3.61 & $-$81.95\sig{***} & $3.17\times10^{-130}$ \\
$\times0.5$ & 77.44 & $-$8.12\sig{***} & $6.9\times10^{-6}$ \\
$\times2$ & 81.95 & $-$3.61\sig{***} & 0.000535 \\
\midrule
NOTXT-REAL & 66.43 & -- & -- \\
NOTXT-MEAN & 65.52 & $-$0.90 & 0.267 \\
NOTXT-SWAP & 66.43 & $+$0.00 & 1 \\
NOTXT-NODE & 66.25 & $-$0.18 & 1 \\
NOTXT-zero & 12.09 & $-$54.33\sig{***} & $2.46\times10^{-66}$ \\
NOTXT-norm $\rightarrow$ word size & 10.83 & $-$55.60\sig{***} & $7.08\times10^{-70}$ \\
\bottomrule
\end{tabular}
\end{center}
\end{table}

\begin{table}[h]
\caption{\textbf{G-Retriever, initialisation 2}, 554 test questions.}
\label{tab:full2}
\begin{center}\small
\begin{tabular}{lccc}
\toprule
Condition & Accuracy & Difference & $p$ \\
\midrule
REAL & 86.28 & -- & -- \\
MEAN & 84.84 & $-$1.44 & 0.134 \\
SWAP & 85.02 & $-$1.26 & 0.265 \\
$\boldsymbol{\mu}+0\cdot(\mathbf{t}_i-\boldsymbol{\mu})$ & 84.84 & $-$1.44 & 0.134 \\
$\boldsymbol{\mu}+2(\mathbf{t}_i-\boldsymbol{\mu})$ & 83.57 & $-$2.71\sig{***} & 0.000275 \\
NODE (real rows) & 86.10 & $-$0.18 & 1 \\
NODE (Gaussian) & 86.64 & $+$0.36 & 0.754 \\
EDGE (rewired) & 84.84 & $-$1.44 & 0.152 \\
REL (resampled) & 85.74 & $-$0.54 & 0.453 \\
zero token & 3.25 & $-$83.03\sig{***} & $4.53\times10^{-135}$ \\
norm $\rightarrow$ word size & 3.43 & $-$82.85\sig{***} & $1.55\times10^{-136}$ \\
$\times0.5$ & 66.06 & $-$20.22\sig{***} & $1.26\times10^{-18}$ \\
$\times2$ & 82.13 & $-$4.15\sig{*} & 0.0451 \\
\midrule
NOTXT-REAL & 71.84 & -- & -- \\
NOTXT-MEAN & 71.48 & $-$0.36 & 0.912 \\
NOTXT-SWAP & 70.76 & $-$1.08 & 0.61 \\
NOTXT-NODE & 73.10 & $+$1.26 & 0.464 \\
NOTXT-zero & 11.37 & $-$60.47\sig{***} & $1.4\times10^{-86}$ \\
NOTXT-norm $\rightarrow$ word size & 11.01 & $-$60.83\sig{***} & $5.72\times10^{-89}$ \\
\bottomrule
\end{tabular}
\end{center}
\end{table}

\begin{table}[h]
\caption{\textbf{G-Retriever, initialisation 3}, 554 test questions.}
\label{tab:full3}
\begin{center}\small
\begin{tabular}{lccc}
\toprule
Condition & Accuracy & Difference & $p$ \\
\midrule
REAL & 87.91 & -- & -- \\
MEAN & 87.18 & $-$0.72 & 0.219 \\
SWAP & 87.73 & $-$0.18 & 1 \\
$\boldsymbol{\mu}+0\cdot(\mathbf{t}_i-\boldsymbol{\mu})$ & 87.73 & $-$0.18 & 1 \\
$\boldsymbol{\mu}+2(\mathbf{t}_i-\boldsymbol{\mu})$ & 87.91 & $+$0.00 & 1 \\
NODE (real rows) & 88.09 & $+$0.18 & 1 \\
NODE (Gaussian) & 87.91 & $+$0.00 & 1 \\
EDGE (rewired) & 87.18 & $-$0.72 & 0.424 \\
REL (resampled) & 87.73 & $-$0.18 & 1 \\
zero token & 3.25 & $-$84.66\sig{***} & $1.31\times10^{-141}$ \\
norm $\rightarrow$ word size & 4.51 & $-$83.39\sig{***} & $1.14\times10^{-135}$ \\
$\times0.5$ & 67.87 & $-$20.04\sig{***} & $3\times10^{-20}$ \\
$\times2$ & 85.56 & $-$2.35\sig{\dag} & 0.066 \\
\midrule
NOTXT-REAL & 70.22 & -- & -- \\
NOTXT-MEAN & 69.68 & $-$0.54 & 0.648 \\
NOTXT-SWAP & 69.86 & $-$0.36 & 0.851 \\
NOTXT-NODE & 71.30 & $+$1.08 & 0.286 \\
NOTXT-zero & 11.37 & $-$58.84\sig{***} & $2.37\times10^{-75}$ \\
NOTXT-norm $\rightarrow$ word size & 11.73 & $-$58.48\sig{***} & $8.31\times10^{-75}$ \\
\bottomrule
\end{tabular}
\end{center}
\end{table}

\begin{table}[h]
\caption{\textbf{G-Retriever, initialisation 4}, 554 test questions.}
\label{tab:full4}
\begin{center}\small
\begin{tabular}{lccc}
\toprule
Condition & Accuracy & Difference & $p$ \\
\midrule
REAL & 89.71 & -- & -- \\
MEAN & 89.71 & $+$0.00 & 1 \\
SWAP & 90.07 & $+$0.36 & 0.625 \\
$\boldsymbol{\mu}+0\cdot(\mathbf{t}_i-\boldsymbol{\mu})$ & 89.71 & $+$0.00 & 1 \\
$\boldsymbol{\mu}+2(\mathbf{t}_i-\boldsymbol{\mu})$ & 90.07 & $+$0.36 & 0.5 \\
NODE (real rows) & 89.71 & $+$0.00 & 1 \\
NODE (Gaussian) & 89.71 & $+$0.00 & 1 \\
EDGE (rewired) & 90.07 & $+$0.36 & 0.688 \\
REL (resampled) & 90.07 & $+$0.36 & 0.625 \\
zero token & 2.17 & $-$87.55\sig{***} & $2\times10^{-146}$ \\
norm $\rightarrow$ word size & 3.25 & $-$86.46\sig{***} & $1.28\times10^{-144}$ \\
$\times0.5$ & 15.88 & $-$73.83\sig{***} & $4.29\times10^{-113}$ \\
$\times2$ & 82.31 & $-$7.40\sig{***} & $1.53\times10^{-6}$ \\
\midrule
NOTXT-REAL & 83.39 & -- & -- \\
NOTXT-MEAN & 84.12 & $+$0.72 & 0.672 \\
NOTXT-SWAP & 82.49 & $-$0.90 & 0.603 \\
NOTXT-NODE & 81.95 & $-$1.44 & 0.358 \\
NOTXT-zero & 12.45 & $-$70.94\sig{***} & $3\times10^{-103}$ \\
NOTXT-norm $\rightarrow$ word size & 13.00 & $-$70.40\sig{***} & $2.57\times10^{-104}$ \\
\bottomrule
\end{tabular}
\end{center}
\end{table}

\clearpage

\begin{table}[h]
\caption{\textbf{G-Retriever (GAT)}, 554 test questions.}
\label{tab:full5}
\begin{center}\small
\begin{tabular}{lccc}
\toprule
Condition & Accuracy & Difference & $p$ \\
\midrule
REAL & 85.20 & -- & -- \\
MEAN & 84.84 & $-$0.36 & 0.754 \\
SWAP & 85.20 & $+$0.00 & 1 \\
$\boldsymbol{\mu}+0\cdot(\mathbf{t}_i-\boldsymbol{\mu})$ & 85.02 & $-$0.18 & 1 \\
$\boldsymbol{\mu}+2(\mathbf{t}_i-\boldsymbol{\mu})$ & 85.20 & $+$0.00 & 1 \\
NODE (real rows) & 85.02 & $-$0.18 & 1 \\
NODE (Gaussian) & 84.66 & $-$0.54 & 0.508 \\
EDGE (rewired) & 85.38 & $+$0.18 & 1 \\
REL (resampled) & 84.84 & $-$0.36 & 0.5 \\
zero token & 3.61 & $-$81.59\sig{***} & $4.3\times10^{-131}$ \\
norm $\rightarrow$ word size & 3.07 & $-$82.13\sig{***} & $1.42\times10^{-133}$ \\
$\times0.5$ & 77.62 & $-$7.58\sig{***} & $3.89\times10^{-5}$ \\
$\times2$ & 83.21 & $-$1.99 & 0.161 \\
\midrule
NOTXT-REAL & 76.90 & -- & -- \\
NOTXT-MEAN & 76.90 & $+$0.00 & 1 \\
NOTXT-SWAP & 76.35 & $-$0.54 & 0.701 \\
NOTXT-NODE & 77.08 & $+$0.18 & 1 \\
NOTXT-zero & 12.09 & $-$64.80\sig{***} & $2.39\times10^{-97}$ \\
NOTXT-norm $\rightarrow$ word size & 10.47 & $-$66.43\sig{***} & $5.65\times10^{-100}$ \\
\bottomrule
\end{tabular}
\end{center}
\end{table}

\begin{table}[h]
\caption{\textbf{G-Retriever (GCN)}, 554 test questions.}
\label{tab:full6}
\begin{center}\small
\begin{tabular}{lccc}
\toprule
Condition & Accuracy & Difference & $p$ \\
\midrule
REAL & 78.88 & -- & -- \\
MEAN & 79.78 & $+$0.90 & 0.405 \\
SWAP & 79.24 & $+$0.36 & 0.86 \\
$\boldsymbol{\mu}+0\cdot(\mathbf{t}_i-\boldsymbol{\mu})$ & 80.14 & $+$1.26 & 0.23 \\
$\boldsymbol{\mu}+2(\mathbf{t}_i-\boldsymbol{\mu})$ & 77.44 & $-$1.44 & 0.215 \\
NODE (real rows) & 79.96 & $+$1.08 & 0.345 \\
NODE (Gaussian) & 78.88 & $+$0.00 & 1 \\
EDGE (rewired) & 79.60 & $+$0.72 & 0.503 \\
REL (resampled) & 79.42 & $+$0.54 & 0.25 \\
zero token & 3.61 & $-$75.27\sig{***} & $1.16\times10^{-120}$ \\
norm $\rightarrow$ word size & 2.89 & $-$75.99\sig{***} & $7.49\times10^{-122}$ \\
$\times0.5$ & 40.61 & $-$38.27\sig{***} & $1.43\times10^{-35}$ \\
$\times2$ & 70.58 & $-$8.30\sig{***} & 0.00086 \\
\midrule
NOTXT-REAL & 69.86 & -- & -- \\
NOTXT-MEAN & 69.49 & $-$0.36 & 0.845 \\
NOTXT-SWAP & 69.31 & $-$0.54 & 0.85 \\
NOTXT-NODE & 67.87 & $-$1.99\sig{\dag} & 0.0895 \\
NOTXT-zero & 12.09 & $-$57.76\sig{***} & $1.22\times10^{-80}$ \\
NOTXT-norm $\rightarrow$ word size & 10.65 & $-$59.21\sig{***} & $1.86\times10^{-85}$ \\
\bottomrule
\end{tabular}
\end{center}
\end{table}

\begin{table}[h]
\caption{\textbf{CONST}, 554 test questions.}
\label{tab:full7}
\begin{center}\small
\begin{tabular}{lccc}
\toprule
Condition & Accuracy & Difference & $p$ \\
\midrule
REAL & 85.56 & -- & -- \\
MEAN & 85.02 & $-$0.54 & 0.375 \\
SWAP & 85.56 & $+$0.00 & 1 \\
$\boldsymbol{\mu}+0\cdot(\mathbf{t}_i-\boldsymbol{\mu})$ & 85.02 & $-$0.54 & 0.375 \\
$\boldsymbol{\mu}+2(\mathbf{t}_i-\boldsymbol{\mu})$ & 84.84 & $-$0.72 & 0.125 \\
NODE (real rows) & 85.56 & $+$0.00 & 1 \\
NODE (Gaussian) & 85.56 & $+$0.00 & 1 \\
EDGE (rewired) & 85.56 & $+$0.00 & 1 \\
REL (resampled) & 85.56 & $+$0.00 & 1 \\
zero token & 2.17 & $-$83.39\sig{***} & $1.14\times10^{-135}$ \\
norm $\rightarrow$ word size & 5.42 & $-$80.14\sig{***} & $6.84\times10^{-126}$ \\
$\times0.5$ & 61.73 & $-$23.83\sig{***} & $2.23\times10^{-23}$ \\
$\times2$ & 83.39 & $-$2.17\sig{\dag} & 0.0884 \\
\midrule
NOTXT-REAL & 51.99 & -- & -- \\
NOTXT-MEAN & 51.62 & $-$0.36 & 0.5 \\
NOTXT-SWAP & 51.99 & $+$0.00 & 1 \\
NOTXT-NODE & 51.99 & $+$0.00 & 1 \\
NOTXT-zero & 12.27 & $-$39.71\sig{***} & $3.29\times10^{-38}$ \\
NOTXT-norm $\rightarrow$ word size & 11.19 & $-$40.79\sig{***} & $1.71\times10^{-40}$ \\
\bottomrule
\end{tabular}
\end{center}
\end{table}

\begin{table}[h]
\caption{\textbf{CONST at word norm}, 554 test questions.}
\label{tab:full8}
\begin{center}\small
\begin{tabular}{lccc}
\toprule
Condition & Accuracy & Difference & $p$ \\
\midrule
REAL & 85.56 & -- & -- \\
MEAN & 85.02 & $-$0.54 & 0.25 \\
SWAP & 85.56 & $+$0.00 & 1 \\
$\boldsymbol{\mu}+0\cdot(\mathbf{t}_i-\boldsymbol{\mu})$ & 85.02 & $-$0.54 & 0.25 \\
$\boldsymbol{\mu}+2(\mathbf{t}_i-\boldsymbol{\mu})$ & 85.56 & $+$0.00 & 1 \\
NODE (real rows) & 85.56 & $+$0.00 & 1 \\
NODE (Gaussian) & 85.56 & $+$0.00 & 1 \\
EDGE (rewired) & 85.56 & $+$0.00 & 1 \\
REL (resampled) & 85.56 & $+$0.00 & 1 \\
zero token & 2.17 & $-$83.39\sig{***} & $1.14\times10^{-135}$ \\
norm $\rightarrow$ word size & 85.56 & $+$0.00 & 1 \\
$\times0.5$ & 4.87 & $-$80.69\sig{***} & $8.83\times10^{-127}$ \\
$\times2$ & 7.04 & $-$78.52\sig{***} & $3.17\times10^{-123}$ \\
\midrule
NOTXT-REAL & 59.75 & -- & -- \\
NOTXT-MEAN & 57.22 & $-$2.53 & 0.261 \\
NOTXT-SWAP & 59.75 & $+$0.00 & 1 \\
NOTXT-NODE & 59.75 & $+$0.00 & 1 \\
NOTXT-zero & 12.27 & $-$47.47\sig{***} & $1.57\times10^{-51}$ \\
NOTXT-norm $\rightarrow$ word size & 59.75 & $+$0.00 & 1 \\
\bottomrule
\end{tabular}
\end{center}
\end{table}

\clearpage

\begin{table}[h]
\caption{\textbf{MLPQ}, 554 test questions.}
\label{tab:full9}
\begin{center}\small
\begin{tabular}{lccc}
\toprule
Condition & Accuracy & Difference & $p$ \\
\midrule
REAL & 84.84 & -- & -- \\
MEAN & 85.02 & $+$0.18 & 1 \\
SWAP & 85.20 & $+$0.36 & 0.688 \\
$\boldsymbol{\mu}+0\cdot(\mathbf{t}_i-\boldsymbol{\mu})$ & 85.02 & $+$0.18 & 1 \\
$\boldsymbol{\mu}+2(\mathbf{t}_i-\boldsymbol{\mu})$ & 85.20 & $+$0.36 & 0.5 \\
NODE (real rows) & 84.84 & $+$0.00 & 1 \\
NODE (Gaussian) & 84.84 & $+$0.00 & 1 \\
EDGE (rewired) & 84.84 & $+$0.00 & 1 \\
REL (resampled) & 84.84 & $+$0.00 & 1 \\
zero token & 3.61 & $-$81.23\sig{***} & $1.7\times10^{-130}$ \\
norm $\rightarrow$ word size & 3.97 & $-$80.87\sig{***} & $6.71\times10^{-130}$ \\
$\times0.5$ & 74.19 & $-$10.65\sig{***} & $1.26\times10^{-7}$ \\
$\times2$ & 84.12 & $-$0.72 & 0.597 \\
\midrule
NOTXT-REAL & 77.62 & -- & -- \\
NOTXT-MEAN & 77.62 & $+$0.00 & 1 \\
NOTXT-SWAP & 77.80 & $+$0.18 & 1 \\
NOTXT-NODE & 77.62 & $+$0.00 & 1 \\
NOTXT-zero & 12.09 & $-$65.52\sig{***} & $3.19\times10^{-91}$ \\
NOTXT-norm $\rightarrow$ word size & 9.75 & $-$67.87\sig{***} & $2.61\times10^{-102}$ \\
\bottomrule
\end{tabular}
\end{center}
\end{table}

\begin{table}[h]
\caption{\textbf{G-Retriever with GELU (no Stage 1)}, 554 test questions.}
\label{tab:full10}
\begin{center}\small
\begin{tabular}{lccc}
\toprule
Condition & Accuracy & Difference & $p$ \\
\midrule
REAL & 73.29 & -- & -- \\
SWAP & 72.20 & $-$1.08 & 0.631 \\
\bottomrule
\end{tabular}
\end{center}
\end{table}

\begin{table}[h]
\caption{\textbf{Stage 1 + Sigmoid projector}, 554 test questions.}
\label{tab:full11}
\begin{center}\small
\begin{tabular}{lccc}
\toprule
Condition & Accuracy & Difference & $p$ \\
\midrule
REAL & 84.48 & -- & -- \\
MEAN & 84.48 & $+$0.00 & 1 \\
SWAP & 84.66 & $+$0.18 & 1 \\
\bottomrule
\end{tabular}
\end{center}
\end{table}

\begin{table}[h]
\caption{\textbf{Stage 1 + GELU, projector only}, 554 test questions.}
\label{tab:full12}
\begin{center}\small
\begin{tabular}{lccc}
\toprule
Condition & Accuracy & Difference & $p$ \\
\midrule
REAL & 57.76 & -- & -- \\
MEAN & 58.84 & $+$1.08 & 0.735 \\
SWAP & 51.08 & $-$6.68\sig{*} & 0.0167 \\
NODE (real rows) & 56.50 & $-$1.26 & 0.685 \\
NODE (Gaussian) & 48.56 & $-$9.21\sig{***} & 0.000616 \\
EDGE (rewired) & 51.62 & $-$6.14\sig{*} & 0.02 \\
REL (resampled) & 55.96 & $-$1.81 & 0.529 \\
norm $\rightarrow$ word size & 3.97 & $-$53.79\sig{***} & $1.32\times10^{-79}$ \\
$\times0.5$ & 33.03 & $-$24.73\sig{***} & $1.18\times10^{-17}$ \\
$\times2$ & 57.58 & $-$0.18 & 1 \\
\bottomrule
\end{tabular}
\end{center}
\end{table}

\clearpage

\begin{table}[h]
\caption{\textbf{Ours, joint (the method)}, 554 test questions.}
\label{tab:full13}
\begin{center}\small
\begin{tabular}{lccc}
\toprule
Condition & Accuracy & Difference & $p$ \\
\midrule
REAL & 85.74 & -- & -- \\
MEAN & 86.10 & $+$0.36 & 0.832 \\
SWAP & 85.20 & $-$0.54 & 0.749 \\
$\boldsymbol{\mu}+0\cdot(\mathbf{t}_i-\boldsymbol{\mu})$ & 86.10 & $+$0.36 & 0.832 \\
NODE (real rows) & 85.20 & $-$0.54 & 0.648 \\
NODE (Gaussian) & 86.10 & $+$0.36 & 0.824 \\
EDGE (rewired) & 86.82 & $+$1.08 & 0.21 \\
GSWAP (other graph) & 84.30 & $-$1.44 & 0.243 \\
SWAP (permutation 2) & 86.82 & $+$1.08 & 0.362 \\
SWAP (permutation 3) & 85.02 & $-$0.72 & 0.585 \\
\midrule
NOTXT-REAL & 69.68 & -- & -- \\
NOTXT-MEAN & 69.86 & $+$0.18 & 1 \\
NOTXT-SWAP & 69.86 & $+$0.18 & 1 \\
NOTXT-GSWAP & 69.13 & $-$0.54 & 0.771 \\
\bottomrule
\end{tabular}
\end{center}
\end{table}

\begin{table}[h]
\caption{\textbf{Ours, joint, cross-fitted}, 554 test questions, cross-fitted recipe.}
\label{tab:full14}
\begin{center}\small
\begin{tabular}{lccc}
\toprule
Condition & Accuracy & Difference & $p$ \\
\midrule
REAL & 76.35 & -- & -- \\
MEAN & 72.74 & $-$3.61\sig{*} & 0.0265 \\
SWAP & 77.44 & $+$1.08 & 0.545 \\
$\boldsymbol{\mu}+0\cdot(\mathbf{t}_i-\boldsymbol{\mu})$ & 72.92 & $-$3.43\sig{*} & 0.037 \\
NODE (real rows) & 77.08 & $+$0.72 & 0.636 \\
NODE (Gaussian) & 75.45 & $-$0.90 & 0.511 \\
EDGE (rewired) & 76.53 & $+$0.18 & 1 \\
\midrule
NOTXT-REAL & 64.44 & -- & -- \\
NOTXT-MEAN & 61.37 & $-$3.07 & 0.292 \\
NOTXT-SWAP & 64.62 & $+$0.18 & 1 \\
\bottomrule
\end{tabular}
\end{center}
\end{table}

\begin{table}[h]
\caption{\textbf{Ours, no entity features}, 554 test questions.}
\label{tab:full47}
\begin{center}\small
\begin{tabular}{lccc}
\toprule
Condition & Accuracy & Difference & $p$ \\
\midrule
REAL & 84.48 & -- & -- \\
MEAN & 87.55 & $+$3.07\sig{**} & 0.00232 \\
SWAP & 84.48 & $+$0.00 & 1 \\
NODE (real rows) & 84.66 & $+$0.18 & 1 \\
EDGE (rewired) & 84.66 & $+$0.18 & 1 \\
GSWAP (other graph) & 84.12 & $-$0.36 & 0.754 \\
\midrule
NOTXT-REAL & 70.22 & -- & -- \\
NOTXT-MEAN & 70.94 & $+$0.72 & 0.834 \\
\bottomrule
\end{tabular}
\end{center}
\end{table}

\subsection{\texttt{SceneGraphs}}

\begin{table}[h]
\caption{\textbf{G-Retriever (graph transformer)}, 2000 test questions.}
\label{tab:full15}
\begin{center}\small
\begin{tabular}{lccc}
\toprule
Condition & Accuracy & Difference & $p$ \\
\midrule
REAL & 81.35 & -- & -- \\
MEAN & 81.00 & $-$0.35 & 0.614 \\
SWAP & 80.15 & $-$1.20\sig{\dag} & 0.0566 \\
$\boldsymbol{\mu}+0\cdot(\mathbf{t}_i-\boldsymbol{\mu})$ & 80.30 & $-$1.05 & 0.119 \\
$\boldsymbol{\mu}+2(\mathbf{t}_i-\boldsymbol{\mu})$ & 82.05 & $+$0.70 & 0.338 \\
NODE (real rows) & 81.55 & $+$0.20 & 0.699 \\
NODE (Gaussian) & 81.80 & $+$0.45 & 0.272 \\
EDGE (rewired) & 81.55 & $+$0.20 & 0.793 \\
REL (resampled) & 81.20 & $-$0.15 & 0.876 \\
zero token & 11.30 & $-$70.05\sig{***} & $<10^{-300}$ \\
norm $\rightarrow$ word size & 11.35 & $-$70.00\sig{***} & $<10^{-300}$ \\
$\times0.5$ & 76.85 & $-$4.50\sig{***} & $1.53\times10^{-7}$ \\
$\times2$ & 80.55 & $-$0.80 & 0.221 \\
\midrule
NOTXT-REAL & 30.20 & -- & -- \\
NOTXT-MEAN & 30.65 & $+$0.45 & 0.686 \\
NOTXT-SWAP & 28.80 & $-$1.40 & 0.187 \\
NOTXT-NODE & 29.75 & $-$0.45 & 0.684 \\
NOTXT-zero & 12.50 & $-$17.70\sig{***} & $3.5\times10^{-45}$ \\
NOTXT-norm $\rightarrow$ word size & 11.60 & $-$18.60\sig{***} & $2.78\times10^{-52}$ \\
\bottomrule
\end{tabular}
\end{center}
\end{table}

\begin{table}[h]
\caption{\textbf{G-Retriever (GAT)}, 2000 test questions.}
\label{tab:full16}
\begin{center}\small
\begin{tabular}{lccc}
\toprule
Condition & Accuracy & Difference & $p$ \\
\midrule
REAL & 80.20 & -- & -- \\
MEAN & 80.80 & $+$0.60 & 0.213 \\
SWAP & 80.40 & $+$0.20 & 0.781 \\
$\boldsymbol{\mu}+0\cdot(\mathbf{t}_i-\boldsymbol{\mu})$ & 80.60 & $+$0.40 & 0.434 \\
$\boldsymbol{\mu}+2(\mathbf{t}_i-\boldsymbol{\mu})$ & 80.15 & $-$0.05 & 1 \\
NODE (real rows) & 80.00 & $-$0.20 & 0.757 \\
NODE (Gaussian) & 80.50 & $+$0.30 & 0.497 \\
EDGE (rewired) & 80.30 & $+$0.10 & 0.856 \\
REL (resampled) & 80.35 & $+$0.15 & 0.25 \\
zero token & 11.30 & $-$68.90\sig{***} & $<10^{-300}$ \\
norm $\rightarrow$ word size & 9.50 & $-$70.70\sig{***} & $<10^{-300}$ \\
$\times0.5$ & 78.90 & $-$1.30\sig{\dag} & 0.0724 \\
$\times2$ & 78.55 & $-$1.65\sig{*} & 0.0266 \\
\midrule
NOTXT-REAL & 28.65 & -- & -- \\
NOTXT-MEAN & 28.90 & $+$0.25 & 0.817 \\
NOTXT-SWAP & 28.20 & $-$0.45 & 0.638 \\
NOTXT-NODE & 29.80 & $+$1.15 & 0.224 \\
NOTXT-zero & 12.50 & $-$16.15\sig{***} & $5.22\times10^{-36}$ \\
NOTXT-norm $\rightarrow$ word size & 10.95 & $-$17.70\sig{***} & $3.5\times10^{-45}$ \\
\bottomrule
\end{tabular}
\end{center}
\end{table}

\clearpage

\begin{table}[h]
\caption{\textbf{G-Retriever (GCN)}, 2000 test questions.}
\label{tab:full17}
\begin{center}\small
\begin{tabular}{lccc}
\toprule
Condition & Accuracy & Difference & $p$ \\
\midrule
REAL & 80.55 & -- & -- \\
MEAN & 80.45 & $-$0.10 & 0.921 \\
SWAP & 79.50 & $-$1.05 & 0.137 \\
$\boldsymbol{\mu}+0\cdot(\mathbf{t}_i-\boldsymbol{\mu})$ & 80.35 & $-$0.20 & 0.752 \\
$\boldsymbol{\mu}+2(\mathbf{t}_i-\boldsymbol{\mu})$ & 79.00 & $-$1.55\sig{*} & 0.0178 \\
NODE (real rows) & 79.30 & $-$1.25\sig{\dag} & 0.0725 \\
NODE (Gaussian) & 79.60 & $-$0.95 & 0.181 \\
EDGE (rewired) & 79.30 & $-$1.25\sig{\dag} & 0.0661 \\
REL (resampled) & 80.55 & $+$0.00 & 1 \\
zero token & 11.30 & $-$69.25\sig{***} & $<10^{-300}$ \\
norm $\rightarrow$ word size & 11.80 & $-$68.75\sig{***} & $<10^{-300}$ \\
$\times0.5$ & 78.25 & $-$2.30\sig{**} & 0.00388 \\
$\times2$ & 78.90 & $-$1.65\sig{*} & 0.019 \\
\midrule
NOTXT-REAL & 29.80 & -- & -- \\
NOTXT-MEAN & 29.55 & $-$0.25 & 0.857 \\
NOTXT-SWAP & 28.85 & $-$0.95 & 0.42 \\
NOTXT-NODE & 30.90 & $+$1.10 & 0.334 \\
NOTXT-zero & 12.50 & $-$17.30\sig{***} & $2.73\times10^{-46}$ \\
NOTXT-norm $\rightarrow$ word size & 12.40 & $-$17.40\sig{***} & $5.23\times10^{-47}$ \\
\bottomrule
\end{tabular}
\end{center}
\end{table}

\begin{table}[h]
\caption{\textbf{CONST}, 2000 test questions.}
\label{tab:full18}
\begin{center}\small
\begin{tabular}{lccc}
\toprule
Condition & Accuracy & Difference & $p$ \\
\midrule
REAL & 81.40 & -- & -- \\
MEAN & 81.25 & $-$0.15 & 0.25 \\
SWAP & 81.40 & $+$0.00 & 1 \\
$\boldsymbol{\mu}+0\cdot(\mathbf{t}_i-\boldsymbol{\mu})$ & 81.25 & $-$0.15 & 0.25 \\
$\boldsymbol{\mu}+2(\mathbf{t}_i-\boldsymbol{\mu})$ & 80.95 & $-$0.45\sig{**} & 0.00391 \\
NODE (real rows) & 81.40 & $+$0.00 & 1 \\
NODE (Gaussian) & 81.40 & $+$0.00 & 1 \\
EDGE (rewired) & 81.40 & $+$0.00 & 1 \\
REL (resampled) & 81.40 & $+$0.00 & 1 \\
zero token & 12.20 & $-$69.20\sig{***} & $<10^{-300}$ \\
norm $\rightarrow$ word size & 14.80 & $-$66.60\sig{***} & $<10^{-300}$ \\
$\times0.5$ & 74.50 & $-$6.90\sig{***} & $2.29\times10^{-16}$ \\
$\times2$ & 79.45 & $-$1.95\sig{**} & 0.00483 \\
\midrule
NOTXT-REAL & 32.55 & -- & -- \\
NOTXT-MEAN & 32.40 & $-$0.15 & 0.755 \\
NOTXT-SWAP & 32.55 & $+$0.00 & 1 \\
NOTXT-NODE & 32.55 & $+$0.00 & 1 \\
NOTXT-zero & 13.80 & $-$18.75\sig{***} & $1.23\times10^{-53}$ \\
NOTXT-norm $\rightarrow$ word size & 12.35 & $-$20.20\sig{***} & $2.46\times10^{-63}$ \\
\bottomrule
\end{tabular}
\end{center}
\end{table}

\begin{table}[h]
\caption{\textbf{Stage 1 + GELU, projector only}, 2000 test questions.}
\label{tab:full19}
\begin{center}\small
\begin{tabular}{lccc}
\toprule
Condition & Accuracy & Difference & $p$ \\
\midrule
REAL & 69.15 & -- & -- \\
MEAN & 48.95 & $-$20.20\sig{***} & $6.8\times10^{-48}$ \\
SWAP & 34.10 & $-$35.05\sig{***} & $1.11\times10^{-127}$ \\
$\boldsymbol{\mu}+0\cdot(\mathbf{t}_i-\boldsymbol{\mu})$ & 49.05 & $-$20.10\sig{***} & $2.05\times10^{-47}$ \\
$\boldsymbol{\mu}+2(\mathbf{t}_i-\boldsymbol{\mu})$ & 38.85 & $-$30.30\sig{***} & $4.36\times10^{-116}$ \\
NODE (real rows) & 60.25 & $-$8.90\sig{***} & $7.43\times10^{-15}$ \\
NODE (Gaussian) & 60.05 & $-$9.10\sig{***} & $2.34\times10^{-16}$ \\
EDGE (rewired) & 66.70 & $-$2.45\sig{*} & 0.0159 \\
REL (resampled) & 68.30 & $-$0.85 & 0.244 \\
zero token & 11.30 & $-$57.85\sig{***} & $6.61\times10^{-293}$ \\
norm $\rightarrow$ word size & 15.50 & $-$53.65\sig{***} & $1.13\times10^{-262}$ \\
$\times0.5$ & 54.70 & $-$14.45\sig{***} & $1.12\times10^{-38}$ \\
$\times2$ & 64.65 & $-$4.50\sig{***} & $2.06\times10^{-7}$ \\
\midrule
NOTXT-REAL & 46.60 & -- & -- \\
NOTXT-MEAN & 22.95 & $-$23.65\sig{***} & $7.41\times10^{-69}$ \\
NOTXT-SWAP & 22.45 & $-$24.15\sig{***} & $1.66\times10^{-78}$ \\
NOTXT-NODE & 38.90 & $-$7.70\sig{***} & $1.31\times10^{-13}$ \\
NOTXT-zero & 12.50 & $-$34.10\sig{***} & $1.03\times10^{-146}$ \\
NOTXT-norm $\rightarrow$ word size & 12.80 & $-$33.80\sig{***} & $1.32\times10^{-143}$ \\
\bottomrule
\end{tabular}
\end{center}
\end{table}

\begin{table}[h]
\caption{\textbf{Ours, joint, standard recipe}, 2,000
\texttt{SceneGraphs} test questions. The reference is \textsc{real} above
the midrule and \textsc{notxt-real} below it.}
\label{tab:scene-standard-full}
\begin{center}
\small
\fitwidth{
\begin{tabular}{lccc}
\toprule
Condition & Accuracy & Difference & $p$ \\
\midrule
\textsc{real}
 & 72.50 & -- & -- \\
\textsc{mean}
 & 71.10 & $-1.40$ & .182 \\
\textsc{swap}
 & 69.10 & $-3.40$\sig{***} & $6.31\times10^{-4}$ \\
\textsc{swap} (permutation 2)
 & 67.85 & $-4.65$\sig{***} & $2.32\times10^{-6}$ \\
\textsc{swap} (permutation 3)
 & 67.85 & $-4.65$\sig{***} & $1.21\times10^{-5}$ \\
$\boldsymbol{\mu}+0(\mathbf{t}_i-\boldsymbol{\mu})$
 & 70.50 & $-2.00$\sig{*} & .0481 \\
\textsc{node} (real rows)
 & 67.40 & $-5.10$\sig{***} & $1.33\times10^{-10}$ \\
\textsc{node} (Gaussian)
 & 67.40 & $-5.10$\sig{***} & $1.57\times10^{-11}$ \\
\textsc{edge} (rewired)
 & 71.45 & $-1.05$ & .186 \\
\textsc{gswap} (other graph)
 & 67.20 & $-5.30$\sig{***} & $3.55\times10^{-9}$ \\
\midrule
\textsc{notxt-real}
 & 32.60 & -- & -- \\
\textsc{notxt-mean}
 & 26.00 & $-6.60$\sig{***} & $3.57\times10^{-8}$ \\
\textsc{notxt-swap}
 & 21.75 & $-10.85$\sig{***} & $8.93\times10^{-19}$ \\
\textsc{notxt-gswap}
 & 29.50 & $-3.10$\sig{**} & .00600 \\
\bottomrule
\end{tabular}}
\end{center}
\end{table}

\begin{table}[h]
\caption{\textbf{G-Retriever, half split}, 2000 test questions.}
\label{tab:full20}
\begin{center}\small
\begin{tabular}{lccc}
\toprule
Condition & Accuracy & Difference & $p$ \\
\midrule
REAL & 77.40 & -- & -- \\
MEAN & 77.30 & $-$0.10 & 0.918 \\
SWAP & 77.20 & $-$0.20 & 0.801 \\
$\boldsymbol{\mu}+0\cdot(\mathbf{t}_i-\boldsymbol{\mu})$ & 77.35 & $-$0.05 & 1 \\
NODE (real rows) & 77.50 & $+$0.10 & 0.856 \\
NODE (Gaussian) & 77.35 & $-$0.05 & 1 \\
EDGE (rewired) & 77.20 & $-$0.20 & 0.775 \\
\midrule
NOTXT-REAL & 25.50 & -- & -- \\
NOTXT-MEAN & 24.90 & $-$0.60 & 0.5 \\
NOTXT-SWAP & 24.40 & $-$1.10 & 0.199 \\
\bottomrule
\end{tabular}
\end{center}
\end{table}

\clearpage

\begin{table}[h]
\caption{\textbf{Ours, joint, cross-fitted, initialisation 1}, 2000 test questions, cross-fitted recipe.}
\label{tab:full21}
\begin{center}\small
\begin{tabular}{lccc}
\toprule
Condition & Accuracy & Difference & $p$ \\
\midrule
REAL & 76.90 & -- & -- \\
MEAN & 75.40 & $-$1.50\sig{\dag} & 0.0829 \\
SWAP & 72.45 & $-$4.45\sig{***} & $2.02\times10^{-7}$ \\
$\boldsymbol{\mu}+0\cdot(\mathbf{t}_i-\boldsymbol{\mu})$ & 74.95 & $-$1.95\sig{*} & 0.0242 \\
NODE (real rows) & 75.85 & $-$1.05 & 0.158 \\
NODE (Gaussian) & 76.10 & $-$0.80 & 0.323 \\
EDGE (rewired) & 74.95 & $-$1.95\sig{**} & 0.00202 \\
\midrule
NOTXT-REAL & 26.20 & -- & -- \\
NOTXT-MEAN & 25.45 & $-$0.75 & 0.474 \\
NOTXT-SWAP & 23.60 & $-$2.60\sig{*} & 0.0123 \\
\bottomrule
\end{tabular}
\end{center}
\end{table}

\begin{table}[h]
\caption{\textbf{Ours, joint, cross-fitted, initialisation 2}, 2000 test questions, cross-fitted recipe.}
\label{tab:full22}
\begin{center}\small
\begin{tabular}{lccc}
\toprule
Condition & Accuracy & Difference & $p$ \\
\midrule
REAL & 77.55 & -- & -- \\
MEAN & 76.45 & $-$1.10 & 0.191 \\
SWAP & 75.10 & $-$2.45\sig{**} & 0.00146 \\
$\boldsymbol{\mu}+0\cdot(\mathbf{t}_i-\boldsymbol{\mu})$ & 76.50 & $-$1.05 & 0.214 \\
NODE (real rows) & 78.55 & $+$1.00 & 0.116 \\
NODE (Gaussian) & 78.15 & $+$0.60 & 0.366 \\
EDGE (rewired) & 77.85 & $+$0.30 & 0.734 \\
\midrule
NOTXT-REAL & 31.25 & -- & -- \\
NOTXT-MEAN & 33.00 & $+$1.75 & 0.13 \\
NOTXT-SWAP & 29.55 & $-$1.70 & 0.136 \\
\bottomrule
\end{tabular}
\end{center}
\end{table}

\begin{table}[h]
\caption{\textbf{Ours, no entity features}, 2000 test questions. Best epoch 3 of 5, with validation
loss 0.326, 0.287, 0.283, 0.2826, 0.302 over epochs 0 to 4.}
\label{tab:full48}
\begin{center}\small
\begin{tabular}{lccc}
\toprule
Condition & Accuracy & Difference & $p$ \\
\midrule
REAL & 79.65 & -- & -- \\
MEAN & 73.90 & $-$5.75\sig{***} & $2.1\times10^{-11}$ \\
SWAP & 70.05 & $-$9.60\sig{***} & $1.2\times10^{-23}$ \\
NODE (real rows) & 79.75 & $+$0.10 & 0.73 \\
EDGE (rewired) & 80.05 & $+$0.40 & 0.56 \\
GSWAP (other graph) & 80.00 & $+$0.35 & 0.54 \\
\midrule
NOTXT-REAL & 30.10 & -- & -- \\
NOTXT-MEAN & 17.25 & $-$12.85\sig{***} & $2.0\times10^{-35}$ \\
\bottomrule
\end{tabular}
\end{center}
\end{table}

\textsc{swap} on our standard-recipe \texttt{SceneGraphs} model is not an artefact of one pairing:
two further derangements give $-4.65$ ($p=2.3\times10^{-6}$) and $-4.65$ ($p=1.2\times10^{-5}$) on
the same 2,000 questions, against the $-3.40$ reported in Table~\ref{tab:geometry-use}, which is the
weakest of the three.

\subsection{\texttt{WebQSP}}

\begin{table}[h]
\caption{\textbf{G-Retriever (graph transformer)}, 1628 test questions.}
\label{tab:full23}
\begin{center}\small
\begin{tabular}{lccc}
\toprule
Condition & Accuracy & Difference & $p$ \\
\midrule
REAL & 70.33 & -- & -- \\
MEAN & 68.49 & $-$1.84\sig{*} & 0.0285 \\
SWAP & 69.41 & $-$0.92 & 0.311 \\
$\boldsymbol{\mu}+0\cdot(\mathbf{t}_i-\boldsymbol{\mu})$ & 70.70 & $+$0.37 & 0.687 \\
$\boldsymbol{\mu}+2(\mathbf{t}_i-\boldsymbol{\mu})$ & 70.95 & $+$0.61 & 0.495 \\
NODE (real rows) & 70.39 & $+$0.06 & 1 \\
NODE (Gaussian) & 70.21 & $-$0.12 & 0.933 \\
EDGE (rewired) & 71.01 & $+$0.68 & 0.46 \\
REL (resampled) & 70.82 & $+$0.49 & 0.575 \\
GSWAP (other graph) & 69.78 & $-$0.55 & 0.569 \\
SWAP (permutation 2) & 69.96 & $-$0.37 & 0.711 \\
SWAP (permutation 3) & 70.15 & $-$0.18 & 0.874 \\
zero token & 10.69 & $-$59.64\sig{***} & $1.59\times10^{-266}$ \\
norm $\rightarrow$ word size & 4.85 & $-$65.48\sig{***} & $<10^{-300}$ \\
$\times0.5$ & 63.02 & $-$7.31\sig{***} & $1.9\times10^{-12}$ \\
$\times2$ & 66.09 & $-$4.24\sig{***} & $7.26\times10^{-6}$ \\
\midrule
NOTXT-REAL & 41.34 & -- & -- \\
NOTXT-MEAN & 42.38 & $+$1.04 & 0.187 \\
NOTXT-SWAP & 41.15 & $-$0.18 & 0.886 \\
NOTXT-NODE & 41.28 & $-$0.06 & 1 \\
NOTXT-GSWAP & 40.48 & $-$0.86 & 0.356 \\
NOTXT-zero & 4.05 & $-$37.29\sig{***} & $1.79\times10^{-157}$ \\
NOTXT-norm $\rightarrow$ word size & 2.76 & $-$38.57\sig{***} & $1.56\times10^{-170}$ \\
\bottomrule
\end{tabular}
\end{center}
\end{table}

\begin{table}[h]
\caption{\textbf{G-Retriever + LoRA}, 1628 test questions.}
\label{tab:full24}
\begin{center}\small
\begin{tabular}{lccc}
\toprule
Condition & Accuracy & Difference & $p$ \\
\midrule
REAL & 74.32 & -- & -- \\
MEAN & 73.89 & $-$0.43 & 0.634 \\
SWAP & 73.59 & $-$0.74 & 0.382 \\
$\boldsymbol{\mu}+0\cdot(\mathbf{t}_i-\boldsymbol{\mu})$ & 73.89 & $-$0.43 & 0.636 \\
NODE (real rows) & 73.34 & $-$0.98 & 0.224 \\
NODE (Gaussian) & 73.59 & $-$0.74 & 0.359 \\
EDGE (rewired) & 74.20 & $-$0.12 & 0.935 \\
\midrule
NOTXT-REAL & 10.38 & -- & -- \\
NOTXT-MEAN & 11.49 & $+$1.11 & 0.215 \\
NOTXT-SWAP & 11.43 & $+$1.04 & 0.221 \\
\bottomrule
\end{tabular}
\end{center}
\end{table}

\clearpage

\begin{table}[h]
\caption{\textbf{G-Retriever with Qwen2.5-3B}, 1628 test questions.}
\label{tab:full25}
\begin{center}\small
\begin{tabular}{lccc}
\toprule
Condition & Accuracy & Difference & $p$ \\
\midrule
REAL & 55.47 & -- & -- \\
MEAN & 54.98 & $-$0.49 & 0.48 \\
SWAP & 54.79 & $-$0.68 & 0.428 \\
$\boldsymbol{\mu}+0\cdot(\mathbf{t}_i-\boldsymbol{\mu})$ & 54.98 & $-$0.49 & 0.48 \\
NODE (real rows) & 54.85 & $-$0.61 & 0.378 \\
NODE (Gaussian) & 54.67 & $-$0.80 & 0.232 \\
EDGE (rewired) & 55.41 & $-$0.06 & 1 \\
\midrule
NOTXT-REAL & 34.40 & -- & -- \\
NOTXT-MEAN & 30.65 & $-$3.75\sig{***} & $2.45\times10^{-5}$ \\
NOTXT-SWAP & 32.31 & $-$2.09\sig{*} & 0.0286 \\
\bottomrule
\end{tabular}
\end{center}
\end{table}

\begin{table}[h]
\caption{\textbf{Stage 1 + GELU, projector only}, 1628 test questions.}
\label{tab:full26}
\begin{center}\small
\begin{tabular}{lccc}
\toprule
Condition & Accuracy & Difference & $p$ \\
\midrule
REAL & 64.62 & -- & -- \\
MEAN & 49.88 & $-$14.74\sig{***} & $1.26\times10^{-30}$ \\
SWAP & 54.73 & $-$9.89\sig{***} & $4.62\times10^{-15}$ \\
$\boldsymbol{\mu}+0\cdot(\mathbf{t}_i-\boldsymbol{\mu})$ & 49.88 & $-$14.74\sig{***} & $1.26\times10^{-30}$ \\
$\boldsymbol{\mu}+2(\mathbf{t}_i-\boldsymbol{\mu})$ & 37.65 & $-$26.97\sig{***} & $8.63\times10^{-82}$ \\
NODE (real rows) & 62.22 & $-$2.40\sig{*} & 0.0289 \\
NODE (Gaussian) & 63.08 & $-$1.54 & 0.173 \\
EDGE (rewired) & 64.68 & $+$0.06 & 1 \\
REL (resampled) & 63.82 & $-$0.80 & 0.424 \\
zero token & 10.69 & $-$53.93\sig{***} & $1.84\times10^{-224}$ \\
norm $\rightarrow$ word size & 14.37 & $-$50.25\sig{***} & $3.14\times10^{-203}$ \\
$\times0.5$ & 54.05 & $-$10.57\sig{***} & $1.04\times10^{-18}$ \\
$\times2$ & 54.30 & $-$10.32\sig{***} & $4.88\times10^{-20}$ \\
\midrule
NOTXT-REAL & 36.30 & -- & -- \\
NOTXT-MEAN & 16.15 & $-$20.15\sig{***} & $1.76\times10^{-53}$ \\
NOTXT-SWAP & 28.01 & $-$8.29\sig{***} & $8.71\times10^{-12}$ \\
NOTXT-NODE & 34.28 & $-$2.03\sig{\dag} & 0.0911 \\
NOTXT-zero & 4.05 & $-$32.25\sig{***} & $1.22\times10^{-136}$ \\
NOTXT-norm $\rightarrow$ word size & 4.12 & $-$32.19\sig{***} & $3.3\times10^{-139}$ \\
\bottomrule
\end{tabular}
\end{center}
\end{table}

\begin{table}[h]
\caption{\textbf{Ours, joint (the method)}, 1628 test questions.}
\label{tab:full27}
\begin{center}\small
\begin{tabular}{lccc}
\toprule
Condition & Accuracy & Difference & $p$ \\
\midrule
REAL & 69.90 & -- & -- \\
MEAN & 66.77 & $-$3.13\sig{***} & 0.000865 \\
SWAP & 67.75 & $-$2.15\sig{*} & 0.0387 \\
$\boldsymbol{\mu}+0\cdot(\mathbf{t}_i-\boldsymbol{\mu})$ & 67.38 & $-$2.52\sig{**} & 0.00807 \\
NODE (real rows) & 68.67 & $-$1.23 & 0.216 \\
NODE (Gaussian) & 68.55 & $-$1.35 & 0.157 \\
EDGE (rewired) & 69.10 & $-$0.80 & 0.415 \\
GSWAP (other graph) & 67.94 & $-$1.97\sig{\dag} & 0.0515 \\
SWAP (permutation 2) & 67.75 & $-$2.15\sig{*} & 0.0344 \\
SWAP (permutation 3) & 66.46 & $-$3.44\sig{***} & 0.000785 \\
\midrule
NOTXT-REAL & 28.75 & -- & -- \\
NOTXT-MEAN & 26.66 & $-$2.09\sig{*} & 0.0452 \\
NOTXT-SWAP & 26.35 & $-$2.40\sig{*} & 0.0283 \\
NOTXT-GSWAP & 26.47 & $-$2.27\sig{*} & 0.0372 \\
\bottomrule
\end{tabular}
\end{center}
\end{table}

\begin{table}[h]
\caption{\textbf{Ours, joint + LoRA}, 1628 test questions.}
\label{tab:full28}
\begin{center}\small
\begin{tabular}{lccc}
\toprule
Condition & Accuracy & Difference & $p$ \\
\midrule
REAL & 74.57 & -- & -- \\
MEAN & 73.46 & $-$1.11 & 0.195 \\
SWAP & 72.24 & $-$2.33\sig{**} & 0.00839 \\
$\boldsymbol{\mu}+0\cdot(\mathbf{t}_i-\boldsymbol{\mu})$ & 72.60 & $-$1.97\sig{*} & 0.0192 \\
NODE (real rows) & 73.77 & $-$0.80 & 0.332 \\
NODE (Gaussian) & 73.40 & $-$1.17 & 0.161 \\
EDGE (rewired) & 74.82 & $+$0.25 & 0.811 \\
\midrule
NOTXT-REAL & 47.91 & -- & -- \\
NOTXT-MEAN & 49.20 & $+$1.29 & 0.239 \\
NOTXT-SWAP & 44.72 & $-$3.19\sig{**} & 0.00518 \\
\bottomrule
\end{tabular}
\end{center}
\end{table}

\clearpage

\begin{table}[h]
\caption{\textbf{CONST}, 1628 test questions.}
\label{tab:full29}
\begin{center}\small
\begin{tabular}{lccc}
\toprule
Condition & Accuracy & Difference & $p$ \\
\midrule
REAL & 71.25 & -- & -- \\
MEAN & 71.44 & $+$0.18 & 0.69 \\
SWAP & 71.25 & $+$0.00 & 1 \\
$\boldsymbol{\mu}+0\cdot(\mathbf{t}_i-\boldsymbol{\mu})$ & 71.44 & $+$0.18 & 0.69 \\
NODE (real rows) & 71.25 & $+$0.00 & 1 \\
NODE (Gaussian) & 71.25 & $+$0.00 & 1 \\
EDGE (rewired) & 71.25 & $+$0.00 & 1 \\
\midrule
NOTXT-REAL & 42.57 & -- & -- \\
NOTXT-MEAN & 42.14 & $-$0.43 & 0.189 \\
NOTXT-SWAP & 42.57 & $+$0.00 & 1 \\
\bottomrule
\end{tabular}
\end{center}
\end{table}

\begin{table}[h]
\caption{\textbf{MLPQ}, 1628 test questions.}
\label{tab:full30}
\begin{center}\small
\begin{tabular}{lccc}
\toprule
Condition & Accuracy & Difference & $p$ \\
\midrule
REAL & 70.70 & -- & -- \\
MEAN & 70.21 & $-$0.49 & 0.115 \\
SWAP & 68.98 & $-$1.72\sig{*} & 0.015 \\
$\boldsymbol{\mu}+0\cdot(\mathbf{t}_i-\boldsymbol{\mu})$ & 70.21 & $-$0.49 & 0.115 \\
NODE (real rows) & 70.70 & $+$0.00 & 1 \\
NODE (Gaussian) & 70.70 & $+$0.00 & 1 \\
EDGE (rewired) & 70.70 & $+$0.00 & 1 \\
\midrule
NOTXT-REAL & 43.37 & -- & -- \\
NOTXT-MEAN & 43.61 & $+$0.25 & 0.608 \\
NOTXT-SWAP & 44.04 & $+$0.68\sig{\dag} & 0.0801 \\
\bottomrule
\end{tabular}
\end{center}
\end{table}

\begin{table}[h]
\caption{\textbf{Ours, joint, Qwen2.5-3B}, 1628 test questions.}
\label{tab:full31}
\begin{center}\small
\begin{tabular}{lccc}
\toprule
Condition & Accuracy & Difference & $p$ \\
\midrule
REAL & 65.79 & -- & -- \\
MEAN & 63.70 & $-$2.09\sig{**} & 0.00187 \\
SWAP & 63.88 & $-$1.90\sig{**} & 0.00573 \\
$\boldsymbol{\mu}+0\cdot(\mathbf{t}_i-\boldsymbol{\mu})$ & 63.70 & $-$2.09\sig{**} & 0.00187 \\
NODE (real rows) & 64.99 & $-$0.80 & 0.25 \\
NODE (Gaussian) & 65.17 & $-$0.61 & 0.373 \\
EDGE (rewired) & 64.43 & $-$1.35\sig{*} & 0.0315 \\
\midrule
NOTXT-REAL & 34.34 & -- & -- \\
NOTXT-MEAN & 34.64 & $+$0.31 & 0.768 \\
NOTXT-SWAP & 35.44 & $+$1.11 & 0.217 \\
\bottomrule
\end{tabular}
\end{center}
\end{table}

\begin{table}[h]
\caption{\textbf{G-Retriever, Qwen2.5-3B, matched budget}, 1628 test questions.}
\label{tab:full32}
\begin{center}\small
\begin{tabular}{lccc}
\toprule
Condition & Accuracy & Difference & $p$ \\
\midrule
REAL & 57.13 & -- & -- \\
MEAN & 56.57 & $-$0.55 & 0.478 \\
SWAP & 55.71 & $-$1.41 & 0.104 \\
$\boldsymbol{\mu}+0\cdot(\mathbf{t}_i-\boldsymbol{\mu})$ & 56.70 & $-$0.43 & 0.595 \\
NODE (real rows) & 56.82 & $-$0.31 & 0.704 \\
NODE (Gaussian) & 56.57 & $-$0.55 & 0.452 \\
EDGE (rewired) & 57.19 & $+$0.06 & 1 \\
\midrule
NOTXT-REAL & 29.30 & -- & -- \\
NOTXT-MEAN & 30.16 & $+$0.86 & 0.403 \\
NOTXT-SWAP & 27.27 & $-$2.03\sig{\dag} & 0.0508 \\
\bottomrule
\end{tabular}
\end{center}
\end{table}

\clearpage

\begin{table}[h]
\caption{\textbf{G-Retriever, half split, initialisation 1}, 1628 test questions.}
\label{tab:full33}
\begin{center}\small
\begin{tabular}{lccc}
\toprule
Condition & Accuracy & Difference & $p$ \\
\midrule
REAL & 70.58 & -- & -- \\
MEAN & 70.02 & $-$0.55 & 0.538 \\
SWAP & 70.15 & $-$0.43 & 0.654 \\
$\boldsymbol{\mu}+0\cdot(\mathbf{t}_i-\boldsymbol{\mu})$ & 69.90 & $-$0.68 & 0.46 \\
NODE (real rows) & 70.33 & $-$0.25 & 0.784 \\
NODE (Gaussian) & 70.58 & $+$0.00 & 1 \\
EDGE (rewired) & 70.64 & $+$0.06 & 1 \\
\midrule
NOTXT-REAL & 42.14 & -- & -- \\
NOTXT-MEAN & 43.06 & $+$0.92 & 0.115 \\
NOTXT-SWAP & 41.77 & $-$0.37 & 0.637 \\
\bottomrule
\end{tabular}
\end{center}
\end{table}

\begin{table}[h]
\caption{\textbf{G-Retriever, half split, initialisation 2}, 1628 test questions.}
\label{tab:full34}
\begin{center}\small
\begin{tabular}{lccc}
\toprule
Condition & Accuracy & Difference & $p$ \\
\midrule
REAL & 67.75 & -- & -- \\
MEAN & 67.69 & $-$0.06 & 1 \\
SWAP & 67.26 & $-$0.49 & 0.614 \\
$\boldsymbol{\mu}+0\cdot(\mathbf{t}_i-\boldsymbol{\mu})$ & 67.57 & $-$0.18 & 0.89 \\
NODE (real rows) & 67.26 & $-$0.49 & 0.598 \\
NODE (Gaussian) & 67.63 & $-$0.12 & 0.942 \\
EDGE (rewired) & 67.20 & $-$0.55 & 0.573 \\
\midrule
NOTXT-REAL & 37.96 & -- & -- \\
NOTXT-MEAN & 38.39 & $+$0.43 & 0.562 \\
NOTXT-SWAP & 37.29 & $-$0.68 & 0.4 \\
\bottomrule
\end{tabular}
\end{center}
\end{table}

\begin{table}[h]
\caption{\textbf{G-Retriever, half split, initialisation 3}, 1628 test questions.}
\label{tab:full35}
\begin{center}\small
\begin{tabular}{lccc}
\toprule
Condition & Accuracy & Difference & $p$ \\
\midrule
REAL & 70.02 & -- & -- \\
MEAN & 70.02 & $+$0.00 & 1 \\
SWAP & 69.78 & $-$0.25 & 0.836 \\
$\boldsymbol{\mu}+0\cdot(\mathbf{t}_i-\boldsymbol{\mu})$ & 68.86 & $-$1.17 & 0.22 \\
NODE (real rows) & 69.16 & $-$0.86 & 0.356 \\
NODE (Gaussian) & 70.52 & $+$0.49 & 0.617 \\
EDGE (rewired) & 69.66 & $-$0.37 & 0.718 \\
\midrule
NOTXT-REAL & 34.09 & -- & -- \\
NOTXT-MEAN & 34.89 & $+$0.80 & 0.275 \\
NOTXT-SWAP & 35.01 & $+$0.92 & 0.248 \\
\bottomrule
\end{tabular}
\end{center}
\end{table}

\begin{table}[h]
\caption{\textbf{Projector only, cross-fitted}, 1628 test questions, cross-fitted recipe.}
\label{tab:full36}
\begin{center}\small
\begin{tabular}{lccc}
\toprule
Condition & Accuracy & Difference & $p$ \\
\midrule
REAL & 63.14 & -- & -- \\
MEAN & 59.71 & $-$3.44\sig{**} & 0.00341 \\
SWAP & 57.80 & $-$5.34\sig{***} & $6.9\times10^{-6}$ \\
$\boldsymbol{\mu}+0\cdot(\mathbf{t}_i-\boldsymbol{\mu})$ & 60.01 & $-$3.13\sig{**} & 0.0045 \\
NODE (real rows) & 61.49 & $-$1.66 & 0.127 \\
NODE (Gaussian) & 62.16 & $-$0.98 & 0.385 \\
EDGE (rewired) & 64.19 & $+$1.04 & 0.32 \\
\midrule
NOTXT-REAL & 36.61 & -- & -- \\
NOTXT-MEAN & 41.58 & $+$4.98\sig{***} & $1.71\times10^{-5}$ \\
NOTXT-SWAP & 33.35 & $-$3.26\sig{**} & 0.00517 \\
\bottomrule
\end{tabular}
\end{center}
\end{table}

\clearpage

\begin{table}[h]
\caption{\textbf{Late-prompt variant, cross-fitted}, 1628 test questions, cross-fitted recipe.}
\label{tab:full37}
\begin{center}\small
\begin{tabular}{lccc}
\toprule
Condition & Accuracy & Difference & $p$ \\
\midrule
REAL & 63.64 & -- & -- \\
MEAN & 60.69 & $-$2.95\sig{**} & 0.0075 \\
SWAP & 58.42 & $-$5.22\sig{***} & $1.41\times10^{-6}$ \\
$\boldsymbol{\mu}+0\cdot(\mathbf{t}_i-\boldsymbol{\mu})$ & 59.28 & $-$4.36\sig{***} & $7.12\times10^{-5}$ \\
NODE (real rows) & 60.87 & $-$2.76\sig{**} & 0.00953 \\
NODE (Gaussian) & 62.59 & $-$1.04 & 0.324 \\
EDGE (rewired) & 62.59 & $-$1.04 & 0.327 \\
\midrule
NOTXT-REAL & 38.33 & -- & -- \\
NOTXT-MEAN & 40.36 & $+$2.03\sig{\dag} & 0.0812 \\
NOTXT-SWAP & 34.77 & $-$3.56\sig{**} & 0.0015 \\
\bottomrule
\end{tabular}
\end{center}
\end{table}

\begin{table}[h]
\caption{\textbf{Projector dropout 0.5, cross-fitted}, 1628 test questions, cross-fitted recipe.}
\label{tab:full38}
\begin{center}\small
\begin{tabular}{lccc}
\toprule
Condition & Accuracy & Difference & $p$ \\
\midrule
REAL & 64.56 & -- & -- \\
MEAN & 60.32 & $-$4.24\sig{***} & $5.2\times10^{-5}$ \\
SWAP & 60.87 & $-$3.69\sig{***} & 0.000608 \\
$\boldsymbol{\mu}+0\cdot(\mathbf{t}_i-\boldsymbol{\mu})$ & 61.86 & $-$2.70\sig{**} & 0.00801 \\
NODE (real rows) & 63.27 & $-$1.29 & 0.216 \\
NODE (Gaussian) & 63.21 & $-$1.35 & 0.196 \\
EDGE (rewired) & 63.45 & $-$1.11 & 0.262 \\
\midrule
NOTXT-REAL & 41.83 & -- & -- \\
NOTXT-MEAN & 43.92 & $+$2.09\sig{*} & 0.0413 \\
NOTXT-SWAP & 38.88 & $-$2.95\sig{**} & 0.00638 \\
\bottomrule
\end{tabular}
\end{center}
\end{table}

\begin{table}[h]
\caption{\textbf{Text-dropout variant, cross-fitted}, 1628 test questions, cross-fitted recipe.}
\label{tab:full39}
\begin{center}\small
\begin{tabular}{lccc}
\toprule
Condition & Accuracy & Difference & $p$ \\
\midrule
REAL & 67.20 & -- & -- \\
MEAN & 68.55 & $+$1.35 & 0.143 \\
SWAP & 66.89 & $-$0.31 & 0.794 \\
$\boldsymbol{\mu}+0\cdot(\mathbf{t}_i-\boldsymbol{\mu})$ & 68.92 & $+$1.72\sig{\dag} & 0.0511 \\
NODE (real rows) & 69.23 & $+$2.03\sig{*} & 0.0281 \\
NODE (Gaussian) & 68.49 & $+$1.29 & 0.172 \\
EDGE (rewired) & 69.10 & $+$1.90\sig{*} & 0.035 \\
\midrule
NOTXT-REAL & 26.97 & -- & -- \\
NOTXT-MEAN & 31.27 & $+$4.30\sig{***} & 0.00043 \\
NOTXT-SWAP & 25.12 & $-$1.84 & 0.132 \\
\bottomrule
\end{tabular}
\end{center}
\end{table}

\begin{table}[h]
\caption{\textbf{Ours, joint, cross-fitted, initialisation 1}, 1628 test questions, cross-fitted recipe.}
\label{tab:full40}
\begin{center}\small
\begin{tabular}{lccc}
\toprule
Condition & Accuracy & Difference & $p$ \\
\midrule
REAL & 69.53 & -- & -- \\
MEAN & 68.00 & $-$1.54\sig{\dag} & 0.0967 \\
SWAP & 66.34 & $-$3.19\sig{**} & 0.00151 \\
$\boldsymbol{\mu}+0\cdot(\mathbf{t}_i-\boldsymbol{\mu})$ & 67.87 & $-$1.66\sig{\dag} & 0.0732 \\
NODE (real rows) & 66.28 & $-$3.26\sig{***} & 0.000443 \\
NODE (Gaussian) & 68.24 & $-$1.29 & 0.174 \\
EDGE (rewired) & 68.98 & $-$0.55 & 0.567 \\
GSWAP (other graph) & 66.71 & $-$2.83\sig{**} & 0.00499 \\
SWAP (permutation 2) & 67.63 & $-$1.90\sig{\dag} & 0.0581 \\
SWAP (permutation 3) & 66.46 & $-$3.07\sig{**} & 0.0018 \\
\midrule
NOTXT-REAL & 19.41 & -- & -- \\
NOTXT-MEAN & 21.99 & $+$2.58\sig{*} & 0.0252 \\
NOTXT-SWAP & 17.87 & $-$1.54 & 0.179 \\
NOTXT-GSWAP & 19.04 & $-$0.37 & 0.776 \\
\bottomrule
\end{tabular}
\end{center}
\end{table}

\clearpage

\begin{table}[h]
\caption{\textbf{Ours, joint, cross-fitted, initialisation 2}, 1628 test questions, cross-fitted recipe.}
\label{tab:full41}
\begin{center}\small
\begin{tabular}{lccc}
\toprule
Condition & Accuracy & Difference & $p$ \\
\midrule
REAL & 66.89 & -- & -- \\
MEAN & 65.11 & $-$1.78\sig{\dag} & 0.0888 \\
SWAP & 65.85 & $-$1.04 & 0.322 \\
$\boldsymbol{\mu}+0\cdot(\mathbf{t}_i-\boldsymbol{\mu})$ & 63.70 & $-$3.19\sig{**} & 0.00139 \\
NODE (real rows) & 64.68 & $-$2.21\sig{*} & 0.0298 \\
NODE (Gaussian) & 65.29 & $-$1.60 & 0.101 \\
EDGE (rewired) & 65.36 & $-$1.54 & 0.12 \\
\midrule
NOTXT-REAL & 22.91 & -- & -- \\
NOTXT-MEAN & 9.34 & $-$13.57\sig{***} & $4.3\times10^{-35}$ \\
NOTXT-SWAP & 21.13 & $-$1.78 & 0.115 \\
\bottomrule
\end{tabular}
\end{center}
\end{table}

\begin{table}[h]
\caption{\textbf{Ours, joint, cross-fitted, initialisation 3}, 1628 test questions, cross-fitted recipe.}
\label{tab:full42}
\begin{center}\small
\begin{tabular}{lccc}
\toprule
Condition & Accuracy & Difference & $p$ \\
\midrule
REAL & 68.43 & -- & -- \\
MEAN & 68.18 & $-$0.25 & 0.848 \\
SWAP & 68.30 & $-$0.12 & 0.948 \\
$\boldsymbol{\mu}+0\cdot(\mathbf{t}_i-\boldsymbol{\mu})$ & 68.73 & $+$0.31 & 0.797 \\
NODE (real rows) & 68.37 & $-$0.06 & 1 \\
NODE (Gaussian) & 68.80 & $+$0.37 & 0.737 \\
EDGE (rewired) & 68.92 & $+$0.49 & 0.598 \\
\midrule
NOTXT-REAL & 28.01 & -- & -- \\
NOTXT-MEAN & 24.69 & $-$3.32\sig{***} & 0.000237 \\
NOTXT-SWAP & 25.86 & $-$2.15\sig{*} & 0.0163 \\
\bottomrule
\end{tabular}
\end{center}
\end{table}

\begin{table}[h]
\caption{\textbf{Ours, joint + LoRA, cross-fitted}, 1628 test questions, cross-fitted recipe.}
\label{tab:full43}
\begin{center}\small
\begin{tabular}{lccc}
\toprule
Condition & Accuracy & Difference & $p$ \\
\midrule
REAL & 70.95 & -- & -- \\
MEAN & 69.78 & $-$1.17 & 0.166 \\
SWAP & 70.02 & $-$0.92 & 0.279 \\
$\boldsymbol{\mu}+0\cdot(\mathbf{t}_i-\boldsymbol{\mu})$ & 69.53 & $-$1.41\sig{\dag} & 0.0845 \\
NODE (real rows) & 69.47 & $-$1.47\sig{\dag} & 0.0669 \\
NODE (Gaussian) & 69.59 & $-$1.35\sig{\dag} & 0.0945 \\
EDGE (rewired) & 69.29 & $-$1.66\sig{*} & 0.0328 \\
\midrule
NOTXT-REAL & 41.09 & -- & -- \\
NOTXT-MEAN & 41.34 & $+$0.25 & 0.845 \\
NOTXT-SWAP & 39.62 & $-$1.47 & 0.161 \\
\bottomrule
\end{tabular}
\end{center}
\end{table}

\begin{table}[h]
\caption{\textbf{G-Retriever + LoRA, cross-fitted}, 1628 test questions, cross-fitted recipe.}
\label{tab:full44}
\begin{center}\small
\begin{tabular}{lccc}
\toprule
Condition & Accuracy & Difference & $p$ \\
\midrule
REAL & 72.17 & -- & -- \\
MEAN & 71.93 & $-$0.25 & 0.805 \\
SWAP & 71.99 & $-$0.18 & 0.859 \\
$\boldsymbol{\mu}+0\cdot(\mathbf{t}_i-\boldsymbol{\mu})$ & 70.82 & $-$1.35\sig{\dag} & 0.0882 \\
NODE (real rows) & 71.87 & $-$0.31 & 0.738 \\
NODE (Gaussian) & 71.56 & $-$0.61 & 0.456 \\
EDGE (rewired) & 71.19 & $-$0.98 & 0.221 \\
\midrule
NOTXT-REAL & 20.64 & -- & -- \\
NOTXT-MEAN & 21.99 & $+$1.35\sig{*} & 0.0448 \\
NOTXT-SWAP & 21.74 & $+$1.11 & 0.181 \\
\bottomrule
\end{tabular}
\end{center}
\end{table}

\clearpage

\begin{table}[h]
\caption{\textbf{Ours, no entity features}, 1628 test questions.}
\label{tab:full45}
\begin{center}\small
\begin{tabular}{lccc}
\toprule
Condition & Accuracy & Difference & $p$ \\
\midrule
REAL & 72.73 & -- & -- \\
MEAN & 66.58 & $-$6.14\sig{***} & $1.28\times10^{-9}$ \\
SWAP & 67.08 & $-$5.65\sig{***} & $5.88\times10^{-9}$ \\
NODE (real rows) & 72.11 & $-$0.61 & 0.415 \\
EDGE (rewired) & 72.30 & $-$0.43 & 0.664 \\
GSWAP (other graph) & 70.76 & $-$1.97\sig{*} & 0.0347 \\
\midrule
NOTXT-REAL & 28.38 & -- & -- \\
NOTXT-MEAN & 20.88 & $-$7.49\sig{***} & $4.94\times10^{-11}$ \\
\bottomrule
\end{tabular}
\end{center}
\end{table}

\begin{table}[h]
\caption{\textbf{Ours, no question conditioning}, 1628 test questions.}
\label{tab:full46}
\begin{center}\small
\begin{tabular}{lccc}
\toprule
Condition & Accuracy & Difference & $p$ \\
\midrule
REAL & 69.84 & -- & -- \\
MEAN & 69.78 & $-$0.06 & 1 \\
SWAP & 66.15 & $-$3.69\sig{***} & 0.000365 \\
NODE (real rows) & 67.75 & $-$2.09\sig{*} & 0.0367 \\
EDGE (rewired) & 69.35 & $-$0.49 & 0.636 \\
GSWAP (other graph) & 67.51 & $-$2.33\sig{*} & 0.0247 \\
\midrule
NOTXT-REAL & 30.90 & -- & -- \\
NOTXT-MEAN & 31.33 & $+$0.43 & 0.742 \\
\bottomrule
\end{tabular}
\end{center}
\end{table}

\clearpage

\section{The token's form: magnitude and residual conditions}
\label{app:magnitude}

These conditions probe the token's geometry rather than which factor it carries, which is why they are not part of Table~\ref{tab:taxonomy}. Table~\ref{tab:magnitude} reports them for the three graph-transformer models and the two graph-free controls on \texttt{ExplaGraphs}; the same conditions for every other model are in Appendix~\ref{app:interventions-full}.

\begin{table}[h]
\caption{Conditions that keep the token's direction and change its length, as differences from REAL. \sig{\dag}~$p<.10$, \sig{*}~$p<.05$, \sig{**}~$p<.01$, \sig{***}~$p<.001$, exact McNemar, unmarked $p\ge.10$. Blank cells were not run. \texttt{ExplaGraphs} 554 questions, \texttt{SceneGraphs} the first 2,000, \texttt{WebQSP} 1,628.}
\label{tab:magnitude}
\begin{center}
\small
\fitwidth{
\begin{tabular}{llccccc}
\toprule
Dataset & Model & REAL & zero token & norm $\rightarrow$ word & $\times0.5$ & $\times2$ \\
\midrule
\multirow{3}{*}{\texttt{ExplaGraphs}}
 & G-Retriever (GT) & 85.56 & $-81.95$\sig{***} & $-81.95$\sig{***} & $-8.12$\sig{***} & $-3.61$\sig{***} \\
 & CONST & 85.56 & $-83.39$\sig{***} & $-80.14$\sig{***} & $-23.83$\sig{***} & $-2.17$\sig{\dag} \\
 & CONST at word norm & 85.56 & $-83.39$\sig{***} & $0.00$ & $-80.69$\sig{***} & $-78.52$\sig{***} \\
\midrule
\multirow{2}{*}{\texttt{SceneGraphs}}
 & G-Retriever (GT) & 81.35 & $-70.05$\sig{***} & $-70.00$\sig{***} & $-4.50$\sig{***} & $-0.80$ \\
 & CONST & 81.40 & $-69.20$\sig{***} & $-66.60$\sig{***} & $-6.90$\sig{***} & $-1.95$\sig{**} \\
\midrule
\texttt{WebQSP} & G-Retriever (GT) & 70.33 & $-59.64$\sig{***} & $-65.48$\sig{***} & $-7.31$\sig{***} & $-4.24$\sig{***} \\
\bottomrule
\end{tabular}}
\end{center}
\end{table}

A trained model depends on the length its token had during training, in both directions, and a large token is not necessary: the constant held at word norm throughout training reaches the same accuracy as the others and then breaks when its length is changed afterwards.

\section{Complete graph-free controls}
\label{app:controls-full}

Table~\ref{tab:controls-full} gives the controls in full. All four models share G-Retriever's prompt, answer loss, optimiser and training schedule. \textsc{const} and \textsc{mlpq} also share its projector, so they differ from G-Retriever only in the token's source; \textsc{pt} differs in both the source and the parameterisation, since its vectors are optimised directly in the embedding space, and its capacity differs by more than two orders of magnitude. The \textsc{const}--\textsc{pt} gap therefore remains confounded between parameterisation and capacity, which is why the main text does not use it as a graph-versus-no-graph comparison.

\begin{table}[h]
\caption{Graph-free controls. Score is accuracy on \texttt{ExplaGraphs} and \texttt{SceneGraphs} and Hit@1 on \texttt{WebQSP}; parameter counts are from the \texttt{WebQSP} checkpoints with the LLM frozen and its embedding table excluded. Dashes are configurations we did not train.}
\label{tab:controls-full}
\begin{center}
\small
\fitwidth{
\begin{tabular}{lcccc}
\toprule
Model & Trainable parameters & \texttt{ExplaGraphs} & \texttt{SceneGraphs} & \texttt{WebQSP} \\
\midrule
G-Retriever & 31,492,099 & 85.56 & 81.35 & 70.33 \\
\textsc{const} & 10,492,928 & 85.56 & 81.40 & 71.25 \\
\textsc{mlpq} & 14,690,304 & 84.84 & -- & 70.70 \\
\textsc{pt} & 40,960 & 60.60 & 70.22 & 52.21 \\
\midrule
ours, Stage~2 & 20,990,979 & 85.74 & 72.50 & 69.90 \\
\bottomrule
\end{tabular}}
\end{center}
\end{table}

Accuracy does not track capacity: \textsc{pt} to \textsc{const} multiplies the trainable parameters by 256 and adds 19 Hit@1 on \texttt{WebQSP}, while \textsc{const} to our Stage~2 multiplies them by two and removes 1.35.

\section{Gold-answer likelihood}
\label{app:nll}

Table~\ref{tab:nll} reports the measurement defined in Equation~\ref{eq:nll}. It is a second reading of conditions that were already run, not a new experiment.

\begin{table}[h]
\caption{Paired gold-answer NLL differences, condition minus \textsc{real}, with 95\% bootstrap intervals; positive means the intervention made the gold answer less likely. Every cell whose $p$-value we measured is significant under a Wilcoxon signed-rank test (largest $p=10^{-4}$), so the criterion we apply is the effect size: the same weights and condition rerun give $+0.0003$ $[-0.0003,+0.0008]$, and we require $0.03$, a hundred times that floor, before reading a likelihood effect. The Qwen2.5-3B rows are the matched-budget pair of Table~\ref{tab:geometry-use}.}
\label{tab:nll}
\begin{center}
\small
\fitwidth{
\begin{tabular}{llcccc}
\toprule
Dataset & Model & REAL & MEAN & SWAP & GSWAP \\
\midrule
\multirow{6}{*}{\texttt{WebQSP}}
 & G-Retriever & 0.7558 & $+0.0131$ $[+.0097,+.0166]$ & $+0.0183$ $[+.0137,+.0229]$ & $+0.0182$ $[+.0136,+.0228]$ \\
 & ours & 0.8540 & $-0.0159$ $[-.0288,-.0041]$ & $+0.0895$ $[+.0727,+.1074]$ & $+0.0332$ $[+.0185,+.0461]$ \\
 & ours, no entity features & 0.8213 & $+0.0378$ $[+.0257,+.0506]$ & $+0.1111$ $[+.0954,+.1261]$ & $+0.0262$ $[+.0175,+.0348]$ \\
 & ours, no question & 0.8517 & $+0.0070$ $[-.0071,+.0203]$ & $+0.0848$ $[+.0671,+.1020]$ & $+0.0845$ $[+.0668,+.1015]$ \\
 & G-Retriever, Qwen2.5-3B & 1.0367 & $+0.0109$ $[+.0038,+.0185]$ & $+0.0734$ $[+.0623,+.0845]$ & $+0.0737$ $[+.0626,+.0850]$ \\
 & ours, Qwen2.5-3B & 1.0553 & $+0.0123$ $[+.0066,+.0177]$ & $+0.0168$ $[+.0106,+.0226]$ & $+0.0144$ $[+.0076,+.0211]$ \\
\midrule
\multirow{2}{*}{\texttt{SceneGraphs}}
 & G-Retriever & 0.2477 & $+0.0053$ $[+.0006,+.0100]$ & $+0.0169$ $[+.0102,+.0240]$ & $+0.0167$ $[+.0101,+.0238]$ \\
 & ours & 0.4122 & $-0.0287$ $[-.0556,-.0027]$ & $+0.0229$ $[-.0041,+.0498]$ & $+0.1665$ $[+.1353,+.1999]$ \\
\midrule
\multirow{2}{*}{\texttt{ExplaGraphs}}
 & G-Retriever & 0.2068 & $+0.0006$ $[-.0048,+.0059]$ & $+0.0014$ $[-.0063,+.0088]$ & $+0.0022$ $[-.0051,+.0095]$ \\
 & ours & 0.1696 & $-0.0047$ $[-.0160,+.0052]$ & $+0.0010$ $[-.0120,+.0130]$ & $+0.0032$ $[-.0091,+.0152]$ \\
\bottomrule
\end{tabular}}
\end{center}
\end{table}

Three readings. In the model whose token can only carry the graph, swapping the graph and swapping the token cost the same likelihood ($+0.0845$ against $+0.0848$), which is the internal consistency check for the instrument. On \texttt{SceneGraphs} the graph swap is seven times the token swap, the largest effect measured anywhere in this study and the distribution-level counterpart of the $-5.30$ accuracy drop, while the mean token \emph{lowers} the likelihood: an empty token is better than a wrong graph. On \texttt{ExplaGraphs} nothing moves for either model, which settles that dataset at the distribution level and not merely below the accuracy resolution.

\section{Complete token-content and use results}
\label{app:decodability}

Table~\ref{tab:decomposition-full} extends the Llama-2 results in
Table~\ref{tab:decomposition} with the complete Qwen2.5-3B replication and
diagnostic variants. The probe is trained on half of the test tokens and
evaluated on the other half by retrieving the corresponding question, answer,
or verbalised graph. The candidate sets contain 814 targets on
\texttt{WebQSP}, 1,000 on \texttt{SceneGraphs}, and 277 on
\texttt{ExplaGraphs}. A constant token therefore has median rank 407, 500,
and 139, respectively. Repeated \texttt{SceneGraphs} answers place the answer control between 476 and 500, depending on the held-out split.

\begin{table}[p]
\caption{\textbf{Complete token-content and use results across LLMs.}
Q, A, and G are median retrieval ranks; lower is better.
$\Delta$Score is the with-text \textsc{gswap} effect, and NLL reports
\textsc{swap}/\textsc{gswap}. Bold score effects satisfy $|\Delta|>2$ and
$p<.05$; bold NLL values satisfy the likelihood criterion. Diagnoses also use
Tables~\ref{tab:geometry-use}, \ref{tab:notext-pipeline}, and
\ref{tab:notext-qwen}. The binary \texttt{ExplaGraphs} answer is omitted.}
\label{tab:decomposition-full}
\begin{center}
\small
\fitwidth{
\begin{tabular}{llcccccl}
\toprule
Dataset & Model & Q rank & A rank & G rank
& $\Delta$Score \textsc{gswap}
& $\Delta$NLL S/GS & Diagnosis \\
\midrule

\multicolumn{8}{l}{\emph{Llama-2-7B}} \\
\cmidrule(lr){1-8}

\multirow{5}{*}{\texttt{WebQSP}}
 & G-Retriever
 & 293 & 317 & 369
 & $-0.55$
 & $+0.0183/+0.0182$
 & \textsc{soft prompt} \\
 & \textsc{mlpq}
 & 407 & 407 & 407
 & -- & -- & -- \\
 & \textbf{Ours}
 & 65 & 60 & 315
 & $-1.97$\sig{\dag}
 & $\mathbf{+0.0895}/\mathbf{+0.0332}$
 & \textsc{masked} \\
 & \quad no entity features
 & 34 & 51 & 303
 & $-1.97$\sig{*}
 & $\mathbf{+0.1111}/+0.0262$
 & \textsc{prior} \\
 & \quad no question
 & 120 & 134 & 329
 & $\mathbf{-2.33}$\sig{*}
 & $\mathbf{+0.0848}/\mathbf{+0.0845}$
 & \textsc{used} \\

\midrule
\multirow{5}{*}{\texttt{SceneGraphs}}
 & G-Retriever
 & 271 & 436 & 248
 & $-1.10$\sig{\dag}
 & $+0.0169/+0.0167$
 & \textsc{soft prompt} \\
 & \textsc{const}
 & 500 & 476 & 500
 & -- & -- & -- \\
 & \textbf{Ours}
 & 47 & 187 & 427
 & $\mathbf{-5.30}$\sig{***}
 & $+0.0229/\mathbf{+0.1665}$
 & \textsc{used} \\
 & \quad no entity features
 & 8 & 59 & 369
 & $+0.35$
 & $\mathbf{+0.1553}/+0.0045$
 & \textsc{prior} \\
 & \quad no question
 & 116 & 395 & 185
 & $-0.25$
 & $+0.0214/+0.0213$
 & \textsc{masked} \\

\midrule
\multirow{4}{*}{\texttt{ExplaGraphs}}
 & G-Retriever
 & 138 & -- & 139
 & $-1.08$
 & $+0.0014/+0.0022$
 & \textsc{soft prompt} \\
 & \textbf{Ours}
 & 54 & -- & 41
 & $-1.44$
 & $+0.0010/+0.0032$
 & \textsc{decodable} \\
 & \quad no entity features
 & 67 & -- & 67
 & $-0.36$
 & $+0.0026/+0.0056$
 & \textsc{decodable} \\
 & \quad no question
 & 84 & -- & 70
 & $+1.26$
 & $-0.0063/-0.0073$
 & \textsc{decodable} \\

\midrule
\multicolumn{8}{l}{\emph{Qwen2.5-3B, matched training budget}} \\
\cmidrule(lr){1-8}

\multirow{4}{*}{\texttt{WebQSP}}
 & G-Retriever
 & 348 & 357 & 388
 & $-1.60$\sig{\dag}
 & $\mathbf{+0.0734}/\mathbf{+0.0737}$
 & \textsc{soft prompt} \\
 & \textbf{Ours}
 & 153 & 156 & 349
 & $\mathbf{-2.09}$\sig{**}
 & $+0.0168/+0.0144$
 & \textsc{used} \\
 & \quad no entity features
 & 109 & 110 & 344
 & $-0.86$
 & $\mathbf{+0.0321}/+0.0166$
 & \textsc{prior} \\
 & \quad no question
 & 188 & 198 & 352
 & $\mathbf{-2.33}$\sig{**}
 & $\mathbf{+0.0336}/\mathbf{+0.0333}$
 & \textsc{used} \\

\midrule
\multirow{4}{*}{\texttt{SceneGraphs}}
 & G-Retriever
 & 326 & 446 & 340
 & $-1.15$\sig{*}
 & $+0.0127/+0.0125$
 & \textsc{soft prompt} \\
 & \textbf{Ours}
 & 112 & 353 & 450
 & $\mathbf{-6.50}$\sig{***}
 & $\mathbf{+0.0392}/\mathbf{+0.0410}$
 & \textsc{used} \\
 & \quad no entity features
 & 29 & 121 & 402
 & $-0.60$
 & $\mathbf{+0.0373}/+0.0019$
 & \textsc{prior} \\
 & \quad no question
 & 155 & 408 & 256
 & $-1.90$\sig{**}
 & $+0.0174/+0.0172$
 & \textsc{decodable} \\

\midrule
\multirow{4}{*}{\texttt{ExplaGraphs}}
 & G-Retriever
 & 139 & -- & 138
 & $\mathbf{-4.51}$\sig{*}
 & $+0.0146/+0.0138$
 & \textsc{used} \\
 & \textbf{Ours}
 & 85 & -- & 62
 & $+0.36$
 & $+0.0008/+0.0042$
 & \textsc{decodable} \\
 & \quad no entity features
 & 92 & -- & 87
 & $-1.99$
 & $+0.0188/+0.0177$
 & \textsc{decodable} \\
 & \quad no question
 & 109 & -- & 99
 & $-1.44$
 & $+0.0017/+0.0008$
 & \textsc{decodable} \\

\bottomrule
\end{tabular}}
\end{center}
\end{table}

The control rows validate the probe: \textsc{mlpq} on \texttt{WebQSP} and
\textsc{const} on \texttt{SceneGraphs} remain at the corresponding median
control ranks. The cross-fitted \texttt{WebQSP} model was not probed and is
therefore omitted.

The Qwen replication reinforces the distinction between recoverability and
use. On \texttt{SceneGraphs}, our model produces the largest
graph-replacement effect in the study ($-6.50$), although its graph rank is
close to the constant control ($450$ versus $500$). On
\texttt{ExplaGraphs}, the graph is substantially more recoverable
($62$ versus $139$), yet replacing it has no resolved effect. A linear probe
can therefore fail to recover information that affects the LLM or recover
information that the LLM ignores.

Qwen G-Retriever on \texttt{ExplaGraphs} provides a further exception: its
tokens remain geometrically collapsed, yet \textsc{gswap} causes a resolved
$-4.51$-point effect. Collapse indicates that little input-specific variation
remains, but does not establish complete insensitivity to that variation. For collapsed baselines, nearly identical \textsc{swap} and \textsc{gswap} likelihood effects indicate generic sensitivity to token perturbation rather than identified graph-specific use.

Three caveats bound the decodability results. The sentence encoder truncates
long graph verbalisations, a single sentence vector compresses a subgraph
lossily, and the fitted map is linear. Failure to decode is therefore not
proof that information is absent, whereas successful decoding is evidence
that it is recoverable by this probe.

The Qwen \texttt{ExplaGraphs} no-entity and no-question variants use seed~1
because their seed-0 runs diverged and scored below the majority baseline.
\section{Validation of the design choices}
\label{app:design}

\textbf{Question conditioning in Stage 1.} Table~\ref{tab:stage1} reports Stage~1 measured on its own, without the LLM. Top-1 is the fraction of validation samples whose own target is the most similar of all validation targets; the graph gap is the mean cosine between representations of different questions about the same graph minus the mean cosine across graphs, and is defined only on \texttt{SceneGraphs}, the one benchmark that asks several questions about one graph. Removing the question conditioning drops top-1 from 0.433 to 0.084, because the answer depends on what is asked and the encoder needs the question to select the relevant part of the scene. The graph gap shows that what it learns is not merely a copy of the question.

\begin{table}[h]
\caption{Stage 1 measured without the LLM. Top-1 is computed against all validation targets of the benchmark, 19,997 on \texttt{SceneGraphs}.}
\label{tab:stage1}
\begin{center}
\small
\fitwidth{
\begin{tabular}{llcc}
\toprule
Run & Target & Validation top-1 & Graph gap \\
\midrule
\texttt{SceneGraphs}, with question & full answer sentence & 0.433 & $0.035 \rightarrow 0.204$ \\
\texttt{SceneGraphs}, without question & full answer sentence & 0.084 & Same embeddings, N/A \\
\texttt{SceneGraphs}, scene-disjoint halves & full answer sentence & 0.376 & $0.035 \rightarrow 0.264$ \\
\texttt{WebQSP}, half split & gold answers & 0.469 & N/A \\
\texttt{ExplaGraphs}, half split & verbalised graph & 0.501 & N/A \\
\bottomrule
\end{tabular}}
\end{center}
\end{table}

\textbf{Stage-1 objective.} With $\mathbf{h}_i$ the pooled encoder output, $\mathbf{s}_i=\mathrm{LM}(c_i)$ the frozen sentence embedding of the target $c_i$, batches of size $B$, temperature $\kappa$ and a memory bank $\mathcal{M}$ of $M$ other samples' targets, Stage~1 minimises
\begin{equation}
\mathcal{L}_{\mathrm{S1}} = -\frac{1}{B}\sum_{i=1}^{B}\log\frac{\exp\!\big(\cos(\mathbf{h}_i,\mathbf{s}_i)/\kappa\big)}{\exp\!\big(\cos(\mathbf{h}_i,\mathbf{s}_i)/\kappa\big)+\sum_{j\in\mathcal{M}}\exp\!\big(\cos(\mathbf{h}_i,\mathbf{s}_j)/\kappa\big)},
\label{eq:infonce}
\end{equation}

\textbf{Stage-1 targets.}
The target is the gold answer on \texttt{WebQSP} and the full answer sentence
on \texttt{SceneGraphs}, which contains 47,487 distinct targets.
\texttt{ExplaGraphs} has only two answer labels, so an answer target cannot
identify a sample and we use the verbalised graph instead.

\textbf{Freezing the encoder in Stage 2.}
With a trainable encoder the answer loss re-collapses the token even with a
GELU projector, to cosine $0.9139$ (Table~\ref{tab:mechanism}). We verified
that the frozen encoder reproduces the intended graph token to within
$1.2\times10^{-7}$.

\textbf{Parameterisation of the prompt token.}
The prompt token is a learned constant passed through a Sigmoid projector,
which matches G-Retriever on all three datasets
(Table~\ref{tab:collapsed-role}). Raw prompt tuning scores 11--25 points
lower, although this comparison does not isolate parameterisation from
capacity.
\textbf{Alternatives that did not work.} Table~\ref{tab:failed} lists three, all trained with the cross-fitted recipe on \texttt{WebQSP} and therefore comparable with the cross-fitted joint model rather than with the main-text numbers. Each answers a question about the design.

\emph{Must the two tokens be trained together?} If the split alone were what matters, adding the prompt token to a converged projector should work as well as training both from the start. It does not: accuracy stays at 63.64 and the best epoch is the first, because by then the model has settled into using the graph token for both roles and the new token has nothing to take over.

\emph{Is a second token needed at all, or only some way of stopping the graph token from being relied on as a constant?} Dropout on the projector is the cheaper alternative, and it leaves one token doing both jobs: accuracy reaches 64.56 and the token drifts back towards collapse, at cosine 0.7563 against 0.1778 for the joint model.

\emph{Could the redundancy be removed from the data instead of the architecture?} Hiding 40\% of the verbalised edges during training makes the text an unreliable channel, which should force the token to carry the graph. Instead the model learns to compensate for missing text, and at test time, when the text is complete again, substitute tokens are as good as the real one or better, with other graphs' node features worth $+2.03$ points ($p=0.03$).

\begin{table}[h]
\caption{Alternatives to the joint Stage 2. \texttt{WebQSP}, cross-fitted recipe, 1,628 test questions. MEAN, SWAP and NODE are differences from each model's own REAL, with the exact McNemar test: \sig{\dag}~$p<.10$, \sig{*}~$p<.05$, \sig{**}~$p<.01$, \sig{***}~$p<.001$, unmarked $p\ge0.1$.}
\label{tab:failed}
\begin{center}
\small
\fitwidth{
\begin{tabular}{lccccc}
\toprule
Variant & Hit@1 & Token cosine & MEAN & SWAP & NODE \\
\midrule
Joint Stage 2, as used in the paper & 69.53 & 0.1778 & $-1.54$\sig{\dag} & $-3.19$\sig{**} & $-3.26$\sig{***} \\
Projector only & 63.14 & 0.5700 & $-3.44$\sig{***} & $-5.34$\sig{***} & $-1.66$ \\
Prompt token added after the projector & 63.64 & 0.5700 & $-2.95$\sig{***} & $-5.22$\sig{***} & $-2.76$\sig{***} \\
Dropout 0.5 on the projector & 64.56 & 0.7563 & $-4.24$\sig{***} & $-3.69$\sig{***} & $-1.29$ \\
40\% of verbalised edges hidden in training & 67.20 & 0.1847 & $+1.35$ & $-0.31$ & $+2.03$\sig{**} \\
\bottomrule
\end{tabular}}
\end{center}
\end{table}

\section{Pipeline and robustness ablations}
\label{app:pipeline-ablations}

Table~\ref{tab:pipeline-ablations} extends
Table~\ref{tab:geometry-use} with the complete set of retrained pipeline
variants and the principal fixed-weight interventions. The no-entity and
no-question rows remove an input during both training and evaluation; they are
therefore pipeline ablations rather than interventions on a fixed model.
Additional relation and geometric interventions are reported in
Appendices~\ref{app:interventions-full} and~\ref{app:magnitude}.

\begin{table}[p]
\caption{\textbf{Pipeline variants under fixed-weight interventions.}
\emph{Ours} denotes the complete two-token pipeline; indented rows are
retrained diagnostic variants. \emph{Projector only} is the single-token model
with the aligned frozen encoder but no separate prompt token. Intervention
columns report condition minus \textsc{real}, so negative values indicate
worse performance. Lower token cosine indicates greater input dependence.
Dashes denote conditions not evaluated.}
\label{tab:pipeline-ablations}
\begin{center}
\small
\fitwidth{
\begin{tabular}{llccccccc}
\toprule
Dataset & Model & \textsc{real} & Token cos. &
$\Delta$\textsc{mean} & $\Delta$\textsc{swap} &
$\Delta$\textsc{node} & $\Delta$\textsc{edge} &
$\Delta$\textsc{gswap} \\
\midrule

\multicolumn{9}{l}{\emph{Llama-2-7B}} \\
\cmidrule(lr){1-9}

\multirow{7}{*}{\texttt{WebQSP}}
 & projector only
 & 64.62 & 0.543
 & $-14.74$\sig{***} & $-9.89$\sig{***}
 & $-2.40$\sig{*} & $+0.06$ & -- \\
 & \textbf{Ours}
 & 69.90 & 0.142
 & $-3.13$\sig{***} & $-2.15$\sig{*}
 & $-1.23$ & $-0.80$ & $-1.97$\sig{\dag} \\
 & \quad no entity features
 & 72.73 & 0.125
 & $-6.14$\sig{***} & $-5.65$\sig{***}
 & $-0.61$ & $-0.43$ & $-1.97$\sig{*} \\
 & \quad no question
 & 69.84 & 0.169
 & $-0.06$ & $-3.69$\sig{***}
 & $-2.09$\sig{*} & $-0.49$ & $-2.33$\sig{*} \\
 & \quad cross-fitted
 & 69.53 & 0.178
 & $-1.54$\sig{\dag} & $-3.19$\sig{**}
 & $-3.26$\sig{***} & $-0.55$ & $-2.83$\sig{**} \\
 & G-Retriever + LoRA
 & 74.32 & 0.997
 & $-0.43$ & $-0.74$
 & $-0.98$ & $-0.12$ & -- \\
 & \textbf{Ours} + LoRA
 & 74.57 & 0.170
 & $-1.11$ & $-2.33$\sig{**}
 & $-0.80$ & $+0.25$ & -- \\

\midrule
\multirow{4}{*}{\texttt{SceneGraphs}}
 & projector only
 & 69.15 & 0.519
 & $-20.20$\sig{***} & $-35.05$\sig{***}
 & $-8.90$\sig{***} & $-2.45$\sig{*} & -- \\
 & \textbf{Ours}
 & 72.50 & 0.133
 & $-1.40$ & $-3.40$\sig{***}
 & $-5.10$\sig{***} & $-1.05$ & $-5.30$\sig{***} \\
 & \quad no entity features
 & 79.65 & 0.125
 & $-5.75$\sig{***} & $-9.60$\sig{***}
 & $+0.10$ & $+0.40$ & $+0.35$ \\
 & \quad no question
 & 76.35 & 0.225
 & $-1.15$ & $-0.25$
 & $-0.75$ & $+0.35$ & $-0.25$ \\

\midrule
\multirow{4}{*}{\texttt{ExplaGraphs}}
 & projector only
 & 57.76 & 0.323
 & $+1.08$ & $-6.68$\sig{*}
 & $-1.26$ & $-6.14$\sig{*} & -- \\
 & \textbf{Ours}
 & 85.74 & 0.099
 & $+0.36$ & $-0.54$
 & $-0.54$ & $+1.08$ & $-1.44$ \\
 & \quad no entity features
 & 84.48 & 0.113
 & $+3.07$\sig{**} & $+0.00$
 & $+0.18$ & $+0.18$ & $-0.36$ \\
 & \quad no question
 & 87.55 & 0.108
 & $+1.08$ & $+0.90$
 & $+0.18$ & $-0.18$ & $+1.26$ \\

\midrule
\multicolumn{9}{l}{\emph{Qwen2.5-3B, matched training budget}} \\
\cmidrule(lr){1-9}

\multirow{4}{*}{\texttt{WebQSP}}
 & G-Retriever
 & 57.13 & 0.997
 & $-0.55$ & $-1.41$
 & $-0.31$ & $+0.06$ & $-1.60$\sig{\dag} \\
 & \textbf{Ours}
 & 65.79 & 0.200
 & $-2.09$\sig{**} & $-1.90$\sig{**}
 & $-0.80$ & $-1.35$\sig{*} & $-2.09$\sig{**} \\
 & \quad no entity features
 & 58.54 & 0.180
 & $-2.03$\sig{**} & $-1.84$\sig{*}
 & $-0.06$ & $-1.23$\sig{\dag} & $-0.86$ \\
 & \quad no question
 & 60.14 & 0.194
 & $-0.86$ & $-2.33$\sig{**}
 & $-2.70$\sig{***} & $-0.55$ & $-2.33$\sig{**} \\

\midrule
\multirow{4}{*}{\texttt{SceneGraphs}}
 & G-Retriever
 & 76.05 & 0.997
 & $-0.30$ & $-1.05$\sig{\dag}
 & $-0.35$ & $+0.30$ & $-1.15$\sig{*} \\
 & \textbf{Ours}
 & 72.65 & 0.139
 & $-7.00$\sig{***} & $-5.75$\sig{***}
 & $-4.40$\sig{***} & $-1.65$\sig{*} & $-6.50$\sig{***} \\
 & \quad no entity features
 & 75.95 & 0.199
 & $-2.35$\sig{**} & $-4.60$\sig{***}
 & $-0.20$ & $-0.55$ & $-0.60$ \\
 & \quad no question
 & 75.40 & 0.557
 & $-0.75$ & $-1.40$\sig{*}
 & $-0.55$ & $+0.50$ & $-1.90$\sig{**} \\

\midrule
\multirow{4}{*}{\texttt{ExplaGraphs}}
 & G-Retriever
 & 74.37 & 0.998
 & $-1.62$ & $-2.17$
 & $-0.72$ & $-2.71$ & $-4.51$\sig{*} \\
 & \textbf{Ours}
 & 83.94 & 0.103
 & $+0.90$ & $+0.00$
 & $+0.00$ & $+0.36$ & $+0.36$ \\
 & \quad no entity features
 & 80.14 & 0.116
 & $+2.17$ & $-3.07$
 & $-0.18$ & $-0.72$ & $-1.99$ \\
 & \quad no question
 & 74.37 & 0.118
 & $-1.44$ & $-0.54$
 & $+0.54$ & $+0.00$ & $-1.44$ \\

\bottomrule
\end{tabular}}
\end{center}
\end{table}

Removing entity features has a benchmark-dependent effect on task performance:
it hurts Qwen on \texttt{WebQSP} and both LLMs on \texttt{ExplaGraphs}, but
improves both LLMs on \texttt{SceneGraphs}. Its effect on token use is more
consistent. Across both LLMs and all three benchmarks, the no-entity model
remains sensitive to \textsc{mean} or \textsc{swap}, while no
\textsc{gswap} effect clears the answer-level criterion. The token can
therefore remain strongly used after its resolved graph-specific contribution
disappears.

Removing question conditioning produces resolved graph use on
\texttt{WebQSP} with both LLMs. On Qwen \texttt{SceneGraphs}, the
\textsc{gswap} effect is statistically significant but does not clear the
two-point criterion; neither LLM shows resolved graph use in the corresponding
\texttt{ExplaGraphs} variant.

For the standard Qwen \texttt{SceneGraphs} pipeline, the earliest checkpoint
was selected and validation loss subsequently increased
($0.2668, 0.2741, 0.2737, 0.2981, 0.3396$). The same immediate overfitting
occurs with Llama-2.

\subsection{Evaluation without the verbalised graph}
\label{app:notext}

Tables~\ref{tab:notext-pipeline} and~\ref{tab:notext-qwen} evaluate the trained
models after deleting the verbalised graph from the prompt. The graph token is
then the only possible graph channel. \textsc{notxt} is the resulting absolute
score, and intervention columns report condition minus \textsc{notxt}. These
evaluations are fixed-weight interventions: the models are not retrained
without text.

A \textsc{notxt-mean} effect establishes dependence on sample-specific token
content, but does not identify that content as graph-specific.
\textsc{notxt-gswap} provides the graph-specific test because it changes the
graph while preserving the remaining inputs. We therefore do not infer that
graph information is masked by text from \textsc{mean} alone.

\begin{table}[p]
\caption{\textbf{Llama-2 evaluation without the verbalised graph.}
Intervention columns report differences from \textsc{notxt}; the final column
gives the corresponding score with the verbalised graph present. Bold
\textsc{gswap} effects satisfy $|\Delta|>2$ and $p<.05$. Dashes denote
conditions not evaluated.}
\label{tab:notext-pipeline}
\begin{center}
\small
\fitwidth{
\begin{tabular}{llccccc}
\toprule
Dataset & Model & \textsc{notxt}
& $\Delta$\textsc{mean} & $\Delta$\textsc{swap}
& $\Delta$\textsc{gswap} & With text \\
\midrule

\multirow{4}{*}{\texttt{WebQSP}}
 & G-Retriever
 & 41.34 & $+1.04$ & $-0.18$ & $-0.86$ & 70.33 \\
 & \textbf{Ours}
 & 28.75 & $-2.09$\sig{*} & $-2.40$\sig{*}
 & $\mathbf{-2.27}$\sig{*} & 69.90 \\
 & \quad no entity features
 & 28.38 & $-7.49$\sig{***} & -- & -- & 72.73 \\
 & \quad no question
 & 30.90 & $+0.43$ & -- & -- & 69.84 \\

\midrule
\multirow{4}{*}{\texttt{SceneGraphs}}
 & G-Retriever
 & 30.20 & $+0.45$ & $-1.40$ & -- & 81.35 \\
 & \textbf{Ours}
 & 32.60 & $-6.60$\sig{***} & $-10.85$\sig{***}
 & $\mathbf{-3.10}$\sig{**} & 72.50 \\
 & \quad no entity features
 & 30.10 & $-12.85$\sig{***} & -- & -- & 79.65 \\
 & \quad no question
 & 26.15 & $-8.25$\sig{***} & $-3.25$\sig{**}
 & $\mathbf{-3.30}$\sig{**} & 76.35 \\

\midrule
\multirow{4}{*}{\texttt{ExplaGraphs}}
 & G-Retriever
 & 66.43 & $-0.90$ & $0.00$ & -- & 85.56 \\
 & \textbf{Ours}
 & 69.68 & $+0.18$ & $+0.18$ & $-0.54$ & 85.74 \\
 & \quad no entity features
 & 70.22 & $+0.72$ & -- & -- & 84.48 \\
 & \quad no question
 & 78.88 & $+1.08$ & -- & -- & 87.55 \\

\bottomrule
\end{tabular}}
\end{center}
\end{table}

\begin{table}[p]
\caption{\textbf{Qwen2.5-3B evaluation without the verbalised graph.}
Intervention columns report differences from \textsc{notxt}; the final column
gives the corresponding score with the verbalised graph present. Bold
\textsc{gswap} effects satisfy $|\Delta|>2$ and $p<.05$. Dashes denote
conditions not evaluated.}
\label{tab:notext-qwen}
\begin{center}
\small
\fitwidth{
\begin{tabular}{llccccc}
\toprule
Dataset & Model & \textsc{notxt}
& $\Delta$\textsc{mean} & $\Delta$\textsc{swap}
& $\Delta$\textsc{gswap} & With text \\
\midrule

\multirow{4}{*}{\texttt{WebQSP}}
 & G-Retriever
 & 29.30 & $+0.86$ & $-2.03$\sig{\dag} & $-1.84$\sig{\dag} & 57.13 \\
 & \textbf{Ours}
 & 34.34 & $+0.31$ & $+1.11$ & $+1.11$ & 65.79 \\
 & \quad no entity features
 & 30.10 & $+1.97$\sig{*} & -- & -- & 58.54 \\
 & \quad no question
 & 32.74 & $+0.61$ & -- & -- & 60.14 \\

\midrule
\multirow{4}{*}{\texttt{SceneGraphs}}
 & G-Retriever
 & 39.50 & $+1.95$\sig{*} & -- & -- & 76.05 \\
 & \textbf{Ours}
 & 43.90 & $-7.10$\sig{***} & -- & -- & 72.65 \\
 & \quad no entity features
 & 37.75 & $-2.90$\sig{*} & -- & -- & 75.95 \\
 & \quad no question
 & 34.70 & $-3.15$\sig{***} & -- & -- & 75.40 \\

\midrule
\multirow{4}{*}{\texttt{ExplaGraphs}}
 & G-Retriever
 & 78.88 & $0.00$ & -- & -- & 74.37 \\
 & \textbf{Ours}
 & 82.13 & $-1.62$ & $+0.18$ & $-0.18$ & 83.94 \\
 & \quad no entity features
 & 75.99 & $-4.51$\sig{*} & -- & -- & 80.14 \\
 & \quad no question
 & 76.17 & $-7.40$\sig{***} & -- & -- & 74.37 \\

\bottomrule
\end{tabular}}
\end{center}
\end{table}

With Llama-2, removing the verbalised graph exposes resolved graph-specific
effects for the standard pipeline on \texttt{WebQSP} and
\texttt{SceneGraphs}, and for the question-free \texttt{SceneGraphs} variant.
No corresponding effect is found on \texttt{ExplaGraphs}.

With Qwen2.5-3B, several models depend on sample-specific token content after
the text is removed, most strongly on \texttt{SceneGraphs} and in the
\texttt{ExplaGraphs} ablations. Most of these rows currently contain only
\textsc{mean}; they therefore establish token dependence but do not identify
the graph as the information being used. The standard Qwen
\texttt{ExplaGraphs} pipeline instead shows no resolved text-free token effect,
despite its improved task score and decodable graph information.

\section{Cross-fitting: is the token's usefulness Stage-1 memorisation?}
\label{app:xfit}

Stage 1 partly memorises its training answers, with top-1 about 0.9 on training samples against about 0.5 on validation samples. A reader may therefore ask whether the LLM's reliance on the token in Section~\ref{sec:rq3} is reliance on answers the encoder memorised. We test this by training Stage~1 on one half of the training questions and Stage~2 on the other, splitting \texttt{SceneGraphs} by scene since one scene carries many questions. No test data is involved in either regime: Stage~1 uses training labels only, and on \texttt{SceneGraphs} the training and test sets share no scenes.

Table~\ref{tab:xfit} shows that the reliance is partly memorisation. When the two stages see disjoint halves the interventions still hurt, but less, while accuracy rises towards the baseline trained on the same half. The two regimes are two operating points of the trade-off described in Section~\ref{sec:rq3}.

\begin{table}[h]
\caption{Cross-fitted runs, in which Stage~1 and Stage~2 see disjoint halves of the training questions. Accuracy is compared with the G-Retriever baseline trained on the same half; the intervention columns are differences from the model's own REAL, with the exact McNemar test. \texttt{WebQSP} 1,628 test questions, \texttt{SceneGraphs} the first 2,000, \texttt{ExplaGraphs} 554.}
\label{tab:xfit}
\begin{center}
\small
\fitwidth{
\begin{tabular}{llcccccc}
\toprule
Dataset & Stage 2 & Accuracy & Baseline & MEAN & SWAP & NODE & EDGE \\
\midrule
\texttt{WebQSP} & projector only & 63.14 & 70.52 & $-3.44$\sig{***} & $-5.34$\sig{***} & $-1.66$ & $+1.04$ \\
\texttt{WebQSP} & projector + prompt & 69.53 & 70.58 & $-1.54$\sig{\dag} & $-3.19$\sig{**} & $-3.26$\sig{***} & $-0.55$ \\
\texttt{SceneGraphs} & projector + prompt, run 1 & 76.90 & 78.57 & $-1.50$\sig{\dag} & $-4.45$\sig{***} & $-1.05$ & $-1.95$\sig{***} \\
\texttt{SceneGraphs} & projector + prompt, run 2 & 77.55 & 78.57 & $-1.10$ & $-2.45$\sig{**} & $+1.00$ & $+0.30$ \\
\texttt{ExplaGraphs} & projector + prompt & 76.35 & 80.87 & $-3.61$\sig{*} & $+1.08$ & $+0.72$ & $+0.18$ \\
\bottomrule
\end{tabular}}
\end{center}
\end{table}

Rewiring the edges costs the cross-fitted joint model 1.95 points on \texttt{SceneGraphs} in one run and nothing in the other, so structure does not reach the LLM reliably in this regime either.

\clearpage

\section{Initialisations and evaluation noise}
\label{app:seeds}

Table~\ref{tab:seeds} reports three initialisations of G-Retriever and of the joint model on \texttt{WebQSP}, trained with the cross-fitted recipe on the same half. The spread across initialisations, 1.3 to 1.5 points, is larger than the difference between the two models, which is what licenses the word parity in Section~\ref{sec:rq3}. Pooling the discordant questions over the three runs, $n=4{,}884$, leaves our token used and G-Retriever's not.

\begin{table}[h]
\caption{Three initialisations on \texttt{WebQSP}, cross-fitted recipe, 1,628 test questions each. The intervention columns are differences from REAL pooled over the three runs, with the exact McNemar test.}
\label{tab:seeds}
\begin{center}
\small
\fitwidth{
\begin{tabular}{lccccccc}
\toprule
Model & Run 1 & Run 2 & Run 3 & Mean $\pm$ sd & MEAN & SWAP & NODE \\
\midrule
G-Retriever & 70.58 & 67.75 & 70.02 & $69.45\pm1.50$ & $\le0.53$ & $\le0.53$ & $\le0.53$ \\
Ours, joint & 69.53 & 66.89 & 68.43 & $68.28\pm1.33$ & $-1.19$\sig{**} & $-1.45$\sig{**} & $-1.84$\sig{***} \\
\bottomrule
\end{tabular}}
\end{center}
\end{table}

Evaluation noise between a training-script score and a separate scoring run is reported in Appendix~\ref{app:protocol}.

\section{The verbalised graph is not enough}
\label{app:motivating}

In this section, we bring some qualitative evidence that LLMs generally struggle to deal with verbalized graphs only. \texttt{ExplaGraphs} is the clean test, because its graph is not retrieved: it is the human-annotated explanation graph of the stance, given in full, so an error cannot be blamed on missing information. We give a frozen Llama-2-7b-chat\footnote{The chat model is used here because the base Llama-2-7b does not answer zero-shot with this prompt; every other experiment in the paper uses the base model, as G-Retriever does.} the complete graph as a list of triples inside G-Retriever's prompt, zero-shot, and score it with the benchmark's evaluator on all 554 test questions.

The majority class is 58.8\%. The arguments alone give 61.0\%, and adding the complete explanation graph gives 64.3\%, with 198 of 554 answers still wrong. The graph fixes 68 questions and breaks 50, so verbalising it misleads the model about as often as it helps. Part of this is a bias towards ``support'', which accounts for 451 of 554 answers and for 161 of the 198 errors, but not all of it: Figure~\ref{fig:example} shows an error in which the graph states the opposition explicitly.

\begin{figure}[h]
\begin{center}
\fbox{\begin{minipage}{0.93\linewidth}
\small
\textbf{\texttt{ExplaGraphs} \#2066.}\quad gold: \emph{counter} \quad|\quad arguments only: \emph{counter} \quad|\quad arguments + full graph: \textbf{\emph{support}}\\[2pt]
\textit{Argument 1:} Legalized organ trade would exploit the poor.\\
\textit{Argument 2:} Legalized organ trade would aid the poor, by giving them a large sum of money to improve their lives.\\
\textit{Graph:} legalized organ trade $\xrightarrow{\text{causes}}$ money for organs $\xrightarrow{\text{capable of}}$ improving finances $\xrightarrow{\text{capable of}}$ improve their lives $\xrightarrow{\text{has context}}$ aid the poor;\quad aid the poor $\xrightarrow{\textbf{is not a}}$ exploit the poor
\end{minipage}}
\end{center}
\caption{\textbf{The verbalised graph can turn a right answer into a wrong one.} Frozen Llama-2-7b-chat, zero-shot, G-Retriever's \texttt{ExplaGraphs} prompt with the complete human-annotated graph as triples. The model is right from the arguments alone and wrong once the graph, which states the opposition explicitly in the relation set in bold, is added. This happens on 50 of the 554 test questions.}
\label{fig:example}
\end{figure}

\section{Extended related work}
\label{app:related}

\textbf{Graph soft prompting.}
A growing line of work connects graph encoders to frozen or adapted LLMs by projecting graph representations into continuous tokens.
GraphToken~\citep{fatemi2024talk} learns an encoding function that emits soft tokens for a frozen LLM on synthetic graph-reasoning tasks, while GraphLLM~\citep{chai2024graphllm} targets graph reasoning with a tuned decoder and GNP~\citep{tian2024gnp} adds a cross-modality pooling module for multiple-choice knowledge-graph questions.
Among these, GNP and G-Retriever~\citep{he2024gretriever} are the only ones that retrieve a question-specific subgraph before encoding; the others operate on the full graph or on synthetic inputs.
G-Retriever additionally verbalises the retrieved subgraph in the prompt, creating a redundant configuration that is the focus of our audit.
GraphGPT~\citep{tang2024graphgpt} combines a text-grounded structural encoder with a lightweight graph--text alignment projector and dual-stage instruction tuning. These systems share the interface we study: a pooled graph representation, a trained projector, and one or a few tokens prepended to the prompt.

\textbf{Multi-token and pooling-based methods.}
Very few works have explored multi-token methods for graph-to-LLM systems.
The single-token bottleneck created by mean pooling over a subgraph has only recently been identified as a structural limitation.
\citet{grover2026one} argue that aggressive mean pooling compresses entire graph substructures into a single token, creating a severe information bottleneck, and propose multi-token pooling operators to increase the bandwidth of the graph-to-LLM interface.
Their work also reports that pooling methods are unstable under soft prompt tuning with a frozen LLM and require LoRA fine-tuning to stabilise.
This is complementary to our diagnosis: we study a single-token approach and show that its collapse is associated with both the projector parameterisation and end-to-end answer training.

\textbf{Graph--language alignment.}
Alignment-based methods add an auxiliary objective intended to bring graph representations closer to language representations.
Align-GRAG~\citep{xu2025aligngrag} is closest to our Stage~1: it contrastively aligns a pooled graph representation with a textual target using KL divergence and contrastive losses.
G-Retriever~\citep{he2024gretriever} can also be viewed as an implicit alignment method, since its projector is trained to map graph representations into the LLM's embedding space under the answer loss.
Our use of contrastive pretraining is similar to Align-GRAG, but our question is different.
Similarity between graph and text spaces does not establish that different graphs produce different tokens after projection, nor that the LLM uses their graph-specific content.
We therefore complement representation alignment with measurements of token variation and fixed-weight interventions on the trained model.

\textbf{Why graph use is not guaranteed.}
Two recent works already point out our diagnosis from different perspectives.
\citet{GTLLM} prove that the autoregressive language objective of graph-tokenizing LLMs decomposes into a textual dependency term and a graph--text alignment term, and that current methods optimize only the former explicitly.
Graph--text alignment is therefore implicit: the training objective does not reward the model for using graph information, and a text-dominant solution can be optimal under the loss.
\citet{petkar2026graph} show empirically that existing graph--language benchmarks can be matched by graph-only or language-only baselines, and introduce CLEGR to require compositional reasoning over structure and text; even there, raw soft-prompted LLMs remain competitive with GNN-backed models.
Together, these results establish that neither the training objective nor the benchmark forces a model to use graph information.
Our work is complementary in level rather than in kind: we study the interface itself and show that a graph token can collapse to a near-constant vector even when the encoder representations are distinct, and that preventing collapse does not guarantee that the LLM uses the token's graph-specific content.
This is what motivates the diagnostic protocol we introduce.

\textbf{Retrieval for graph-grounded generation.}
G-Retriever~\citep{he2024gretriever} formulates subgraph retrieval as a prize-collecting Steiner tree problem~\citep{bienstock1993pcst} and is the reference implementation we study.
GRAG~\citep{hu2024grag} retrieves subgraphs for generation with a different scoring rule, emphasising the importance of subgraph structure and topology.
GraSP~\citep{wang2026topology} encodes subgraphs into soft prompts using GNNs for multi-hop KBQA and combines this with a two-stage LLM reasoning procedure.
GNN-RAG~\citep{mavromatis2024gnn} uses a GNN to retrieve reasoning paths that are then verbalised, rather than projecting them as soft tokens.
Retrieval quality is orthogonal to our question: our interventions keep the retrieved subgraph fixed and change only the token computed from it.

\textbf{Parameter-efficient tuning and continuous prompting.}
Fine-tuning an LLM can be done in several ways.
Full-weight fine-tuning updates all parameters.
Parameter-efficient methods update a small subset or add new parameters, and include LoRA~\citep{lora}, which learns low-rank updates to existing weights, and continuous prompting methods.
Continuous prompting conditions a frozen language model with learned continuous vectors.
This family includes prompt tuning~\citep{pt}, which learns input-independent embeddings, and prefix-tuning~\citep{li2021prefixtuningoptimizingcontinuousprompts}, which learns continuous prefixes made available across the model's layers.
A graph token is a form of continuous prompting, but with a different intended role: it is computed from each input graph and is therefore expected to preserve input-specific information, unlike standard prompt tuning where the prompt is shared across all inputs.
Our results connect these settings by showing that Graph Soft prompting, as it is, collapses into a nearly input-independent vector and consequently behaves like a conventional soft prompt.
LoRA is used in our work only as a robustness check on the frozen-LLM setting, and the effect is consistent.

\end{document}